%% file: main.tex
\pdfoutput=1
\PassOptionsToPackage{most}{tcolorbox}
\PassOptionsToPackage{hidelinks}{hyperref}
\documentclass[11pt]{article}
\usepackage[final]{acl}
\usepackage{times}
\usepackage{latexsym}
\usepackage[T1]{fontenc}
\usepackage[utf8]{inputenc}
\usepackage{microtype}
\usepackage{inconsolata}
\usepackage{graphicx}
\usepackage{booktabs}
\usepackage{makecell}
\usepackage{multirow}
\usepackage{amsmath}
\usepackage{tabularx} 
\usepackage{amssymb}
\usepackage{float}
\usepackage{caption}
\usepackage{titletoc}
\usepackage{xcolor}
\usepackage{colortbl}
\usepackage{array}
\usepackage{tcolorbox}

\definecolor{inputgray}{gray}{0.93}
\definecolor{chunkblue}{RGB}{220,230,245}
\definecolor{chunkgreen}{RGB}{225,245,225}
\definecolor{chunkyellow}{RGB}{250,245,210}
\definecolor{finalanswer}{RGB}{245,225,225}

\newtcolorbox{inputbox}{
  colback=inputgray,
  colframe=black,
  boxrule=0.4pt,
  arc=2pt,
  left=6pt,
  right=6pt,
  top=4pt,
  bottom=4pt
}

\newtcolorbox{bluebox}{
  colback=chunkblue,
  colframe=black,
  boxrule=0.4pt,
  arc=2pt,
  left=6pt,
  right=6pt,
  top=4pt,
  bottom=4pt
}

\newtcolorbox{greenbox}{
  colback=chunkgreen,
  colframe=black,
  boxrule=0.4pt,
  arc=2pt,
  left=6pt,
  right=6pt,
  top=4pt,
  bottom=4pt
}

\newtcolorbox{yellowbox}{
  colback=chunkyellow,
  colframe=black,
  boxrule=0.4pt,
  arc=2pt,
  left=6pt,
  right=6pt,
  top=4pt,
  bottom=4pt
}

\newtcolorbox{answerbox}{
  colback=finalanswer,
  colframe=black,
  boxrule=0.4pt,
  arc=2pt,
  left=6pt,
  right=6pt,
  top=4pt,
  bottom=4pt
}

\usepackage{amssymb}  
\usepackage{pifont}   
\usepackage{longtable,booktabs}
\usepackage{caption}
\DeclareCaptionFont{ninept}{\fontsize{9}{10.8}\selectfont}
\usepackage{tabularx} 
\newcolumntype{L}{>{\raggedright\arraybackslash}X}

\usepackage{url}
\usepackage[most]{tcolorbox} 

\tcbset{
  colback=gray!5!white,
  colframe=black,
  boxrule=0.5pt,
  arc=2mm,
  left=2mm,
  right=2mm,
  top=1mm,
  bottom=1mm,
}

\makeatletter
\DeclareRobustCommand{\breakabletexttt}[1]{%
  \begingroup
  \ttfamily
  \hyphenchar\font=`\- 
  \frenchspacing
  \sloppy
  \catcode`_=12 
  \catcode`-=12 
  \catcode`\.=12 
  \catcode`/=12 
  \@vobeyspaces
  \seqsplit{#1}%
  \endgroup
}
\makeatother

\usepackage{seqsplit} 

\title{How Reliable are Confidence Estimators for Large Reasoning Models?\\A Systematic Benchmark on High-Stakes Domains}

\author{
  \textbf{Reza Khanmohammadi\textsuperscript{\textnormal{1}}\thanks{Corresponding author: \href{mailto:khanreza@msu.edu}{\texttt{khanreza@msu.edu}}}}\textnormal{,}
  \textbf{Erfan Miahi\textsuperscript{\textnormal{2}}}\textnormal{,}
  \textbf{Simerjot Kaur\textsuperscript{\textnormal{3}}}\textnormal{,}
\\
  \textbf{Charese H. Smiley\textsuperscript{\textnormal{3}}}\textnormal{,}
  \textbf{Ivan Brugere\textsuperscript{\textnormal{3}}}\textnormal{,}
  \textbf{Kundan Thind\textsuperscript{\textnormal{4,\textdagger}}}\textnormal{,}
  \textbf{Mohammad M. Ghassemi\textsuperscript{\textnormal{1},}\thanks{Shared senior authorship}}
\\
\\
  \textsuperscript{1}Michigan State University \quad
  \textsuperscript{2}Independent AI Researcher \\
  \textsuperscript{3}JPMorgan AI Research \quad
  \textsuperscript{4}Henry Ford Health
\\
      \normalsize\textsuperscript{1} \texttt{\{khanreza, ghassem3\}@msu.edu} \quad \textsuperscript{2} \texttt{mhi.erfan1@gmail.com} \\
      \normalsize\textsuperscript{3} \texttt{\{simerjot.kaur, ivan.brugere, charese.h.smiley\}@jpmchase.com} \quad \textsuperscript{4} \texttt{kthind1@hfhs.org}
}

\begin{document}
\maketitle
\begin{abstract}
\input{sections/abstract}
\end{abstract}
{\normalsize \textbf{Code} --- {\small \href{https://github.com/Ledengary/RMCB}{https://github.com/ledengary/RMCB}}} \\
{\normalsize \textbf{Data} --- {\small \href{https://huggingface.co/datasets/ledengary/RMCB}{https://huggingface.co/datasets/ledengary/RMCB}}}

\section{Introduction}
\input{sections/introduction}
\section{Related Work}
\label{sec:related_work}
\input{sections/related_work}
\section{The RMCB Benchmark}
\label{sec:RMCB_benchmark}
\input{sections/RMCB_benchmark}
\section{Experimental Methodology}
\label{sec:methodology}
\input{sections/experimental_methodology}
\section{Results}
\label{sec:results}
\input{sections/results}
\section{Discussion}
\label{sec:discussion}
\input{sections/discussion}
\section{Conclusion}
\label{sec:conclusion}
\input{sections/conclusion}
\section*{Limitations}
\label{sec:limitations}
\input{sections/limitations}
\section*{Ethical Considerations}
\label{sec:ethical_considerations}
\input{sections/ethical_considerations}

\section*{Acknowledgments}
This work was supported in part by the Henry Ford Health + Michigan State University Health Sciences Cancer Seed Funding Program and by the JPMorgan Chase AI Research Faculty Research Award. The authors are solely responsible for the contents of this paper; the opinions expressed do not necessarily reflect those of the funding organizations. The authors also acknowledge the use of Large Language Models to assist in polishing the language and grammar of this manuscript.

\section*{Disclaimer}
This paper was prepared for informational purposes by the Artificial Intelligence Research group of JPMorgan Chase \& Co and its affiliates (“JP Morgan”), and is not a product of the Research Department of JP Morgan. JP Morgan makes no representation and warranty whatsoever and disclaims all liability, for the completeness, accuracy or reliability of the information contained herein. This document is not intended as investment research or investment advice, or a recommendation, offer or solicitation for the purchase or sale of any security, financial instrument, financial product or service, or to be used in any way for evaluating the merits of participating in any transaction, and shall not constitute a solicitation under any jurisdiction or to any person, if such solicitation under such jurisdiction or to such person would be unlawful.



\bibliography{custom}

\clearpage
\appendix
\begin{center}
\huge\textbf{Appendix}
\end{center}
\label{sec:appendix}

\begin{center}
\Large\textbf{Table of Contents}
\rule{\linewidth}{1pt}
\end{center}
\startcontents[sections]
\printcontents[sections]{}{1}{}

\rule{\linewidth}{1pt}

\input{sections/appendix_llm}
\input{sections/appendix_dataset}
\input{sections/appendix_baseline_methods}
\input{sections/appendix_evaluation_metrics}
\input{sections/appendix_optuna}
\input{sections/appendix_extended_results}
\input{sections/appendix_ellipse_plots}
\input{sections/appendix_calibration_diagrams}

\end{document}

%% file: sections/abstract.tex
The miscalibration of Large Reasoning Models (LRMs) undermines their reliability in high-stakes domains, necessitating methods to accurately estimate the confidence of their long-form, multi-step outputs. To address this gap, we introduce the Reasoning Model Confidence estimation Benchmark (RMCB), a public resource of 347,496 reasoning traces from six popular LRMs across different architectural families. The benchmark is constructed from a diverse suite of datasets spanning high-stakes domains, including clinical, financial, legal, and mathematical reasoning, alongside complex general reasoning benchmarks, with correctness annotations provided for all samples. Using RMCB, we conduct a large-scale empirical evaluation of over ten distinct representation-based methods, spanning sequential, graph-based, and text-based architectures. Our central finding is a persistent trade-off between discrimination (AUROC) and calibration (ECE): text-based encoders achieve the best AUROC (0.672), while structurally-aware models yield the best ECE (0.148), with no single method dominating both. Furthermore, we find that increased architectural complexity does not reliably outperform simpler sequential baselines, suggesting a performance ceiling for methods relying solely on chunk-level hidden states. This work provides the most comprehensive benchmark for this task to date, establishing rigorous baselines and demonstrating the limitations of current representation-based paradigms.

%% file: sections/introduction.tex
Despite impressive performance, large language models (LLMs) often struggle with confidence calibration. The model's estimated probability that an answer is correct frequently misaligns with the actual outcome, leading to high confidence in wrong answers and low confidence in right ones. This unreliability forces costly manual review of every output, which undermines the primary benefits of automation. The problem is especially critical in high-stakes domains like medicine, finance, and law, where a single miscalibrated output can have significant consequences. Therefore, accurate confidence scores are essential to build trustworthy systems, allowing users to efficiently determine which outputs require human verification. A reliable confidence score must be both well-calibrated, ensuring the predicted probability aligns with the true likelihood of correctness, and discriminative, effectively distinguishing correct answers from incorrect ones.

This challenge is particularly pronounced for Large Reasoning Models (LRMs), which articulate their problem-solving process through a sequence of intermediate thoughts before providing a final answer. While these models often achieve superior performance, their long and detailed outputs substantially increase the cost and complexity of manual verification. Current confidence estimation techniques, which primarily analyze logits or probe hidden states, were developed for short-form, token-level prediction and are not designed for these long-form outputs. The confidence in a multi-step argument is an emergent property of the entire reasoning trajectory, not a static feature of any single component. Our comprehensive benchmark confirms this mismatch, revealing that existing methods consistently fail to achieve both high discriminative power and good calibration for LRMs. Consequently, a significant gap remains for a generalizable framework that can reliably assess the confidence of multi-step reasoning across diverse model architectures and high-stakes domains.

In this work, we make two primary contributions. First, we introduce the \textbf{R}easoning \textbf{M}odel \textbf{C}onfidence estimation \textbf{B}enchmark (\textbf{RMCB}), the first large-scale, publicly available resource dedicated to this task. The benchmark comprises 347,496 reasoning traces generated by six popular LRMs. Whereas prior studies have often focused on a narrow set of domains like programming or mathematics, or drawn from specific tasks within broader evaluations, RMCB is constructed from a diverse suite of full datasets spanning high-stakes domains, including clinical, financial, and legal reasoning, alongside general knowledge understanding. A core component of RMCB is its detailed annotation; we release all model inferences with corresponding correctness labels covering every evaluated response.

Building on this resource, we conduct a comprehensive empirical study evaluating over ten distinct confidence estimation methods. Our investigation covers the full spectrum of representation-based techniques, from baseline approaches like verbalized confidence and text classification with specialized encoders, to more complex models that operate on sequences of hidden states, including stacked sequential models and various graph-based architectures. This large-scale evaluation reveals our second primary contribution: the empirical finding that methods relying on hidden-state representations, regardless of their architectural complexity, hit a consistent performance ceiling and face a persistent trade-off between discriminative power (AUROC) and calibration (ECE). We show that even sophisticated models designed to capture the relational structure of the reasoning trace fail to consistently outperform simpler baselines. Ultimately, this work rigorously quantifies the limitations of the current paradigm and establishes a clear set of baselines that highlight the need for future research to explore alternative signal sources beyond static hidden states.


%% file: sections/related_work.tex
    The estimation of an LLM's confidence is a well-established field of study, with methods generally falling into several distinct categories. One of the most direct approaches is \textit{verbalized confidence}, where a model is explicitly prompted to state its own certainty level in natural language, which is then parsed into a numerical score \citep{verbalizedconfidenceelicitation}. Another common black-box technique is \textit{self-consistency}, which assesses the robustness of an answer by generating multiple responses with sampling and measuring the consensus among them \citep{selfconsistency}. Moving to white-box methods, a significant body of work focuses on \textit{generative signals} derived from the model's output logits. These techniques analyze the properties of the token-level probability distributions, using features like the log-probability of the generated sequence \cite{HADEMIF} or applying post-hoc calibration methods such as temperature scaling \citep{tempscaling}. Other white-box approaches operate on the model's \textit{internal hidden states}, training lightweight classifiers on hidden-state representations to predict correctness. This includes methods like P(IK), which uses the hidden state of the input prompt \citep{pik}, SAPLMA, which identifies particular hidden layers whose activations best capture correctness signals \citep{SAPLMA}, and InternalInspector, which leverages the full set of hidden states across all layers for confidence estimation \citep{i2inspector}. A more recent and distinct category involves probing the model's \textit{internal stability}. For instance, CCPS \citep{CCPS} introduces targeted perturbations to a model's final hidden states and measures the resulting representational shift, using the magnitude of this shift as a proxy for the model's underlying confidence.

However, applying these established methods to LRMs presents a series of fundamental challenges. The long-form, multi-step nature of LRM outputs makes many of these techniques computationally intractable or conceptually misaligned. Self-consistency, for example, becomes prohibitively expensive due to the high cost of generating multiple, thousand-token reasoning traces for a single query. Verbalized confidence may be unreliable, as the step-by-step reasoning process can lead a model to become overconfident in its final conclusion, even when the underlying logic is flawed. Methods relying on token-level signals, such as those analyzing generative log-probabilities or probing internal stability like CCPS, struggle with extreme representational scale—storing and processing full logit vectors or computing perturbations across thousands of tokens per sample is not a feasible strategy for producing a single, coherent representation.

This infeasibility of token-level analysis for long-form reasoning has naturally led to a focus on higher-level representations, with two notable approaches shaping the initial exploration of this problem. The first, from \citet{YVCE} (YVCE), adapts the verbalized confidence approach for LRMs. The second, and more foundational for representation-based methods, is Probing Hidden States for Self-Verification (PHSV) by \citet{PHSV}. PHSV trains a lightweight MLP on the hidden state of each intermediate reasoning “chunk,” defined as a contiguous segment of the reasoning trace that ends in an intermediate answer. At inference time, the classifier is applied sequentially over chunks, and the confidence score is determined by the first chunk exceeding a threshold, or by the final chunk if none do.

While these methods provide a crucial starting point, the field of LRM confidence estimation remains underexplored. The initial works were largely evaluated in isolation, without a comprehensive comparison against a broader set of architectural alternatives. Furthermore, their evaluations have often focused on a limited set of domains, such as mathematics or general reasoning, leaving their generalizability to diverse, high-stakes applications an open question. This work is motivated by three critical gaps in the literature: (1) the lack of a large-scale, public benchmark for LRM confidence estimation that spans multiple high-stakes domains; (2) the absence of a systematic comparison between the foundational chunk-level methods and more complex sequential, graph-based, and text-based architectures; and (3) an incomplete understanding of the trade-offs between discrimination and calibration for these representation-based methods. We aim to fill these gaps by providing the first comprehensive benchmark that directly addresses these challenges.

%% file: sections/RMCB_benchmark.tex
To address the gap in LRM confidence estimation, we constructed the RMCB benchmark, a large-scale, publicly available resource designed to facilitate the systematic evaluation of confidence estimation methods. This section details the data sources, model suite, and annotation methodology used to build the benchmark.
\begin{table}[h]
\centering
\caption{Distribution of datasets used for the RMCB training and evaluation splits. The test set is entirely disjoint from the training set. Domains indicate the primary reasoning type assessed.}
\label{tab:dataset_splits}
\begingroup
\footnotesize                           

\begin{tabular}{@{}l l l r@{}}
\toprule
\textbf{Dataset} & \textbf{Domain} & \textbf{Split} & \textbf{Samples} \\
\midrule
\multicolumn{4}{c}{\textit{Training Datasets (Total = 10,000)}} \\
\midrule
GSM8K & Mathematical & Train & 1,000 \\
TAT-QA & Financial & Train & 1,000 \\
MedQA & Medical & Train & 1,000 \\
LEXam & Legal & Train & 1,000 \\
ARC & General & Train & 1,000 \\
CommonsenseQA2 & General & Train & 1,000 \\
LogiQA & General & Train & 1,000 \\
OpenBookQA & General & Train & 1,000 \\
QuaRTz & General & Train & 1,000 \\
ReClor & General & Train & 1,000 \\
\midrule
\multicolumn{4}{c}{\textit{Evaluation Datasets (Total = 51,951)}} \\
\midrule
MATH & Mathematical & Test & 5,000 \\
FinQA & Financial & Test & 1,138 \\
MedMCQA & Medical & Test & 6,150 \\
LegalBench & Legal & Test & 21,167 \\
MMLU-Pro & General & Test & 11,987 \\
BBH & General & Test & 6,509 \\
\midrule
\multicolumn{4}{c}{\textbf{Grand Total = 61,951}} \\
\bottomrule
\end{tabular}
\endgroup
\vspace{-10pt}
\end{table}

\subsection{Data Sources \& LRM Suite}
The foundation of RMCB is a diverse collection of reasoning problems sourced from well-established public benchmarks. Whereas prior studies have often focused on a narrow set of domains or specific tasks within broader evaluations, our benchmark is constructed from full datasets spanning multiple high-stakes domains—specifically clinical, financial, and legal reasoning—alongside challenging mathematical and general commonsense reasoning tasks. As shown in Table~\ref{tab:dataset_splits}, the benchmark is split into two entirely disjoint sets: a balanced training set of 10,000 samples and a challenging evaluation set of over 50,000 samples. A full breakdown of these datasets, including our standardized curation process, is provided in Appendix~\ref{app:appendix_dataset}. A complete list of dataset sources, access methods, and exact versions used in this benchmark is provided in Appendix~\ref{app:dataset_sources}.

To generate the reasoning traces for our benchmark, we selected a suite of six popular and powerful open-weight LRMs. These models represent diverse architectural families (Phi, Qwen, Mistral, EXAONE) and a wide range of parameter counts (from ~4B to ~33B), establishing a broad and representative sample of modern reasoning capabilities. Specifically, our evaluation includes \texttt{Phi-4-mini-flash-reasoning} (3.85B parameters), \texttt{Qwen3-8B} (8.19B parameters), \texttt{Qwen3-14B} (14.8B parameters), \texttt{Magistral-Small-2506} (23.6B parameters), \texttt{EXAONE-Deep-32B} (32.0B parameters), and \texttt{QwQ-32B} (32.8B parameters). Detailed architectural parameters for each LRM are provided in Appendix~\ref{app:appendix_llm}.

\subsection{Data Generation \& Annotation}
We generated responses from each of the six LRMs for every prompt, resulting in a total corpus of 347,496 unique reasoning traces. This count reflects the valid, high-quality generations remaining after filtering. During large-scale inference, a small fraction of outputs (approximately 6.5\%) were excluded to ensure benchmark integrity. These exclusions were primarily due to generation failures such as empty outputs or repetition loops. As a result, the number of successfully annotated samples varies slightly across LRMs, as detailed in Table ~\ref{tab:test_merged_llm} (e.g., 44,409 test samples for \texttt{Phi-4-mini-flash-reasoning} versus 50,792 for \texttt{QwQ-32B}). We prioritize the validity and quality of reasoning traces over enforcing a uniform sample count across architectures. All remaining outputs were generated using deterministic decoding (temperature of 0.0) with a maximum length of 4096 tokens to ensure reproducibility; further details on the generation setup are available in Appendix~\ref{app:response_generation}. To enable a granular, step-by-step analysis, each generated response was first segmented into coherent units of thought, or ``chunks.'' Our process is an enhancement of the methodology first introduced by \citet{PHSV}, which groups paragraphs based on a set of linguistic keywords that signal a shift in the reasoning process (e.g., self-correction, verification). Further details on our segmentation methodology and the expanded keyword list can be found in Appendix~\ref{app:response_segmentation}. An example reasoning trace illustrating the chunking process is provided in Appendix~\ref{app:example_trace}.

A core contribution of RMCB is its detailed annotation of correctness. Each of the 347,496 reasoning traces was systematically graded to ensure comprehensive evaluation across all responses. For overall correctness, multiple-choice answers were checked via string matching, while open-ended responses were evaluated for semantic equivalence by a state-of-the-art LLM judge (\texttt{GPT-5-nano}), a practice validated by recent literature \citep{calibration-tuning, PHSV, emit, CCPS}. To provide a more granular supervisory signal, we extended this grading to the per-chunk level, applying the same LLM-based judging process to each individual reasoning chunk. This allows us to obtain a correctness label for each intermediate step in the reasoning process. Full details on the prompts and methodology for this automated grading are provided in Appendix~\ref{app:response_grading}. The final sample- and chunk-level statistics after this comprehensive annotation process are available in Appendix Tables~\ref{tab:train_merged_llm} and \ref{tab:test_merged_llm}.

\subsection{Benchmark Components}
We publicly release all components of the RMCB to accelerate research on reasoning confidence estimation. The released artifacts include: (a) a comprehensive dataset of 347{,}496 reasoning traces, including input prompts and model-generated responses from six LRMs; (b) complete correctness annotations provided for every trace; and (c) the full implementation of all evaluated confidence estimation methods, enabling reproducible and extensible experimentation.

%% file: sections/experimental_methodology.tex
To systematically evaluate confidence estimation methods on our RMCB benchmark, we designed a rigorous experimental framework. All LRMs are kept frozen throughout our experiments; no fine-tuning is performed on the underlying reasoning models. All learning occurs exclusively in the downstream, post-hoc confidence estimators. This section details the task formulation, the metrics used for evaluation, the suite of methods investigated, and the consistent protocol for training and hyperparameter optimization.

\subsection{Task Formulation}
We formulate LRM confidence estimation as a binary classification problem. For a given reasoning trace generated by an LRM, the task is to produce a single scalar probability score $p \in [0,1]$ that predicts whether the final answer is correct (label~1) or incorrect (label~0). This unified formulation enables direct comparison across all methods.

\subsection{Evaluation Metrics}
A reliable confidence score must satisfy two distinct criteria: it must be discriminative (able to distinguish correct from incorrect answers) and well-calibrated (its predicted probability must align with the true likelihood of correctness). To capture this fundamental trade-off, we focus on two primary metrics:

\noindent \textbf{Area Under the ROC Curve (AUROC):} \ Measures the model's discriminative power by evaluating its ability to rank correct answers higher than incorrect ones across all thresholds. A score of 1.0 represents perfect discrimination.

\noindent \textbf{Expected Calibration Error (ECE):} \ Measures the alignment between a model's predicted confidence and its empirical accuracy. A lower ECE indicates better calibration, with 0 being a perfectly calibrated model.

Our analysis centers on the inherent tension between maximizing AUROC and minimizing ECE, a core challenge in confidence estimation. We also report a suite of supplementary metrics, including the Brier Score and F1 Score, to provide a more complete picture of performance. Detailed definitions for all metrics are provided in Appendix~\ref{app:evaluation_metrics}.

\subsection{Confidence Estimation Methods}
Our investigation covers a wide spectrum of over fifteen distinct representation-based techniques, which we group into logical families to systematically explore different architectural hypotheses. These methods operate on one of two primary signal sources: the LRM's raw chunk hidden states, or correctness-focused features derived from a pre-trained probe (\texttt{PHSV-half}). We began by implementing established \textbf{baseline methods}, including verbalized confidence (\texttt{YVCE}), probing the initial prompt state (\texttt{P(IK)}), and the state-of-the-art chunk-level probing method (\texttt{PHSV}). To move beyond single-chunk analysis, we developed \textbf{Stacked Final Hidden State (\texttt{SFHS})} models that process the entire stack of chunk hidden states using various backbones. In addition, we introduce \textbf{Token-Level Correctness Classification (\texttt{TLCC})}, a sequential model that replaces chunk hidden states with low-dimensional statistics derived solely from token-level logit distributions, enabling a controlled evaluation of generative uncertainty signals without access to internal representations. To explicitly model the non-linear structure of reasoning, we benchmarked several \textbf{graph-based architectures}, including models that learn from the simple chronological flow (\texttt{GNN-SB}), the global logical and semantic coherence using rich edge features (\texttt{GNN-SR}), and the meta-level dynamics of the confidence trajectory itself (\texttt{GNN-CD}). Finally, we evaluated \textbf{hybrid and text-based models} that fuse different signal types (\texttt{LateFusion} and \texttt{CE}) or operate on the raw text of the entire reasoning trace (\texttt{ETTIN}). A comprehensive description of each method's architecture and implementation details is provided in Appendix~\ref{app:confidence_estimation_methods}.

\subsection{Training \& Hyperparameter Optimization}
To ensure a fair and rigorous comparison, all trainable models were developed using a consistent training and optimization protocol. Hyperparameter tuning was performed for each method using Optuna \citep{optuna}, with each study running for up to 100 trials. For each LLM, we allocated $20\%$ of the training portion as a validation subset, stratified by dataset to preserve the domain distribution within each LLM’s training pool (see Tables~\ref{tab:train_merged_llm} and~\ref{tab:test_merged_llm} for the full train and test sizes).

The selection of an appropriate objective function for optimization was critical. Given the class imbalance inherent in our data—where a majority of LRM responses are correct—simply optimizing for accuracy would be a misleading metric. Our preliminary experiments also revealed that optimizing solely for discriminative power (AUROC) often produced models with very poor calibration, sacrificing ECE to maximize ranking performance. As our goal is to evaluate methods on their ability to perform well on \textit{both} axes, we designed a composite score to guide the optimization process. The objective function was a composite score designed to jointly address discrimination and calibration:
\[
\text{CompositeScore} = \alpha \cdot \text{AUROC} + (1 - \alpha) \cdot (1 - \text{ECE})
\]
We set $\alpha = 0.6$ to place a slight emphasis on AUROC, prioritizing models with strong discriminative power while still imposing a significant penalty for poor calibration.

For final model selection, we imposed an additional practical constraint: a trial was only considered "feasible" if its best-performing epoch also achieved a minimum sensitivity and specificity of 0.50 at its Youden's J optimal threshold. This ensures our final reported models demonstrate a tangible predictive ability better than random chance on both positive and negative classes. Among all feasible trials, the one with the highest composite score was selected. To ensure a fair comparison of \textit{architectural efficiency}, all model configurations were constrained to a maximum of 3.2 million trainable parameters. Each trial was trained for up to 200 epochs with an early-stopping patience of 20 epochs. The full details of our optimization strategy and the hyperparameter search space for each method are provided in Appendix~\ref{app:training_details}.

%% file: sections/results.tex
\begingroup
\scriptsize
\setlength{\tabcolsep}{3pt}
\begin{table*}[t]
\centering
\caption{Overall performance metrics for each method aggregated across all LLMs and test datasets. Each metric value represents a double-averaged result with standard deviation: first, each method's performance is averaged across all datasets for each LLM (unweighted mean $\pm$ std), then these LLM-specific means are averaged across all LLMs (unweighted mean $\pm$ std). \textbf{Bold} values indicate the overall best-performing method for each metric. The full table with detailed values is provided in Appendix Table~\ref{tab:method_only_appendix}.}
\label{tab:method_only_main}
\resizebox{0.9\textwidth}{!}{%
\scriptsize
\begin{tabular}{@{}l c c c c c c c@{}}
\toprule
\textbf{Method} & \textbf{ECE} $\downarrow$ & \textbf{Brier} $\downarrow$ & \textbf{Acc} $\uparrow$ & \textbf{F1} $\uparrow$ & \textbf{Spec} $\uparrow$ & \textbf{AUCPR} $\uparrow$ & \textbf{AUROC} $\uparrow$\\
\midrule
YVCE & 0.279 {\tiny $\pm$0.14} & 0.307 {\tiny $\pm$0.12} & 0.586 {\tiny $\pm$0.14} & 0.671 {\tiny $\pm$0.15} & 0.143 {\tiny $\pm$0.15} & \textbf{0.696} {\tiny $\pm$0.10} & 0.603 {\tiny $\pm$0.06} \\
TLCC-CONV & 0.178 {\tiny $\pm$0.03} & 0.222 {\tiny $\pm$0.02} & 0.665 {\tiny $\pm$0.03} & 0.669 {\tiny $\pm$0.09} & 0.424 {\tiny $\pm$0.14} & 0.655 {\tiny $\pm$0.10} & 0.639 {\tiny $\pm$0.04} \\
PHSV & 0.197 {\tiny $\pm$0.05} & 0.251 {\tiny $\pm$0.03} & 0.609 {\tiny $\pm$0.05} & 0.600 {\tiny $\pm$0.12} & 0.389 {\tiny $\pm$0.20} & 0.577 {\tiny $\pm$0.11} & 0.598 {\tiny $\pm$0.04} \\
SFHS-Conv & 0.165 {\tiny $\pm$0.02} & 0.222 {\tiny $\pm$0.01} & 0.676 {\tiny $\pm$0.02} & 0.656 {\tiny $\pm$0.10} & 0.453 {\tiny $\pm$0.14} & 0.659 {\tiny $\pm$0.10} & 0.653 {\tiny $\pm$0.02} \\
GNN-SB-GCN & 0.150 {\tiny $\pm$0.02} & 0.216 {\tiny $\pm$0.01} & 0.678 {\tiny $\pm$0.02} & 0.645 {\tiny $\pm$0.10} & 0.432 {\tiny $\pm$0.14} & 0.657 {\tiny $\pm$0.10} & 0.653 {\tiny $\pm$0.03} \\
GNN-SB-GraphSAGE & 0.154 {\tiny $\pm$0.02} & 0.217 {\tiny $\pm$0.01} & 0.675 {\tiny $\pm$0.02} & 0.643 {\tiny $\pm$0.09} & 0.446 {\tiny $\pm$0.15} & 0.664 {\tiny $\pm$0.09} & 0.659 {\tiny $\pm$0.03} \\
GNN-SR-Transformer & 0.175 {\tiny $\pm$0.02} & \textbf{0.208} {\tiny $\pm$0.01} & 0.677 {\tiny $\pm$0.02} & 0.648 {\tiny $\pm$0.12} & 0.432 {\tiny $\pm$0.19} & 0.664 {\tiny $\pm$0.09} & 0.656 {\tiny $\pm$0.03} \\
GNN-CD-noft-GCN2Conv-dual & 0.174 {\tiny $\pm$0.04} & 0.223 {\tiny $\pm$0.02} & 0.663 {\tiny $\pm$0.03} & 0.611 {\tiny $\pm$0.10} & \textbf{0.523} {\tiny $\pm$0.14} & 0.657 {\tiny $\pm$0.11} & 0.651 {\tiny $\pm$0.02} \\
ETTIN & 0.160 {\tiny $\pm$0.01} & 0.217 {\tiny $\pm$0.01} & 0.677 {\tiny $\pm$0.02} & \textbf{0.698} {\tiny $\pm$0.10} & 0.290 {\tiny $\pm$0.22} & 0.680 {\tiny $\pm$0.09} & \textbf{0.672} {\tiny $\pm$0.04} \\
ETTIN-HGA & \textbf{0.148} {\tiny $\pm$0.02} & 0.211 {\tiny $\pm$0.01} & \textbf{0.689} {\tiny $\pm$0.01} & 0.693 {\tiny $\pm$0.12} & 0.282 {\tiny $\pm$0.23} & 0.678 {\tiny $\pm$0.10} & 0.670 {\tiny $\pm$0.03} \\
\bottomrule
\end{tabular}}
\end{table*}
\vspace{-10pt}
\endgroup 

\begin{figure*}[t]
\centering
\makebox[\textwidth][c]{\includegraphics[width=0.9\textwidth]{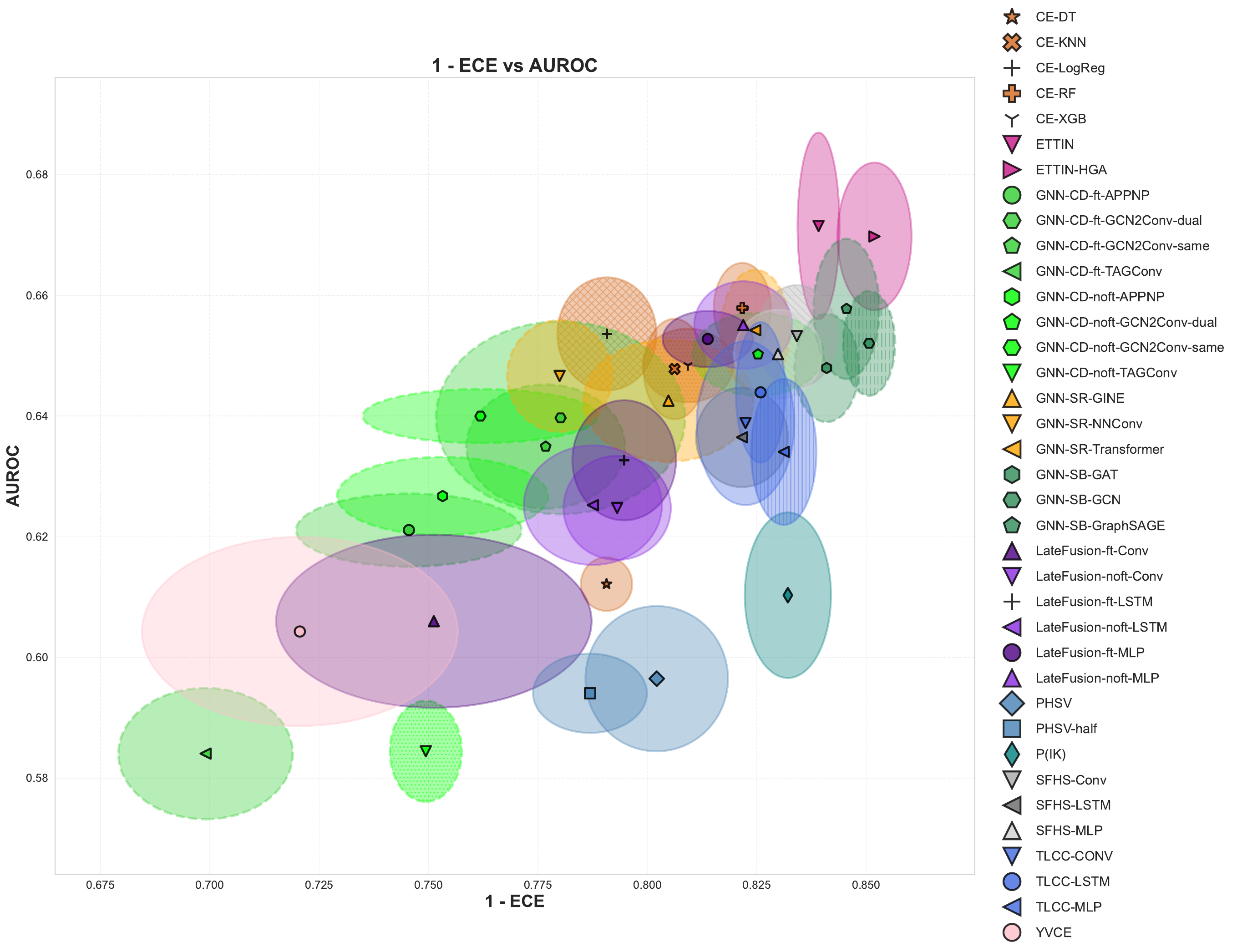}}
\caption{Performance trade-off between calibration (1-ECE) and discrimination (AUROC). Each ellipse denotes a method, with its center showing the unweighted doubly-averaged mean performance (first across datasets per LRM, then across LRMs). Ellipse width and height represent the standard deviation of these LRM-specific means, reflecting consistency across model architectures. The top-right corner marks ideal performance.}
\label{fig:ECE_vs_AUROC_ellipse_distribution}
\vspace{-10pt}
\end{figure*}

\begin{figure*}[t]
\centering
\makebox[\textwidth][c]{\includegraphics[width=0.98\textwidth]{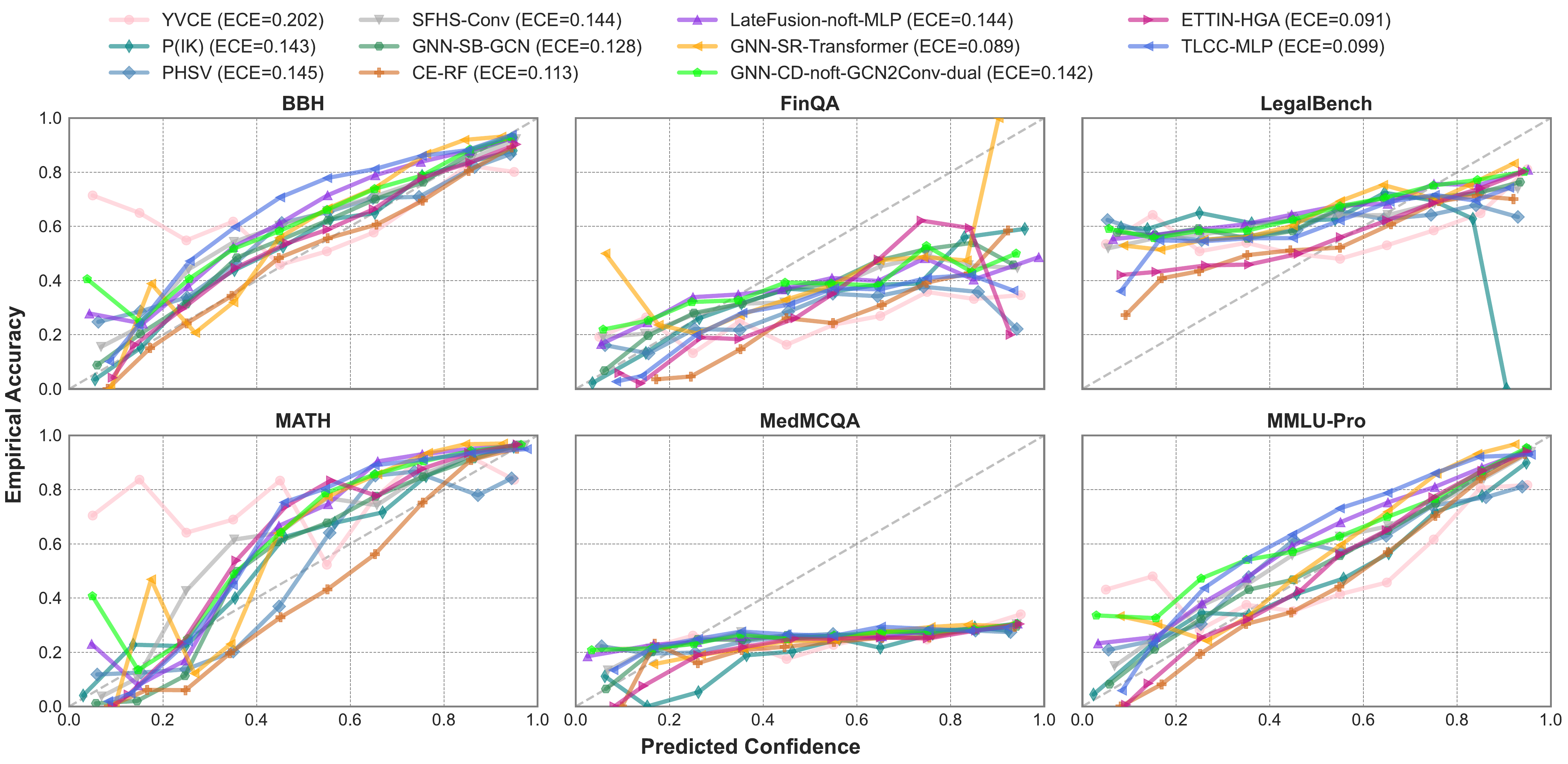}}
\caption{Calibration curves for each test dataset, aggregated across all LRMs. Each subplot shows one dataset's reliability diagram with methods averaged across all six LRM families. The ECE values in the legend represent each method's average performance across LRMs for that specific dataset. Points closer to the diagonal (dashed line) indicate better calibration.}
\label{fig:calibration_grid_v2_dataset_cells_best_10bins_2x3}
\vspace{-10pt}
\end{figure*}

We evaluated over ten distinct confidence estimation methods across six LRM families and six challenging test datasets. The overall performance, aggregated across all models and datasets, is presented in Table~\ref{tab:method_only_main}. While this table provides the highest-level summary of our findings, a comprehensive set of detailed results—including performance breakdowns per-LRM, per-dataset, and combinations thereof—is provided in Appendix~\ref{app:comprehensive_tables} to facilitate a deeper, more granular analysis. This section reports the key empirical findings from our benchmark.

\noindent\textbf{Discrimination Performance.}\quad
The primary metrics for discrimination are AUROC, AUCPR, and F1 score. Across these, the text-based encoder methods, \texttt{ETTIN} and \texttt{ETTIN-HGA}, emerge as the top performers. \texttt{ETTIN} achieves the highest overall AUROC (0.672) and F1 Score (0.698), with \texttt{ETTIN-HGA} following closely (AUROC 0.670, F1 0.693). The best-performing hidden-state-based methods, such as \texttt{GNN-SB-GraphSAGE} (AUROC 0.659) and \texttt{GNN-SR-Transformer} (AUROC 0.656), form a competitive second tier but do not surpass the text-based approaches. In contrast, foundational baselines like \texttt{PHSV} (AUROC 0.598) and prompting-based methods like \texttt{YVCE} (AUROC 0.603) show significantly weaker discriminative ability. A surprising result is the strong performance of \texttt{YVCE} on AUCPR (0.696), suggesting that while poorly calibrated, it is effective at identifying some high-confidence correct answers.

\noindent\textbf{Calibration Performance.}\quad
The primary metrics for calibration are ECE and Brier score. Here, a different set of models excel. The best overall calibration is achieved by \texttt{ETTIN-HGA}, with an ECE of 0.148. This is closely followed by the simple graph models, specifically \texttt{GNN-SB-GCN} (0.150) and \texttt{GNN-SB-GraphSAGE} (0.154). Notably, the baseline \texttt{ETTIN} model, despite its top discriminative performance, has a comparatively weaker ECE of 0.160. The best Brier score, which balances discrimination and calibration, is achieved by \texttt{GNN-SR-Transformer} (0.208). The poorest calibration is observed in the \texttt{YVCE} baseline (ECE 0.279). A key observation across all methods is that no single model is dominant, with top performers in discrimination often being distinct from top performers in calibration.

\noindent\textbf{Visualizing the Performance Landscape.}\quad
Figure~\ref{fig:ECE_vs_AUROC_ellipse_distribution} provides a visual summary of these results, plotting each method's calibration (1-ECE) against its discrimination (AUROC). The center of each ellipse marks the method's mean performance, while its height and width represent the standard deviation of AUROC and ECE, respectively, indicating consistency across the benchmark. The plot shows that no single method occupies the ideal top-right corner. The text-based methods \texttt{ETTIN} and \texttt{ETTIN-HGA} are positioned furthest to the right, indicating the highest mean AUROC. The \texttt{GNN-SB} family models are clustered towards the top of the plot, indicating strong mean calibration. Many methods, such as the \texttt{SFHS} family and \texttt{GNN-SR-Transformer}, occupy a dense cluster in the center, representing a balance between the two metrics. Supplementary plots for other metric pairs are provided in Appendix~\ref{app:allipse_plots}.

\noindent\textbf{Visualizing Calibration Across LRMs.}\quad
To examine how calibration varies across reasoning domains, Figure~\ref{fig:calibration_grid_v2_dataset_cells_best_10bins_2x3} presents reliability diagrams for each test dataset, with methods aggregated across all six LRM families. The ECE values in the legend represent dataset-specific averages across LRMs. Each subplot plots predicted confidence against empirical accuracy, with the dashed diagonal representing perfect calibration. The plots reveal substantial variation in calibration quality across both methods and datasets. Methods like \texttt{ETTIN-HGA} (avg. ECE=0.091) and \texttt{GNN-SR-Transformer} (avg. ECE=0.089) maintain consistently strong calibration with curves close to the diagonal across most datasets. Conversely, \texttt{YVCE} (avg. ECE=0.202) exhibits the poorest overall calibration. Notably, MedMCQA proves exceptionally challenging, with nearly all methods producing flat, poorly-aligned curves indicating severe miscalibration, while MMLU-Pro and BBH demonstrate better calibration potential with curves more closely following the diagonal. FinQA and MATH show moderate calibration with substantial method-dependent variation. For per-LRM calibration analysis and further interpretation details, see Appendix~\ref{app:appendix_calibration_diagrams}.

%% file: sections/discussion.tex
Our comprehensive benchmark reveals several key insights into the challenges of representation-based confidence estimation for LRMs. We discuss the main findings below.

\noindent \textbf{A persistent trade-off exists between discrimination and calibration.} \quad
The central finding of our study is a consistent trade-off between a model's ability to distinguish correct from incorrect answers (AUROC) and its ability to produce calibrated scores (ECE). This is evident in Table~\ref{tab:method_only_main}, where the top-performing model for AUROC, \texttt{ETTIN}, is not the best for ECE. This trade-off is visually confirmed in Figure~\ref{fig:ECE_vs_AUROC_ellipse_distribution}, where no single method occupies the ideal top-right corner. Instead, top-performing methods are scattered along a performance frontier, suggesting a fundamental tension: architectures that capture holistic features for discrimination may be less effective at modeling the structural nuances crucial for calibration.

\noindent \textbf{Text-based encoders excel at discrimination but require structural awareness for calibration.} \quad
The strong performance of the \texttt{ETTIN} model on AUROC demonstrates that treating the entire reasoning trace as a single text input is a powerful strategy for discrimination. This approach likely captures global semantic and stylistic patterns that are indicative of a correct answer. However, the performance of \texttt{ETTIN-HGA} provides a crucial insight. The only difference between these models is the HGA layer (described in \ref{app:ETTIN-HGA-Model-Desc}), which explicitly models the chunk-level structure of the reasoning. This single architectural change improves the ECE from 0.160 to 0.148 (a 7.5\% relative improvement), establishing it as the best-calibrated model in our benchmark. The reliability diagrams (Figure~\ref{fig:calibration_grid_v2_dataset_cells_best_10bins_2x3}) further illustrate its strong calibration, with \texttt{ETTIN-HGA} closely following the ideal diagonal line across multiple LRMs. This demonstrates that while holistic text representations are sufficient for ranking, achieving reliable calibration requires explicit awareness of the model's internal, step-by-step reasoning structure. However, even \texttt{ETTIN-HGA}'s strong calibration degrades substantially on certain datasets (particularly MedMCQA), suggesting that dataset characteristics significantly impact calibration quality beyond architectural choices.

\noindent \textbf{Architectural complexity does not guarantee improved performance.} \quad
A surprising result from our benchmark is that increased architectural complexity does not reliably lead to better performance. For example, the sophisticated \texttt{GNN-SR-Transformer} model, which processes a densely forward-connected graph with rich, 5-dimensional edge features, achieves a similar AUROC (0.656 vs. 0.653) and a worse ECE (0.175 vs. 0.165) compared to the much simpler \texttt{SFHS-Conv}, which applies a 1D convolution over a flat sequence of hidden states. This finding is reflected in Figure~\ref{fig:ECE_vs_AUROC_ellipse_distribution}, where the ellipses for the more complex \texttt{GNN-SR} family are largely overlapping with, or even outperformed by, the simpler \texttt{SFHS} models. This pattern suggests that the primary limitation may not be the sophistication of the architecture, but rather the information content available in the chunk-level hidden states themselves.

\noindent \textbf{Generative uncertainty signals are strong but insufficient for discrimination.} \quad
The inclusion of \texttt{TLCC} provides a critical additional perspective on the observed performance ceiling. By relying exclusively on token-level logit statistics aggregated at the chunk level, \texttt{TLCC} removes hidden state representations entirely while retaining access to the model’s intrinsic uncertainty signals. Across LRMs, \texttt{TLCC} achieves competitive calibration and, in several cases, improved specificity relative to hidden-state-based baselines, confirming that logit-derived statistics are effective indicators of uncertainty. However, \texttt{TLCC} consistently underperforms text-based encoders such as \texttt{ETTIN} in AUROC, indicating a reduced ability to separate correct from incorrect reasoning trajectories. This result reinforces a key conclusion of our benchmark: uncertainty signals alone, even when aggregated over full reasoning traces, are insufficient to capture the semantic and logical distinctions required for maximal discriminative power. High-performing confidence estimation for LRMs therefore appears to require both uncertainty-aware signals and access to global semantic context.

\noindent \textbf{Two-stage feature extraction is a viable but inconsistent strategy.} \quad
Several of our methods (\texttt{GNN-CD}, \texttt{LateFusion}, \texttt{CE}) use a two-stage setup where a \texttt{PHSV-half} model is trained first and then used as a feature extractor. This approach is reasonable but not always effective. The best two-stage model, \texttt{GNN-CD-noft-GCN2Conv-dual} (AUROC 0.651), performs comparably to the best single-stage hidden state models, yet many other \texttt{GNN-CD} variants perform poorly, as shown in the lower-left quadrant of Figure~\ref{fig:ECE_vs_AUROC_ellipse_distribution}. End-to-end fine-tuning of \texttt{PHSV-half} features (\texttt{ft} variants) also failed to consistently improve results. This suggests that while local confidence signals help, integrating them into a global model remains challenging.

\noindent \textbf{Simple baselines and direct confidence probing are unreliable.} \quad
Our benchmark confirms that direct confidence estimation methods perform poorly on long-form reasoning tasks. \texttt{YVCE}, which relies on the LRM's own self-assessment, is the least calibrated model we tested. Similarly, \texttt{PHSV}, which uses a local classifier in an early-exit fashion, performs weakly on both AUROC and ECE. The poor calibration of these methods is evident in Figure~\ref{fig:calibration_grid_v2_dataset_cells_best_10bins_2x3}. The reliability curve for \texttt{YVCE} is unstable, showing erratic fluctuations across the confidence spectrum. The curve for \texttt{PHSV}, though less volatile, remains misaligned across different LRM architectures. This finding shows that the problem is non-trivial and that more advanced modeling of the full reasoning trace is necessary for reliable confidence estimation.

%% file: sections/conclusion.tex
In this work, we address the critical and underexplored challenge of confidence estimation for LRMs by introducing the Reasoning Model Confidence estimation Benchmark (RMCB), a publicly available resource of 347,496 reasoning traces from six popular LRMs across diverse high-stakes domains, each paired with correctness annotations. Using this benchmark, we conducted a comprehensive empirical evaluation of over ten representation-based methods, revealing a persistent trade-off between a method's discriminative power (AUROC) and its calibration (ECE). Our results show that text-based encoders like \texttt{ETTIN} achieve the best discrimination, while architectures that explicitly model the reasoning structure, such as \texttt{ETTIN-HGA} and simple graph-based models, yield superior calibration. Furthermore, we find that increased architectural complexity does not guarantee improved performance, with sophisticated graph neural networks failing to consistently outperform simpler sequential baselines. Ultimately, our work provides a clear map of the current performance landscape, rigorously quantifies the limitations of relying solely on chunk-level hidden states, and establishes a robust set of baselines that highlight the need for future research to explore alternative signal sources, such as token-level generative signals, to overcome the performance plateau we have identified.

%% file: sections/limitations.tex
While our work provides a comprehensive benchmark for representation-based LRM confidence estimation, its scope is defined by several key methodological choices that present avenues for future research. Primarily, our investigation focuses on methods that operate on chunk-level representations, including both internal hidden states and aggregated token-level generative statistics. This design choice enables a computationally tractable and scalable feature set, avoiding the significant storage and processing overhead required to model full token-level distributions (e.g., complete logit vectors for every token in each of the 347,496 traces). While the inclusion of aggregated logit statistics addresses a simple and computationally efficient form of generative uncertainty, our results indicate that more expressive modeling of reasoning dynamics beyond chunk-level summaries remains an important direction for future work. Furthermore, our methodological scope includes several other boundaries. Our response segmentation into ``chunks,'' while an enhancement of the method proposed by \citet{PHSV}, remains a heuristic based on linguistic keywords. Our correctness annotations, while scalable, rely on an LLM judge (\texttt{GPT-5-nano}) rather than expert human annotation. We also excluded consistency-based methods from our benchmark due to the prohibitive inference cost of generating multiple long-form reasoning traces per query. Finally, the RMCB is currently limited to the English language and a specific suite of LRM architectures. We believe these limitations do not detract from our core findings but instead provide a clear and promising roadmap for future investigations into the reliability of LRMs.

%% file: sections/ethical_considerations.tex
Although the goal of this work is to improve LRM reliability, several ethical considerations are important. First, a key risk is over-relying on any automated confidence score. Our results show that even the best methods have trade-offs and no method is perfect. In high-stakes fields like medicine, finance, or law, using these scores to automatically approve LRM outputs without human oversight could lead to harmful outcomes if a model error goes unnoticed. Second, fairness is a critical issue. The LRMs used in our benchmark may carry biases from their own training data, and the confidence models we trained could inherit or even amplify these issues. As a result, the confidence scores might be less reliable for certain demographic groups or types of questions, which could lead to unfair outcomes. Therefore, any method developed or evaluated on this benchmark should be treated as a tool to assist human experts, not to replace their critical judgment. We strongly recommend that any real-world deployment of these confidence estimators be preceded by thorough fairness testing and ongoing monitoring to ensure they are used responsibly.

%% file: sections/appendix_llm.tex
\section{Reasoning Language Models}
\label{app:appendix_llm}

We evaluate a set of open-weight reasoning-oriented LLMs from multiple families, including Microsoft's Phi-4 series, Qwen's Qwen3 and QwQ lines, Mistral's Magistral, and LG AI Research's EXAONE. These models range in size from $\sim$3.85B to $\sim$32.8B parameters and span both dense and hybrid architectures, with context lengths from 32K up to 131K tokens. Table~\ref{tab:models_summary_arch} and Table~\ref{tab:models_summary_ctx} summarize the key configuration details extracted directly from the model configuration files.

\begin{table*}[t]
\centering
\caption{Architectural parameters of evaluated models.}
\label{tab:models_summary_arch}
\footnotesize
\begin{tabular}{l r r r r}
\toprule
\textbf{Hub ID} & \textbf{Params (B)} & \textbf{Hidden Size} & \textbf{Layers} & \textbf{Attention Heads (Q/KV)} \\
\midrule
\href{https://huggingface.co/microsoft/Phi-4-mini-flash-reasoning}{microsoft/Phi-4-mini-flash-reasoning} & 3.85 & 2560 & 32 & 40 / 20 \\
\href{https://huggingface.co/Qwen/Qwen3-8B}{Qwen/Qwen3-8B} & 8.19 & 4096 & 36 & 32 / 8 \\
\href{https://huggingface.co/Qwen/Qwen3-14B}{Qwen/Qwen3-14B} & 14.8 & 5120 & 40 & 40 / 8 \\
\href{https://huggingface.co/mistralai/Magistral-Small-2506}{mistralai/Magistral-Small-2506} & 23.6 & 5120 & 40 & 32 / 8 \\
\href{https://huggingface.co/LGAI-EXAONE/EXAONE-Deep-32B}{LGAI-EXAONE/EXAONE-Deep-32B} & 32.0 & 5120 & 64 & 40 / 8 \\
\href{https://huggingface.co/Qwen/QwQ-32B}{Qwen/QwQ-32B} & 32.8 & 5120 & 64 & 40 / 8 \\
\bottomrule
\end{tabular}
\end{table*}

\begin{table*}[t]
\centering
\caption{Context, feed-forward, and vocabulary parameters of evaluated models.}
\label{tab:models_summary_ctx}
\footnotesize
\begin{tabular}{l r r r}
\toprule
\textbf{Hub ID} & \textbf{Feed-Forward Size} & \textbf{Context Length} & \textbf{Vocabulary Size} \\
\midrule
\href{https://huggingface.co/microsoft/Phi-4-mini-flash-reasoning}{microsoft/Phi-4-mini-flash-reasoning} & 10240 & 64K & 200{,}064 \\
\href{https://huggingface.co/Qwen/Qwen3-8B}{Qwen/Qwen3-8B} & 12288 & 32K & 151{,}936 \\
\href{https://huggingface.co/Qwen/Qwen3-14B}{Qwen/Qwen3-14B} & 17408 & 32K & 151{,}936 \\
\href{https://huggingface.co/mistralai/Magistral-Small-2506}{mistralai/Magistral-Small-2506} & 32768 & 128K & 131{,}072 \\
\href{https://huggingface.co/LGAI-EXAONE/EXAONE-Deep-32B}{LGAI-EXAONE/EXAONE-Deep-32B} & 27392 & 32K & 102{,}400 \\
\href{https://huggingface.co/Qwen/QwQ-32B}{Qwen/QwQ-32B} & 27648 & 131K & 152{,}064 \\
\bottomrule
\end{tabular}

\vspace{2pt}
\raggedright
\end{table*}

%% file: sections/appendix_dataset.tex
\section{RMCB Benchmark Construction}
\label{app:appendix_dataset}
This section provides a detailed breakdown of the construction of the RMCB benchmark. We describe the full suite of datasets used to create our training and evaluation splits, the standardized schema for data curation, and the methodology for LRM response generation and annotation. All datasets are in English. For comprehensive information regarding the original construction and domain coverage of the source benchmarks, we refer readers to their respective publications.

\subsection{Data Curation and Standard Schema}

To create a consistent format for our experiments, all raw datasets were processed into a standardized JSONL format. Each line in the resulting files corresponds to a single reasoning problem and contains the following fields, which are derived from the properties available in each source dataset:

\begin{itemize}
    \item \textbf{\texttt{prompt}}: The input question from the dataset, formatted into a string that is provided to the language model.
    \item \textbf{\texttt{explanation}}: Any relevant supporting information for the ground-truth answer provided by the source dataset. This can range from a detailed, step-by-step reasoning trace to shorter contextual details, or it may be empty if no such information is available.
    \item \textbf{\texttt{answer}}: The ground-truth final answer to the prompt, as specified by the source dataset.
    \item \textbf{\texttt{category}}: A categorization of the sample if it exists in the source dataset (e.g., "Surgery" for a sample in MedMCQA).
    \item \textbf{\texttt{dataset}}: The name of the source dataset.
    \item \textbf{\texttt{record\_id}}: A unique and deterministic identifier for each sample, generated using a dataset-specific hashing function applied to the preprocessed input fields used during evaluation. The exact hashing procedures are provided in the dataset reconstruction script to ensure full reproducibility of identifiers.
\end{itemize}

This standardized schema allows for consistent data handling and feature extraction across all models and datasets in our benchmark.

\begin{table*}[t]
\centering
\caption{Overall dataset distribution for training. Each dataset contributes exactly 1,000 unique examples (seed=23) to form a balanced 10,000-sample training set. Domains indicate the primary reasoning type assessed.}
\footnotesize
\label{tab:train_dataset_distribution}
\begin{tabular}{l l r r}
\toprule
\textbf{Dataset} & \textbf{Domain} & \textbf{Train Samples} & \textbf{Train (\%)} \\
\midrule
GSM8K & Mathematical Reasoning & 1,000 & 10.00 \\
TAT-QA & Financial Reasoning & 1,000 & 10.00 \\
MedQA & Medical Reasoning & 1,000 & 10.00 \\
LEXam & Legal Reasoning & 1,000 & 10.00 \\
ARC & General Reasoning & 1,000 & 10.00 \\
CommonsenseQA2 & General Reasoning & 1,000 & 10.00 \\
LogiQA & General Reasoning & 1,000 & 10.00 \\
OpenBookQA & General Reasoning & 1,000 & 10.00 \\
QuaRTz & General Reasoning & 1,000 & 10.00 \\
ReClor & General Reasoning & 1,000 & 10.00 \\
\bottomrule
\end{tabular}
\end{table*}

\subsection{Dataset Sources and Versions}
\label{app:dataset_sources}

Table~\ref{tab:dataset_sources} provides a complete overview of all datasets used to construct the RMCB benchmark, including their domains, splits, access methods, and exact versions or revisions. For datasets hosted on HuggingFace, we record the repository identifier, configuration where applicable, and commit hash to ensure reproducibility. For datasets requiring manual download, we explicitly note the source and provide step-by-step download and preprocessing instructions in the public RMCB code repository associated with this paper. This table is intended to make dataset provenance explicit and verifiable.

\begin{table*}[p]
\centering
\small
\caption{Dataset sources and exact revisions or versions used to construct the RMCB benchmark.}
\label{tab:dataset_sources}
\begin{tabular}{ll}
\toprule
\textbf{Dataset} & \textbf{Details} \\
\midrule

\multirow{2}{*}{GSM8K}
& Source: HuggingFace \texttt{openai/gsm8k} (config: \texttt{socratic}) \\
& Revision: \texttt{cc7b047b6e5bb11b4f1af84efc572db110a51b3c} \\
\addlinespace

\multirow{2}{*}{TAT-QA}
& Source: GitHub \texttt{NExTplusplus/TAT-QA} \\
& Revision: not versioned \\
\addlinespace

\multirow{2}{*}{MedQA}
& Source: Google Drive release \\
& Revision: \texttt{ddef95d268cdad413693d634279a9a679d468469} \\
\addlinespace

\multirow{2}{*}{LEXam}
& Source: HuggingFace \texttt{LEXam-Benchmark/LEXam} \\
& Revision: \texttt{68f21a324eb0e14837be42f10b644c40847c3ed4} \\
\addlinespace

\multirow{2}{*}{ARC}
& Source: HuggingFace \texttt{allenai/ai2\_arc} (config: \texttt{ARC-Challenge}) \\
& Revision: \texttt{210d026faf9955653af8916fad021475a3f00453} \\
\addlinespace

\multirow{2}{*}{CommonsenseQA-2}
& Source: HuggingFace \texttt{chiayewken/commonsense-qa-2} \\
& Revision: \texttt{15e7dc364f7906ad69cbe4a0bed697ba12f07bdf} \\
\addlinespace

\multirow{2}{*}{LogiQA}
& Source: HuggingFace \texttt{lucasmccabe/logiqa} \\
& Revision: \texttt{3c19b0488d794d30c36f73d132d8a22e64f42f2e} \\
\addlinespace

\multirow{2}{*}{OpenBookQA}
& Source: HuggingFace \texttt{allenai/openbookqa} (config: \texttt{main}) \\
& Revision: \texttt{388097ea7776314e93a529163e0fea805b8a6454} \\
\addlinespace

\multirow{2}{*}{QuaRTz}
& Source: HuggingFace \texttt{allenai/quartz} \\
& Revision: \texttt{28c1dbb56caf81799296cb17892fa73402e23464} \\
\addlinespace

\multirow{2}{*}{ReClor}
& Source: HuggingFace \texttt{voidful/ReClor} \\
& Revision: \texttt{809ebe44b8dde882c4190f4178b27676b941b933} \\
\addlinespace

\multirow{2}{*}{MATH}
& Source: Kaggle \texttt{awsaf49/math-dataset} \\
& Version: 1 \\
\addlinespace

\multirow{2}{*}{FinQA}
& Source: GitHub \texttt{czyssrs/FinQA} \\
& Revision: \texttt{0f16e2867befa6840783e58be38c9efb9229d742} \\
\addlinespace

\multirow{2}{*}{MedMCQA}
& Source: HuggingFace \texttt{openlifescienceai/medmcqa} \\
& Revision: \texttt{91c6572c454088bf71b679ad90aa8dffcd0d5868} \\
\addlinespace

\multirow{2}{*}{LegalBench}
& Source: HuggingFace \texttt{nguha/legalbench} \\
& Revision: \texttt{e042ea68c19df12b737fe768572f22ead61e8e37} \\
\addlinespace

\multirow{2}{*}{MMLU-Pro}
& Source: HuggingFace \texttt{TIGER-Lab/MMLU-Pro} \\
& Revision: \texttt{dd36ce4b34827164989f100331f82c5a29741747} \\
\addlinespace

\multirow{2}{*}{BBH}
& Source: HuggingFace \texttt{maveriq/bigbenchhard} \\
& Revision: \texttt{d53c5b10a77edeb29da195f47e6086b29f2f7f74} \\

\bottomrule
\end{tabular}
\end{table*}

\subsection{Training Datasets}
The trainable confidence models are developed on a balanced training set of 10,000 samples aggregated from ten reasoning datasets, shown in Table~\ref{tab:train_dataset_distribution}. From each dataset, we deterministically sampled exactly 1,000 unique examples using \texttt{seed=23}, then merged them to form the final training set. These datasets span a variety of high-stakes and general reasoning domains—including mathematical, financial, medical, legal, and general reasoning—ensuring that the training data covers a wide spectrum of reasoning types. Below, we detail the processing for selected training datasets as representative examples of how raw benchmarks were mapped to our standard schema.

\subsubsection{GSM8K (Mathematical Reasoning)}
The Grade School Math 8K (GSM8K) dataset \cite{GSM8K}, released under the MIT License, is a benchmark designed to test multi-step mathematical reasoning. It consists of high-quality, linguistically diverse word problems whose solutions require a sequence of elementary calculations. We utilize the \texttt{socratic} configuration, which provides detailed, chain-of-thought style solutions for each problem. The raw dataset examples are mapped to our standard schema as follows:
\begin{itemize}
    \item \textbf{\texttt{prompt}}: The input question is formatted using the following template:
    \begin{tcolorbox}[title=Prompt Template,fonttitle=\bfseries,fontupper=\small]
Question: \{question\}  

Answer:
    \end{tcolorbox}
    \item \textbf{\texttt{explanation}}: The full, unaltered reasoning trace from the original \texttt{answer} field of the dataset.
    \item \textbf{\texttt{answer}}: The final numerical answer is parsed from the text that follows the \texttt{\#\#\#\#} marker at the end of the reasoning trace.
    \item \textbf{\texttt{category}}: This field is not applicable to the GSM8K dataset and is set to "N/A".
\end{itemize}

\subsubsection{TAT-QA (Financial Reasoning)}
The Tabular and Textual Question Answering (TAT-QA) dataset \cite{TATQA}, released for non-commercial use, is a benchmark for financial reasoning over hybrid data. Each sample contains both unstructured text and a structured table extracted from real-world financial reports, requiring a model to synthesize information from both sources. Answering questions correctly often involves complex numerical reasoning, such as addition, subtraction, and comparison. The raw dataset examples are mapped to our standard schema as follows:
\begin{itemize}
    \item \textbf{\texttt{prompt}}: A comprehensive prompt is constructed for each question. First, a context is created by verbalizing the structured table into natural language. This process follows the methodology of FinQA \cite{FinQA}, where each row is converted into a descriptive sentence using a template similar to  \textit{`the \{column name\} of \{row name\} is \{cell value\};`}. This verbalized table text is then concatenated with the original paragraphs to form a complete context. This context is then embedded into a template akin to \citet{TATLLM}, which is formatted as follows:

    \begin{tcolorbox}[title=Prompt Template,fonttitle=\bfseries,fontupper=\small]
Below is an instruction that describes a question answering task in the finance domain, 
paired with an input table and its relevant text that provide further context. 
Generate an appropriate answer to the given question.

Question: \{question\}

Context:
\{paragraphs and verbalized table\}

Answer:
    \end{tcolorbox}

    \item \textbf{\texttt{explanation}}: This field is populated with the \texttt{derivation} from the source dataset, which provides a step-by-step reasoning trace for the answer.
    \item \textbf{\texttt{answer}}: The ground-truth \texttt{answer} from the source dataset is normalized into a consistent string format.
    \item \textbf{\texttt{category}}: This field is a composite created by joining the \texttt{answer\_type} and \texttt{answer\_from} fields from the source data (e.g., "multi-span[SEP]table-text"). Distributions of \texttt{answer\_type} and \texttt{answer\_from} are provided in Tables~\ref{tab:tatqa_answer_types} and~\ref{tab:tatqa_answer_sources}, respectively.
\end{itemize}

\begin{table}[h]
\footnotesize
\centering
\caption{TAT-QA answer types distribution (Train).}
\label{tab:tatqa_answer_types}
\begin{tabular}{lrr}
\toprule
\textbf{Answer Type} & \textbf{Train Samples} & \textbf{Train (\%)} \\
\midrule
span & 438 & 43.8 \\
arithmetic & 414 & 41.4 \\
multi-span & 121 & 12.1 \\
count & 27 & 2.7 \\
\bottomrule
\end{tabular}
\end{table}

\begin{table}[h]
\footnotesize
\centering
\caption{TAT-QA answer sources distribution (Train).}
\label{tab:tatqa_answer_sources}
\begin{tabular}{lrr}
\toprule
\textbf{Answer Source} & \textbf{Train Samples} & \textbf{Train (\%)} \\
\midrule
table & 448 & 44.8 \\
table-text & 334 & 33.4 \\
text & 218 & 21.8 \\
\bottomrule
\end{tabular}
\end{table}

\subsubsection{MedQA (Medical Reasoning)}
The MedQA dataset \cite{MedQA}, released under a research-use-only license, is a large-scale, multiple-choice question benchmark designed to test professional medical knowledge, with questions sourced from medical board exams in the US, Mainland China, and Taiwan. The questions are varied and often require a deep understanding of clinical scenarios to arrive at the correct diagnosis or treatment. The raw dataset examples are mapped to our standard schema as follows:
\begin{itemize}
    \item \textbf{\texttt{prompt}}: Each example is formatted into a standardized multiple-choice prompt. To ensure consistency, the options are sorted alphabetically by their key (e.g., A, B, C) before being inserted into the following template:
    \begin{tcolorbox}[title=Prompt Template,fonttitle=\bfseries,fontupper=\small]
Question: {question}

Choices:

(A) \{option\_A\}

(B) \{option\_B\}

...

Answer:
    \end{tcolorbox}
    \item \textbf{\texttt{explanation}}: This field is intentionally left empty as the source dataset does not provide reasoning traces for the answers.
    \item \textbf{\texttt{answer}}: The ground-truth \texttt{answer} is the single letter corresponding to the correct option (e.g., "B").
    \item \textbf{\texttt{category}}: This field is populated with the \texttt{meta\_info} from the source dataset, which typically specifies the medical sub-domain (e.g., "Internal Medicine"). In addition, the dataset is categorized by USMLE exam step (Step 1, Step 2 \& 3), as shown in Table~\ref{tab:medqa_categories}.
\end{itemize}

\begin{table}[h]
\footnotesize
\centering
\caption{MedQA categories distribution (Train).}
\label{tab:medqa_categories}
\begin{tabular}{lrr}
\toprule
\textbf{Category} & \textbf{Train Samples} & \textbf{Train (\%)} \\
\midrule
Step 1 & 552 & 55.2 \\
Step 2 \& 3 & 448 & 44.8 \\
\bottomrule
\end{tabular}
\end{table}

\subsubsection{LEXam (Legal Reasoning)}
The LEXam benchmark \cite{LEXam}, released under the CC BY 4.0 License, is a collection of legal examination questions designed to test complex legal reasoning across various jurisdictions and legal areas, sourced from 340 real law exams. We process all English-language samples from three distinct variants: \texttt{open\_question}, \texttt{mcq\_4\_choices}, and \texttt{mcq\_perturbation}. The raw dataset examples are mapped to our standard schema as follows:
\begin{itemize}
    \item \textbf{\texttt{prompt}}: The prompts are tailored to the question type. 
    
    For the \texttt{open\_question} variant, a detailed system prompt guides the model to provide a structured, exam-style legal analysis (see Figure~\ref{fig:lexam-open-prompt}).
    \begin{figure*}[t]
    \centering
    \begin{tcolorbox}[title=Open Question Prompt Template,fonttitle=\bfseries,fontupper=\small]
You are an expert in \{course\_name\} and address legal issues in a structured, exam-style manner.
Assume the applicable jurisdiction unless specifically mentioned; if the course context justifies, address legal issues beyond the stated jurisdiction as well.

Use precise legal language and formal academic tone when answering.

Do NOT state any disclaimer or refer to the need for external legal advice.

Do NOT request the user to consult laws or to research on their own.

Offer focused legal analyses and individualized advice.

Speak directly and authoritatively without mentioning that your response is merely for general information.

Incorporate jurisdiction-specific legal terminology where appropriate.

If you have discovered relevant legal considerations, respond with a concise, clear legal analysis.

Cite the specific legal provision, explicitly indicating sections, subsections, or paragraphs where available (e.g., “Section 74(2)(b) of the Contracts Act”).

Avoid vague references without specifying applicable subsections or clauses.

If no relevant considerations are found, explicitly state that no pertinent information is available.

If you do have reliable sources, share practical guidance or insights from them.

Respond in the same language as the question.

If the question specifically requests a short answer, provide a concise response.

If the prompt asks you to analyze a specific case provided in the exam, but the text or details of that case have not been provided in the prompt, explicitly flag that the required case material is missing.

Question: 

\{question\}

Answer:
    \end{tcolorbox}
    \caption{Open Question Prompt Template for the \texttt{open\_question} variant.}
    \label{fig:lexam-open-prompt}
    \end{figure*}

    For the \texttt{mcq} variants, a different prompt instructs the model to use a step-by-step, chain-of-thought process to analyze the facts, explain relevant legal rules, and justify its final choice (see Figure~\ref{fig:lexam-mcq-prompt}).
    \begin{figure*}[t]
    \centering
    \begin{tcolorbox}[title=MCQ Prompt Template,fonttitle=\bfseries,fontupper=\small]
You are an expert in \{course\_name\} and address legal issues in a structured, exam-style manner.
You are given a multiple-choice question, where only one choice (e.g., A, B, C, etc.) is correct.
Assume the applicable jurisdiction unless specifically stated otherwise. If the context of the course justifies it, consider legal frameworks beyond the stated jurisdiction as well.

Please reason through the question step by step, using a chain-of-thought approach:

- Clarify the facts: Briefly restate or highlight the key facts in the question to anchor your reasoning.

- Issue Identification: What legal issue(s) arise from the facts?

- Rule Explanation: What legal rules or principles are relevant, and what are their sources (e.g., statutes, case law, doctrine)?

- Application and Reasoning: Apply the relevant rules to the facts, carefully weighing any ambiguities, exceptions, or competing interpretations.

- Eliminate Incorrect Answers: Briefly explain why each incorrect answer is wrong or less convincing.

- Conclusion: Clearly state the correct answer choice (e.g., A, B, C, etc.) with a brief justification for why it best fits the legal analysis.

Format your final answer as follows:

Correct Answer: C

Question:

\{question\_with\_choices\}

Answer:
    \end{tcolorbox}
    \caption{Multiple-choice (MCQ) Prompt Template for the \texttt{mcq} variants.}
    \label{fig:lexam-mcq-prompt}
    \end{figure*}
    Both prompt templates above are reproduced directly from the main manuscript of the LEXam benchmark~\cite{LEXam}.
    \item \textbf{\texttt{explanation}}: This field is intentionally left empty as the source dataset provides reference answers but not step-by-step reasoning traces.
    \item \textbf{\texttt{answer}}: This field contains the full text of the ground-truth answer for open questions and the single correct letter for MCQ variants.
    \item \textbf{\texttt{category}}: This field is a composite created by joining the \texttt{area}, \texttt{jurisdiction}, and \texttt{course} fields from the source data (e.g., "Private[SEP]International[SEP]Comparative Private Law"). Distributions for each of these components are shown in Tables~\ref{tab:lexam_area}, \ref{tab:lexam_jurisdiction}, and~\ref{tab:lexam_course}.
\end{itemize}

\begin{table}[h]
\footnotesize
\centering
\caption{LEXam area distribution (Train).}
\label{tab:lexam_area}
\begin{tabular}{lrr}
\toprule
\textbf{Area} & \textbf{Train Samples} & \textbf{Train (\%)} \\
\midrule
Interdisciplinary & 493 & 49.3 \\
Private & 249 & 24.9 \\
Public & 122 & 12.2 \\
Unknown & 118 & 11.8 \\
Criminal & 18 & 1.8 \\
\bottomrule
\end{tabular}
\end{table}

\begin{table}[h]
\footnotesize
\centering
\caption{LEXam jurisdiction distribution (Train).}
\label{tab:lexam_jurisdiction}
\begin{tabular}{lrr}
\toprule
\textbf{Jurisdiction} & \textbf{Train Samples} & \textbf{Train (\%)} \\
\midrule
Swiss & 471 & 47.1 \\
International & 366 & 36.6 \\
Unknown & 118 & 11.8 \\
Generic & 45 & 4.5 \\
\bottomrule
\end{tabular}
\end{table}

\begin{table*}[t]
\footnotesize
\centering
\caption{LEXam course distribution (Train).}
\label{tab:lexam_course}
\begin{tabular}{lrr}
\toprule
\textbf{Course} & \textbf{Train Samples} & \textbf{Train (\%)} \\
\midrule
Swiss Law & 582 & 58.2 \\
US Business Law & 96 & 9.6 \\
International Organisations & 34 & 3.4 \\
Chinese Business Law & 29 & 2.9 \\
International Finance Law & 23 & 2.3 \\
International Commercial Arbitration & 23 & 2.3 \\
Comparative Private Law & 20 & 2.0 \\
European Economic Law & 20 & 2.0 \\
Legal Theory & 20 & 2.0 \\
International Sales Law & 18 & 1.8 \\
Foundations and Trusts & 18 & 1.8 \\
International Criminal Law & 18 & 1.8 \\
History of Business Law & 16 & 1.6 \\
Transnational Public Security Law & 15 & 1.5 \\
Legal Sociology & 9 & 0.9 \\
Principles of Corporate Law & 9 & 0.9 \\
International Financial Law & 8 & 0.8 \\
International Human Rights & 8 & 0.8 \\
Introduction to Sports Law & 7 & 0.7 \\
International Economic Law & 7 & 0.7 \\
Comparative Constitutional Law & 7 & 0.7 \\
Gesellschaftsrecht & 7 & 0.7 \\
Comparative Corporate Law & 6 & 0.6 \\
\bottomrule
\end{tabular}
\end{table*}

\subsubsection{ARC-Challenge (General Science Reasoning)}
The AI2 Reasoning Challenge (ARC) dataset \cite{ARC}, released under the CC BY-SA 4.0 License, is a collection of multiple-choice science questions designed to be challenging for AI systems, requiring knowledge and reasoning beyond simple retrieval. We use the more difficult \texttt{ARC-Challenge} subset and process examples exclusively from its \texttt{train} split. The raw dataset examples are mapped to our standard schema as follows:
\begin{itemize}
    \item \textbf{\texttt{prompt}}: Each question is formatted into a standard multiple-choice prompt, with the choices presented in their original order from the dataset. The template is as follows:
    \begin{tcolorbox}[title=Prompt Template,fonttitle=\bfseries,fontupper=\small]
Question: \{question\}

Choices:

(A) \{option\_A\}

(B) \{option\_B\}

...

Answer:
    \end{tcolorbox}
    \item \textbf{\texttt{explanation}}: This field is intentionally left empty as the source dataset does not provide reasoning traces.
    \item \textbf{\texttt{answer}}: The ground-truth \texttt{answer} is the correct option letter from the original \texttt{answerKey} field.
    \item \textbf{\texttt{category}}: This field is derived from the prefix of the question’s \texttt{id} field (e.g., ``Mercury'', ``MCAS''). The category distribution is shown in Table~\ref{tab:arc_categories}.
\end{itemize}

\begin{table}[h]
\footnotesize
\centering
\caption{ARC-Challenge categories distribution (Train).}
\label{tab:arc_categories}
\begin{tabular}{lrr}
\toprule
\textbf{Category} & \textbf{Train Samples} & \textbf{Train (\%)} \\
\midrule
Mercury      & 647 & 64.7 \\
MCAS         & 89  & 8.9 \\
NYSEDREGENTS & 38  & 3.8 \\
ACTAAP       & 36  & 3.6 \\
MDSA         & 31  & 3.1 \\
TIMSS        & 24  & 2.4 \\
NCEOGA       & 23  & 2.3 \\
AKDE\&ED     & 20  & 2.0 \\
VASoL        & 16  & 1.6 \\
MEA          & 15  & 1.5 \\
LEAP         & 13  & 1.3 \\
MSA          & 12  & 1.2 \\
CSZ          & 7   & 0.7 \\
AIMS         & 5   & 0.5 \\
TAKS         & 4   & 0.4 \\
MEAP         & 4   & 0.4 \\
NAEP         & 4   & 0.4 \\
OHAT         & 3   & 0.3 \\
CSZ30494     & 1   & 0.1 \\
CSZ30564     & 1   & 0.1 \\
NCEOGA2013   & 1   & 0.1 \\
WASL         & 1   & 0.1 \\
CSZ10245     & 1   & 0.1 \\
CSZ20740     & 1   & 0.1 \\
CSZ30771     & 1   & 0.1 \\
CSZ20059     & 1   & 0.1 \\
FCAT         & 1   & 0.1 \\
\bottomrule
\end{tabular}
\end{table}

\subsubsection{CommonsenseQA 2.0 (General Reasoning)}
The CommonsenseQA 2.0 dataset \cite{commonsenseqa2}, released under the CC BY 4.0 License, is a challenging benchmark of 14,343 yes/no questions designed to test a model's commonsense reasoning capabilities. The questions were created through a gamified process to be adversarial to language models. The raw dataset examples are mapped to our standard schema as follows:
\begin{itemize}
    \item \textbf{\texttt{prompt}}: The input question is formatted into a standardized yes/no multiple-choice prompt using the following template:
    \begin{tcolorbox}[title=Prompt Template,fonttitle=\bfseries,fontupper=\small]
Question: \{question\}

Choices:

(A) yes

(B) no

Answer:
    \end{tcolorbox}
    \item \textbf{\texttt{explanation}}: This field is populated with the \texttt{relational\_prompt} from the source dataset (e.g., "is capable of", "causes"), which indicates the type of commonsense reasoning being tested.
    \item \textbf{\texttt{answer}}: The ground-truth \texttt{answer} is converted to a letter, where "yes" maps to "A" and "no" maps to "B".
    \item \textbf{\texttt{category}}: This field is a composite created by joining the \texttt{topic\_prompt} and boolean flags indicating whether the topic and relational prompts were used in the question's construction (e.g., ``world trade center[SEP]True[SEP]True''). The distribution of prompt usage is provided in Table~\ref{tab:commonsenseqa2_prompt_usage}.
\end{itemize}

\begin{table*}[t]
\footnotesize
\centering
\caption{CommonsenseQA2 prompt usage distribution (Train).}
\label{tab:commonsenseqa2_prompt_usage}
\begin{tabular}{llrr}
\toprule
\textbf{relational\_prompt\_used} & \textbf{topic\_prompt\_used} & \textbf{Train Samples} & \textbf{Train (\%)} \\
\midrule
False & False & 21  & 2.1 \\
False & True  & 123 & 12.3 \\
True  & False & 7   & 0.7 \\
True  & True  & 849 & 84.9 \\
\bottomrule
\end{tabular}
\end{table*}

\subsubsection{LogiQA (General Reasoning)}
The LogiQA dataset \cite{LogiQA}, released under the CC BY-NC-SA 4.0 License, is a benchmark designed to test a model's capability for logical reasoning within a machine reading comprehension context. It is sourced from expert-written questions for civil servant exams and covers multiple types of deductive reasoning, such as categorical, conditional, and disjunctive reasoning. A key challenge of this dataset is that the correct answer is typically not a direct span in the text but must be inferred through logical steps. The raw dataset examples are mapped to our standard schema as follows:
\begin{itemize}
    \item \textbf{\texttt{prompt}}: The input is formatted as a multiple-choice question that includes a context paragraph, the query, and four options. The template is as follows:
    \begin{tcolorbox}[title=Prompt Template,fonttitle=\bfseries,fontupper=\small]
Context: \{context\}

Question: \{query\}

Choices:

(A) \{option\_A\}

(B) \{option\_B\}

(C) \{option\_C\}

(D) \{option\_D\}

Answer:
    \end{tcolorbox}
    \item \textbf{\texttt{explanation}}: This field is intentionally left empty as the source dataset does not provide explicit reasoning traces.
    \item \textbf{\texttt{answer}}: The ground-truth \texttt{answer} is the single letter corresponding to the correct option.
    \item \textbf{\texttt{category}}: This field is not applicable to the LogiQA dataset and is set to "N/A".
\end{itemize}

\subsubsection{OpenBookQA (General Science Reasoning)}
The OpenBookQA dataset \cite{OpenBookQA}, released under the Apache License 2.0, is a multiple-choice question answering benchmark designed to test a deeper understanding of elementary-level science. The questions require combining a core science fact from a provided "open book" with broad common knowledge to arrive at the correct answer. We process examples from the \texttt{main} configuration's \texttt{train} split. The raw dataset examples are mapped to our standard schema as follows:
\begin{itemize}
    \item \textbf{\texttt{prompt}}: Each question is formatted into a standard multiple-choice prompt, with the choices presented in their original order from the dataset. The template is as follows:
    \begin{tcolorbox}[title=Prompt Template,fonttitle=\bfseries,fontupper=\small]
Question: \{question\}

Choices:

(A) \{option\_A\}

(B) \{option\_B\}

...

Answer:
    \end{tcolorbox}
    \item \textbf{\texttt{explanation}}: This field is intentionally left empty as the source dataset does not provide reasoning traces.
    \item \textbf{\texttt{answer}}: The ground-truth \texttt{answer} is the correct option letter from the original \texttt{answerKey} field.
    \item \textbf{\texttt{category}}: This field is not applicable to the OpenBookQA dataset and is set to "N/A".
\end{itemize}

\subsubsection{QuaRTz (General Science Reasoning)}
The QuaRTz dataset \cite{QuaRTz}, released under the CC BY 4.0 License, is a benchmark for reasoning about textual qualitative relationships. Each question is a 2-way multiple-choice problem that is paired with a background sentence expressing a general qualitative relationship (e.g., "More pollutants mean poorer air quality."). Answering correctly requires applying this general knowledge to a specific, novel situation presented in the question. The raw dataset examples are mapped to our standard schema as follows:
\begin{itemize}
    \item \textbf{\texttt{prompt}}: The input is formatted as a multiple-choice question that includes a context paragraph, the question itself, and two answer choices. The template is as follows:
    \begin{tcolorbox}[title=Prompt Template,fonttitle=\bfseries,fontupper=\small]
Context: \{context\}

Question: \{question\}

Choices:

(A) \{option\_A\}

(B) \{option\_B\}

...

Answer:
    \end{tcolorbox}
    \item \textbf{\texttt{explanation}}: This field is intentionally left empty as the source dataset does not provide reasoning traces.
    \item \textbf{\texttt{answer}}: The ground-truth \texttt{answer} is the correct option letter from the original \texttt{answerKey} field.
    \item \textbf{\texttt{category}}: This field is not applicable to the QuaRTz dataset and is set to "N/A".
\end{itemize}

\subsubsection{ReClor (General Reasoning)}
The ReClor dataset \cite{ReClor}, released under the CC BY-NC 4.0 License, is a reading comprehension benchmark that requires complex logical reasoning. The questions are sourced from standardized tests such as the GMAT and LSAT, ensuring a high level of difficulty and quality. The dataset is designed to evaluate a model's ability to understand arguments, identify flaws, and make inferences, rather than simple text matching. The raw dataset examples are mapped to our standard schema as follows:
\begin{itemize}
    \item \textbf{\texttt{prompt}}: The input is formatted as a multiple-choice question that includes a context paragraph, the question itself, and four answer choices. The template is as follows:
    \begin{tcolorbox}[title=Prompt Template,fonttitle=\bfseries,fontupper=\small]
Context: \{context\}

Question: \{question\}

Choices:

(A) \{option\_A\}

(B) \{option\_B\}

...

Answer:
    \end{tcolorbox}
    \item \textbf{\texttt{explanation}}: This field is intentionally left empty as the source dataset does not provide explicit reasoning traces.
    \item \textbf{\texttt{answer}}: The ground-truth \texttt{answer} is the single letter corresponding to the correct option.
    \item \textbf{\texttt{category}}: This field is not applicable to the ReClor dataset and is set to "N/A".
\end{itemize}

\begin{table*}[t]
\footnotesize
\centering
\caption{Overall dataset distribution for testing. Each dataset is entirely disjoint from the training set and used exclusively for evaluation. Domains indicate the primary reasoning type assessed.}
\label{tab:test_dataset_distribution}
\begin{tabular}{l l r r}
\toprule
\textbf{Dataset} & \textbf{Domain} & \textbf{Test Samples} & \textbf{Test (\%)} \\
\midrule
MATH & Mathematical Reasoning & 5,000 & 9.62 \\
FinQA & Financial Analysis & 1,138 & 2.19 \\
MedMCQA & Medical Diagnosis & 6,150 & 11.84 \\
LegalBench & Legal Reasoning & 21,167 & 40.74 \\
MMLU-Pro & General Understanding & 11,987 & 23.07 \\
BBH & Complex Reasoning & 6,509 & 12.53 \\
\bottomrule
\end{tabular}
\end{table*}

\subsection{Evaluation Datasets}
The final evaluation of the confidence estimation models is performed on a challenging suite of 51,951 samples aggregated from six reasoning-intensive datasets. These datasets, shown in Table~\ref{tab:test_dataset_distribution}, are entirely disjoint from the training set and are used exclusively for testing. The evaluation suite was selected to probe model generalization in both in-domain and out-of-domain settings. While none of the datasets overlap with those used for training, they span the same high-level reasoning domains—mathematical, financial, medical, legal, and complex reasoning—ensuring that evaluation reflects the model’s ability to transfer knowledge. Below, we detail the processing for these evaluation datasets, mapping them to our standard schema.

\subsubsection{MATH (Mathematical Reasoning)}
The MATH dataset \cite{MATH}, released under the MIT License, is a benchmark designed to test mathematical problem-solving ability, consisting of 12,500 problems from high school mathematics competitions. We utilize the full test set of 5,000 records for our evaluation. The problems cover seven subjects, including algebra, geometry, and number theory, and each is accompanied by a full, step-by-step solution in \LaTeX{}. The raw dataset examples are mapped to our standard schema as follows:
\begin{itemize}
    \item \textbf{\texttt{prompt}}: The input question is formatted using the following template:
    \begin{tcolorbox}[title=Prompt Template,fonttitle=\bfseries,fontupper=\small]
Question: \{problem\}

Answer:
    \end{tcolorbox}
    \item \textbf{\texttt{explanation}}: This field is populated with the full, unaltered step-by-step solution from the original \texttt{solution} field of the dataset.
    \item \textbf{\texttt{answer}}: The ground-truth \texttt{answer} is the final result extracted from the \texttt{\textbackslash boxed\{\dots\}} command within the solution text.
    \item \textbf{\texttt{category}}: This field is a composite created by joining the problem \texttt{type} and \texttt{level} from the source data (e.g., ``Prealgebra[SEP]Level 4''). Distributions by type, level, and their combination are reported in Tables~\ref{tab:math_types}, \ref{tab:math_levels}, and~\ref{tab:math_type_level}, respectively.
\end{itemize}

\begin{table}[h]
\footnotesize
\centering
\caption{MATH problem types distribution (Test).}
\label{tab:math_types}
\begin{tabular}{lrr}
\toprule
\textbf{Type} & \textbf{Test Samples} & \textbf{Test (\%)} \\
\midrule
Algebra & 1187 & 23.74 \\
Intermediate Algebra & 903 & 18.06 \\
Prealgebra & 871 & 17.42 \\
Precalculus & 546 & 10.92 \\
Number Theory & 540 & 10.80 \\
Geometry & 479 & 9.58 \\
Counting \& Probability & 474 & 9.48 \\
\bottomrule
\end{tabular}
\end{table}

\begin{table}[h]
\footnotesize
\centering
\caption{MATH problem levels distribution (Test).}
\label{tab:math_levels}
\begin{tabular}{lrr}
\toprule
\textbf{Level} & \textbf{Test Samples} & \textbf{Test (\%)} \\
\midrule
Level 5 & 1324 & 26.48 \\
Level 4 & 1214 & 24.28 \\
Level 3 & 1131 & 22.62 \\
Level 2 & 894  & 17.88 \\
Level 1 & 437  & 8.74 \\
\bottomrule
\end{tabular}
\end{table}

\begin{table*}[t]
\footnotesize
\centering
\caption{MATH combined type--level distribution (Test).}
\label{tab:math_type_level}
\begin{tabular}{llrr}
\toprule
\textbf{Type} & \textbf{Level} & \textbf{Test Samples} & \textbf{Test (\%)} \\
\midrule
Algebra                & Level 5 & 307 & 6.14 \\
Algebra                & Level 4 & 283 & 5.66 \\
Intermediate Algebra   & Level 5 & 280 & 5.60 \\
Algebra                & Level 3 & 261 & 5.22 \\
Intermediate Algebra   & Level 4 & 248 & 4.96 \\
Prealgebra             & Level 3 & 224 & 4.48 \\
Algebra                & Level 2 & 201 & 4.02 \\
Intermediate Algebra   & Level 3 & 195 & 3.90 \\
Prealgebra             & Level 5 & 193 & 3.86 \\
Prealgebra             & Level 4 & 191 & 3.82 \\
Prealgebra             & Level 2 & 177 & 3.54 \\
Number Theory          & Level 5 & 154 & 3.08 \\
Number Theory          & Level 4 & 142 & 2.84 \\
Algebra                & Level 1 & 135 & 2.70 \\
Precalculus            & Level 5 & 135 & 2.70 \\
Geometry               & Level 5 & 132 & 2.64 \\
Intermediate Algebra   & Level 2 & 128 & 2.56 \\
Precalculus            & Level 3 & 127 & 2.54 \\
Geometry               & Level 4 & 125 & 2.50 \\
Counting \& Probability & Level 5 & 123 & 2.46 \\
Number Theory          & Level 3 & 122 & 2.44 \\
Precalculus            & Level 4 & 114 & 2.28 \\
Precalculus            & Level 2 & 113 & 2.26 \\
Counting \& Probability & Level 4 & 111 & 2.22 \\
Geometry               & Level 3 & 102 & 2.04 \\
Counting \& Probability & Level 2 & 101 & 2.02 \\
Counting \& Probability & Level 3 & 100 & 2.00 \\
Number Theory          & Level 2 & 92  & 1.84 \\
Prealgebra             & Level 1 & 86  & 1.72 \\
Geometry               & Level 2 & 82  & 1.64 \\
Precalculus            & Level 1 & 57  & 1.14 \\
Intermediate Algebra   & Level 1 & 52  & 1.04 \\
Counting \& Probability & Level 1 & 39  & 0.78 \\
Geometry               & Level 1 & 38  & 0.76 \\
Number Theory          & Level 1 & 30  & 0.60 \\
\bottomrule
\end{tabular}
\end{table*}

\subsubsection{FinQA (Financial Reasoning)}
The FinQA dataset \cite{FinQA}, released under the MIT License, is a large-scale benchmark designed to test numerical reasoning over financial reports. We utilize the full test set of 1,147 records for our evaluation. Each example consists of a question authored by a financial expert, along with both unstructured text and a structured table from S\&P 500 company earnings reports. Answering questions correctly requires synthesizing information from both data types and performing complex, multi-step calculations. The raw dataset examples are mapped to our standard schema as follows:
\begin{itemize}
    \item \textbf{\texttt{prompt}}: A comprehensive prompt is constructed for each question. First, a context is created by verbalizing the structured table into natural language. This process follows the methodology of the original FinQA paper, where each row is converted into a descriptive sentence using a template similar to \textit{`the \{column name\} of \{row name\} is \{cell value\};`}. This verbalized table text is then concatenated with the original paragraphs to form a complete context, which is then embedded into the same prompt template used for the TAT-QA dataset.
    \item \textbf{\texttt{explanation}}: This field is populated with the executable reasoning program (e.g., "divide(100, 100), divide(3.8, \#0)") from the source dataset, which provides a fully explainable, step-by-step reasoning trace.
    \item \textbf{\texttt{answer}}: The ground-truth \texttt{answer} is the final numerical result from the source dataset.
    \item \textbf{\texttt{category}}: This field indicates whether the necessary information to answer the question was found in the text, the table, or both (e.g., ``text\_retrieved-table\_retrieved''). The category distribution is shown in Table~\ref{tab:finqa_categories}.
\end{itemize}

\begin{table}[h]
\footnotesize
\centering
\caption{FinQA categories distribution (Test).}
\label{tab:finqa_categories}
\begin{tabular}{lrr}
\toprule
\textbf{Category} & \textbf{Test Samples} & \textbf{Test (\%)} \\
\midrule
table\_retrieved & 657 & 57.73 \\
text\_retrieved & 272 & 23.90 \\
text\_retrieved-table\_retrieved & 125 & 10.98 \\
N/A & 84 & 7.38 \\
\bottomrule
\end{tabular}
\end{table}

\subsubsection{MedMCQA (Medical Diagnosis)}
The MedMCQA dataset \cite{MedMCQA}, released under the Apache License 2.0, is a large-scale, multiple-choice question answering benchmark containing over 194,000 questions from Indian medical entrance exams (AIIMS \& NEET PG). The dataset is designed to test deep medical knowledge and reasoning across 21 different subjects. We utilize the full test set of 6,150 records for our evaluation. The raw dataset examples are mapped to our standard schema as follows:
\begin{itemize}
    \item \textbf{\texttt{prompt}}: Each example is formatted into a standardized multiple-choice prompt. The four options are presented in their original order. The template is as follows:
    \begin{tcolorbox}[title=Prompt Template,fonttitle=\bfseries,fontupper=\small]
Question: \{question\}

Choices:

(A) \{option\_A\}

(B) \{option\_B\}

(C) \{option\_C\}

(D) \{option\_D\}

Answer:
    \end{tcolorbox}
    \item \textbf{\texttt{explanation}}: This field is populated with the expert's explanation for the correct answer, taken directly from the \texttt{exp} field of the source dataset.
    \item \textbf{\texttt{answer}}: The ground-truth \texttt{answer} is the single letter corresponding to the correct option, derived from the \texttt{cop} (correct option) field.
    \item \textbf{\texttt{category}}: This field is populated with the \texttt{subject\_name} from the source data (e.g., ``Pathology'', ``Anatomy''). The category distribution is shown in Table~\ref{tab:medmcqa_categories}.
\end{itemize}

\begin{table}[t]
\footnotesize
\centering
\caption{MedMCQA categories distribution (Test).}
\label{tab:medmcqa_categories}
\begin{tabular}{lrr}
\toprule
\textbf{Category} & \textbf{Test Samples} & \textbf{Test (\%)} \\
\midrule
Dental & 1203 & 19.56 \\
Unknown & 682 & 11.09 \\
Gynaecology \& Obstetrics & 532 & 8.65 \\
Surgery & 501 & 8.15 \\
Physiology & 388 & 6.31 \\
Medicine & 372 & 6.05 \\
Biochemistry & 352 & 5.72 \\
Pharmacology & 317 & 5.15 \\
Pathology & 305 & 4.96 \\
Anatomy & 259 & 4.21 \\
Social \& Preventive Medicine & 243 & 3.95 \\
Pediatrics & 190 & 3.09 \\
Ophthalmology & 177 & 2.88 \\
Microbiology & 167 & 2.72 \\
Forensic Medicine & 132 & 2.15 \\
Radiology & 119 & 1.93 \\
ENT & 86 & 1.40 \\
Skin & 60 & 0.98 \\
Anaesthesia & 59 & 0.96 \\
Psychiatry & 6 & 0.10 \\
\bottomrule
\end{tabular}
\end{table}

\subsubsection{LegalBench (Legal Reasoning)}
The LegalBench dataset \cite{LegalBench} is a comprehensive benchmark consisting of 162 tasks designed to evaluate legal reasoning. Given the scale of the benchmark, we curated a representative subset from its test data for a more controlled evaluation. To ensure a balanced and fair selection, we implemented a deterministic, multi-stage sampling algorithm (\texttt{seed=23}). First, tasks were stratified into the six core legal reasoning types defined by the original authors (e.g., Issue-spotting, Rule-application, Interpretation). Then, from each reasoning category, we randomly sampled a quarter of the available tasks, up to a maximum of five, to form our evaluation suite. This process resulted in a final test set of 15 distinct tasks, comprising 21,167 test records.

The raw dataset examples are mapped to our standard schema as follows:
\begin{itemize}
    \item \textbf{\texttt{prompt}}: Constructed using templates adapted from the official implementation provided for each task in LegalBench. The templates for the selected tasks are summarized in Table~\ref{tab:legalbench_prompts}.
    \item \textbf{\texttt{explanation}}: Intentionally left empty, as the source dataset provides final answers but not step-by-step reasoning traces.
    \item \textbf{\texttt{answer}}: The ground-truth \texttt{answer} is taken directly from the source data for each task.
    \item \textbf{\texttt{category}}: Populated with the task identifier from the LegalBench dataset (e.g., ``international\_citizenship\_questions'', ``opp115\_first\_party\_collection\_use''). The category distribution is shown in Table~\ref{tab:legalbench_categories}.
\end{itemize}

\onecolumn
\pagebreak
\footnotesize
\begin{longtable}{p{0.5\textwidth} p{0.15\textwidth} p{0.2\textwidth}}
\caption{Prompt templates for the selected LegalBench tasks used in our evaluation suite.}
\label{tab:legalbench_prompts}\\
\toprule
\footnotesize\textbf{Task / Prompt} & \footnotesize\textbf{Records} & \footnotesize\textbf{License} \\
\midrule
\endfirsthead
\multicolumn{3}{l}{\scriptsize\emph{(Continued from previous page)}}\\
\toprule
\footnotesize\textbf{Task / Prompt} & \footnotesize\textbf{Records} & \footnotesize\textbf{License} \\
\midrule
\endhead
\midrule
\multicolumn{3}{r}{\scriptsize\emph{(Continued on next page)}}\\
\endfoot
\bottomrule
\endlastfoot

\multicolumn{3}{l}{\footnotesize\textbf{- International citizenship questions}} \\
Prompt: Answer the following questions considering the state of international law on January 1st, 2020.

Question: \{\{question\}\} Answer ``Yes'' or ``No''.

Answer: & 9{,}306 & CC BY 4.0 \\
\addlinespace

\multicolumn{3}{l}{\footnotesize\textbf{- Learned hands housing}} \\
Prompt: Does the post discuss issues with paying your rent or mortgage, landlord-tenant issues, housing subsidies and public housing, eviction, and other problems with your apartment, mobile home, or house? Answer ``Yes'' or ``no''.

Post: \{text\}

Label: & 4{,}494 & CC BY-NC-SA 4.0 \\
\addlinespace

\multicolumn{3}{l}{\footnotesize\textbf{- Opp115 first party collection use}} \\
Prompt: Does the clause describe how and why a service provider collects user information? Answer ``Yes'' or ``no''.

Clause: \{text\}

Label: & 2{,}086 & CC BY-NC \\
\addlinespace

\multicolumn{3}{l}{\footnotesize\textbf{- Cuad license grant}} \\
Prompt: Does the clause contain a license granted by one party to its counterparty? Answer ``Yes'' or ``no''.

Clause: \{text\}

Label: & 1{,}396 & CC BY 4.0 \\
\addlinespace

\multicolumn{3}{l}{\footnotesize\textbf{- Definition classification}} \\
Prompt: Identify if the sentence defines a term. Answer ``Yes'' or ``no''.

Sentence: \{text\}

Label: & 1{,}337 & CC BY-SA 4.0 \\
\addlinespace

\multicolumn{3}{l}{\footnotesize\textbf{- Opp115 international and specific audiences}} \\
Prompt: Does the clause describe practices that pertain only to a specific group of users (e.g., children, Europeans, or California residents)? Answer ``Yes'' or ``no''.

Clause: \{text\}

Label: & 980 & CC BY-NC \\
\addlinespace

\multicolumn{3}{l}{\footnotesize\textbf{- Learned hands torts}} \\
Prompt: Does the post discuss problems that one person has with another person (or animal), like when there is a car accident, a dog bite, bullying or possible harassment, or neighbors treating each other badly? Answer ``Yes'' or ``no''.

Post: \{text\}

Label: & 432 & CC BY-NC-SA 4.0 \\
\addlinespace

\multicolumn{3}{l}{\footnotesize\textbf{- Diversity 5}} \\
Prompt: Diversity jurisdiction exists when there is (1) complete diversity between plaintiffs and defendants, and (2) the amount-in-controversy (AiC) is greater than \$75k.

Q: \{text\} Is there diversity jurisdiction?

A: & 300 & CC BY 4.0 \\
\addlinespace

\multicolumn{3}{l}{\footnotesize\textbf{- Diversity 6}} \\
Prompt: Diversity jurisdiction exists when there is (1) complete diversity between plaintiffs and defendants, and (2) the amount-in-controversy (AiC) is greater than \$75k.

Q: \{text\} Is there diversity jurisdiction?

A: & 300 & CC BY 4.0 \\
\addlinespace

\multicolumn{3}{l}{\footnotesize\textbf{- Learned hands domestic violence}} \\
Prompt: Does the post discuss dealing with domestic violence and abuse, including getting protective orders, enforcing them, understanding abuse, reporting abuse, and getting resources and status if there is abuse? Answer ``Yes'' or ``no''.

Post: \{text\}

Label: & 174 & CC BY-NC-SA 4.0 \\
\addlinespace

\multicolumn{3}{l}{\footnotesize\textbf{- UCC v common law}} \\
Prompt: The UCC (through Article 2) governs the sale of goods, which are defined as moveable tangible things (cars, apples, books, etc.), whereas the common law governs contracts for real estate and services. For the following contracts, determine if they are governed by the UCC or by common law.

Contract: \{contract\} Is this contract governed by the UCC or the common law?

Governed by: & 94 & CC BY 4.0 \\
\addlinespace

\multicolumn{3}{l}{\footnotesize\textbf{- Maud cor standard (intervening event)}} \\
Prompt: Instruction: Read the segment of a merger agreement and answer the multiple-choice question by choosing the option that best characterizes the agreement.

Question: What standard should the board follow when determining whether to change its recommendation in response to an intervening event?

Option A: ``Breach'' of fiduciary duties

...

Option I: Other specified standard

Merger Agreement: \{text\}

Answer: & 84 & CC BY 4.0 \\
\addlinespace

\multicolumn{3}{l}{\footnotesize\textbf{- Cuad third party beneficiary}} \\
Prompt: Does the clause specify that there is a non-contracting party who is a beneficiary to some or all of the clauses in the contract and therefore can enforce its rights against a contracting party? Answer ``Yes'' or ``no''.

Clause: \{text\}

Label: & 68 & CC BY 4.0 \\
\addlinespace

\multicolumn{3}{l}{\footnotesize\textbf{- Learned hands benefits}} \\
Prompt: Does the post discuss public benefits and social services that people can get from the government, like for food, disability, old age, housing, medical help, unemployment, child care, or other social needs? Answer ``Yes'' or ``no''.

Post: \{text\}

Label: & 66 & CC BY-NC-SA 4.0 \\
\addlinespace

\multicolumn{3}{l}{\footnotesize\textbf{- Personal jurisdiction}} \\
Prompt: There is personal jurisdiction over a defendant in the state where the defendant is domiciled, or when (1) the defendant has sufficient contacts with the state, such that they have availed itself of the privileges of the state and (2) the claim arises out of the nexus of the defendant's contacts with the state.

Q: \{text\} Is there personal jurisdiction?

A: & 50 & CC BY 4.0 \\
\end{longtable}
\twocolumn

\begin{table*}[t]
\footnotesize
\centering
\caption{LegalBench categories distribution (Test).}
\label{tab:legalbench_categories}
\begin{tabular}{lrr}
\toprule
\textbf{Category} & \textbf{Test Samples} & \textbf{Test (\%)} \\
\midrule
international\_citizenship\_questions & 9306 & 43.96 \\
learned\_hands\_housing & 4494 & 21.23 \\
opp115\_first\_party\_collection\_use & 2086 & 9.85 \\
cuad\_license\_grant & 1396 & 6.60 \\
definition\_classification & 1337 & 6.32 \\
opp115\_international\_and\_specific\_audiences & 980 & 4.63 \\
learned\_hands\_torts & 432 & 2.04 \\
diversity\_5 & 300 & 1.42 \\
diversity\_6 & 300 & 1.42 \\
learned\_hands\_domestic\_violence & 174 & 0.82 \\
ucc\_v\_common\_law & 94 & 0.44 \\
maud\_cor\_standard\_(intervening\_event) & 84 & 0.40 \\
cuad\_third\_party\_beneficiary & 68 & 0.32 \\
learned\_hands\_benefits & 66 & 0.31 \\
personal\_jurisdiction & 50 & 0.24 \\
\bottomrule
\end{tabular}
\end{table*}
\normalsize
\subsubsection{MMLU-PRO (General Understanding)}
The MMLU-Pro dataset \cite{MMLUPRO}, released under the Apache-2.0 License, is a more challenging version of the MMLU benchmark, designed to elevate the assessment of multi-task language understanding by incorporating more complex, reasoning-intensive questions across 14 challenging tasks. We utilize the full test set of 12,032 records for our evaluation. The raw dataset examples are mapped to our standard schema as follows:
\begin{itemize}
    \item \textbf{\texttt{prompt}}: Each question is formatted into a standard multiple-choice prompt, with the choices presented in their original order. The template is as follows:
    \begin{tcolorbox}[title=Prompt Template,fonttitle=\bfseries,fontupper=\small]
Question: \{question\}

Choices:

(A) \{option\_A\}

(B) \{option\_B\}

...

Answer:
    \end{tcolorbox}
    \item \textbf{\texttt{explanation}}: This field is populated with the \texttt{cot\_content} from the source dataset, which provides a ground-truth chain-of-thought reasoning trace.
    \item \textbf{\texttt{answer}}: The ground-truth \texttt{answer} is the single letter corresponding to the correct option.
    \item \textbf{\texttt{category}}: This field is populated with the \texttt{category} field from the source data (e.g., ``math'', ``physics''). The distribution of categories in the test set is shown in Table~\ref{tab:mmlupro_categories}.
\end{itemize}

\begin{table*}[t]
\footnotesize
\centering
\caption{MMLU-Pro categories distribution (Test).}
\label{tab:mmlupro_categories}
\begin{tabular}{lrr}
\toprule
\textbf{Category} & \textbf{Test Samples} & \textbf{Test (\%)} \\
\midrule
math               & 1351 & 11.27 \\
physics            & 1299 & 10.84 \\
chemistry          & 1132 & 9.44 \\
law                & 1080 & 9.01 \\
engineering        & 969  & 8.08 \\
other              & 924  & 7.71 \\
economics          & 843  & 7.03 \\
health             & 807  & 6.73 \\
psychology         & 798  & 6.66 \\
business           & 788  & 6.57 \\
biology            & 707  & 5.90 \\
philosophy         & 499  & 4.16 \\
computer science   & 409  & 3.41 \\
history            & 381  & 3.18 \\
\bottomrule
\end{tabular}
\end{table*}

\subsubsection{Big-Bench Hard (Complex Reasoning)}
The Big-Bench Hard (BBH) dataset \cite{BBH}, released under the MIT License, is a suite of 27 challenging tasks designed to be beyond the capabilities of contemporary language models. The tasks are diverse, covering areas such as logical deduction, causal judgment, and tracking shuffled objects. We utilize the full test set, combining all 27 sub-tasks for a total of 6,511 records in our evaluation. The raw dataset examples are mapped to our standard schema as follows:
\begin{itemize}
    \item \textbf{\texttt{prompt}}: The input question is formatted using the following simple template:
    \begin{tcolorbox}[title=Prompt Template,fonttitle=\bfseries,fontupper=\small]
Question: \{input\}

Answer:
    \end{tcolorbox}
    \item \textbf{\texttt{explanation}}: This field is intentionally left empty as the source dataset does not provide reasoning traces.
    \item \textbf{\texttt{answer}}: The ground-truth \texttt{answer} is the value from the original \texttt{target} field.
    \item \textbf{\texttt{category}}: This field contains the name of the specific BBH sub-task (e.g., ``boolean\_expressions'', ``causal\_judgement''). The category distribution is shown in Table~\ref{tab:bbh_categories}.
\end{itemize}

\begin{table*}[t]
\footnotesize
\centering
\caption{Big-Bench Hard (BBH) categories distribution (Test).}
\label{tab:bbh_categories}
\begin{tabular}{lrr}
\toprule
\textbf{Category} & \textbf{Test Samples} & \textbf{Test (\%)} \\
\midrule
boolean\_expressions & 250 & 3.84 \\
multistep\_arithmetic\_two & 250 & 3.84 \\
web\_of\_lies & 250 & 3.84 \\
tracking\_shuffled\_objects\_three\_objects & 250 & 3.84 \\
tracking\_shuffled\_objects\_seven\_objects & 250 & 3.84 \\
tracking\_shuffled\_objects\_five\_objects & 250 & 3.84 \\
temporal\_sequences & 250 & 3.84 \\
salient\_translation\_error\_detection & 250 & 3.84 \\
ruin\_names & 250 & 3.84 \\
reasoning\_about\_colored\_objects & 250 & 3.84 \\
object\_counting & 250 & 3.84 \\
navigate & 250 & 3.84 \\
movie\_recommendation & 250 & 3.84 \\
logical\_deduction\_three\_objects & 250 & 3.84 \\
logical\_deduction\_seven\_objects & 250 & 3.84 \\
logical\_deduction\_five\_objects & 250 & 3.84 \\
hyperbaton & 250 & 3.84 \\
geometric\_shapes & 250 & 3.84 \\
formal\_fallacies & 250 & 3.84 \\
dyck\_languages & 250 & 3.84 \\
disambiguation\_qa & 250 & 3.84 \\
date\_understanding & 250 & 3.84 \\
word\_sorting & 250 & 3.84 \\
sports\_understanding & 248 & 3.81 \\
causal\_judgement & 187 & 2.87 \\
snarks & 178 & 2.73 \\
penguins\_in\_a\_table & 146 & 2.24 \\
\bottomrule
\end{tabular}
\end{table*}

\begin{table*}[h!]
\footnotesize
\centering
\small
\caption{Reasoning-path keywords used to detect reconsideration, verification, or alternate exploration behaviors in generated responses.}
\label{tab:reasoning_keywords}
\begin{tabular}{p{0.25\textwidth}|p{0.65\textwidth}}
\toprule
\textbf{Category} & \textbf{Keywords} \\
\midrule
Verification & 
\begin{minipage}[t]{\linewidth}
wait, double-check, make sure, verify, to confirm, let me verify, let me double-check, let me confirm
\end{minipage} \\
\midrule
Alternative Approach & 
\begin{minipage}[t]{\linewidth}
alternatively, another way, another approach, different approach
\end{minipage} \\
\midrule
Reconsideration & 
\begin{minipage}[t]{\linewidth}
but let me, let me try, on second thought, let me reconsider, let me check, hold on, wait a minute, let me think again, but what if
\end{minipage} \\
\bottomrule
\end{tabular}
\end{table*}

\subsection{Response Generation}
\label{app:response_generation}
All model outputs were generated using vLLM \citep{vllm} servers hosted locally and accessed through the OpenAI-compatible API interface. Each reasoning prompt was submitted as a user message, and the model’s completion was recorded as the response. To ensure deterministic and reproducible outputs, inference was performed with a temperature of 0.0 and a maximum generation length of 4096 tokens. To mitigate repetitive or looping text generation, which some reasoning models can exhibit even under deterministic decoding, we applied model-specific penalties following recommendations from their original releases. Specifically, we set the frequency penalty to 1.5 for \texttt{Magistral-Small-2506} and the frequency penalty to 0.8 with a presence penalty of 1.5 for \texttt{Phi-4-mini-flash-reasoning}. All responses were produced in batch mode via parallel threaded API calls to the vLLM endpoint, using a compute setup with 8 NVIDIA H200 GPUs, enabling efficient large-scale inference.

\subsection{Response Segmentation}
\label{app:response_segmentation}
To analyze the step-by-step reasoning, each generated \texttt{model\_response} was segmented into coherent units of thought, or ``chunks.'' This process follows the methodology of \citet{PHSV}, which first splits the reasoning trace into paragraphs and then groups them based on a set of keywords that signal a new reasoning path. Based on our development experiments, we found that an expanded set of keywords provided more robust segmentation across the diverse model families in our study. Each chunk is designed to represent a single, coherent line of reasoning that often ends with an intermediate conclusion (see Table~\ref{tab:reasoning_keywords}).

\begin{table*}[h!]
\footnotesize
\centering
\small
\caption{System and user prompts employed for GPT-5-nano to assign correctness labels to individual model responses.}
\label{tab:grading_prompts}
\begin{tabular}{p{0.15\textwidth}|p{0.75\textwidth}}
\toprule
\textbf{Prompt Type} & \textbf{Content} \\
\midrule
System Prompt &  
\begin{minipage}[t]{\linewidth}
You are a meticulous grading assistant. A teacher has asked a student a question, and the student provided a step-by-step answer as a series of 'chunks'. Your task is to assist the teacher by evaluating each chunk of the student's reasoning and provide an overall assessment. You must follow the instructions precisely and provide your output only in the specified XML format.
\end{minipage} \\
\midrule
User Prompt &  
\begin{minipage}[t]{\linewidth}
\textbf{\#\#\# Instruction}

For each reasoning chunk from the student, evaluate whether its intermediate result exactly matches the Final Ground-Truth Answer. Mark each chunk with:  

- 1 if the chunk's intermediate result matches the ground-truth answer.  

- 0 if the chunk's intermediate result does not match the ground-truth answer.  

- null if the chunk does not contain any intermediate result (e.g., pure reflection/setup).  

After grading each chunk, provide a final grade that evaluates whether the model's final answer/conclusion matches the ground truth:  

- 1 if the final answer/conclusion matches the ground truth.  

- 0 if the final answer/conclusion does not match the ground truth.  

Your output must be a series of chunk evaluations in XML format, followed by a final grade:  

$<$chunk id="1"$>$0/1/null$<$/chunk$>$  

$<$chunk id="2"$>$0/1/null$<$/chunk$>$  

...  

$<$final\_grade$>$0/1$<$/final\_grade$>$  

---  

\textbf{\#\#\# Context}  

* Question: "\{prompt\}"  

* Final Ground-Truth Answer: "\{answer\}"  

---  

\textbf{\#\#\# Task: Grade Each Chunk}  

\{reasoning\_chunks\}
\end{minipage} \\
\bottomrule
\end{tabular}
\end{table*}

\subsection{Response Grading}
\label{app:response_grading}
Each generated response was subsequently labeled as correct or incorrect. For multiple-choice questions, correctness was determined by a straightforward string match. For open-ended responses, we employed a state-of-the-art language model as a judge to assess semantic equivalence, a practice that has been validated and widely adopted in recent literature \cite{calibration-tuning, PHSV, emit, CCPS}. The reliability of using powerful LLMs for the constrained task of comparing a generated answer to a ground-truth answer has been demonstrated by \citet{calibration-tuning}, who found that GPT-4's judgments closely align with human assessments (4.5\% average difference). Building upon their findings, we utilized the more recent and capable \texttt{GPT-5-nano} model to ensure the highest quality labels. This approach does not use the grader as a knowledge oracle; rather, it provides the grader with the question, the ground-truth answer, and the model's generated answer, asking only if the two are semantically equivalent.

To provide a more granular analysis, we applied this grading on a per-chunk basis, similar to the approach in \citet{PHSV}. This allows us to obtain a correctness label for each intermediate step in the reasoning process. The prompt used for this per-chunk grading task, which is an adjusted version of the one used in prior work, is detailed in Table~\ref{tab:grading_prompts}. The total cost for this automated labeling process---calculated across all training and testing datasets and all LRMs---was approximately \$746. The final sample- and chunk-level statistics for each LRM after grading are provided in Table~\ref{tab:train_merged_llm} for the train split and Table~\ref{tab:test_merged_llm} for the test split.

\clearpage
\onecolumn
\begingroup
\scriptsize
\begin{longtable}{l r r r r r r r r}
\caption{A breakdown of sample- and chunk-level train statistics across various datasets, aggregated by LRMs. \textbf{Notation:} $S$ = number of samples (problems); $C$ = chunks per sample; $p$ = proportion; labels $L \in \{\checkmark,\text{\ding{55}},\varnothing\}$ denote correct, incorrect, and no-result. \textbf{Columns:} $S$; $\checkmark$ = correct samples (\#S, $p$); \ding{55} = incorrect samples (\#S, $p$); $C(\mu\pm\sigma)$ = mean$\pm$sd chunks per sample; $\sum C$ = total chunks; $\checkmark$ = correct chunks (\#C, $p$); \ding{55} = incorrect chunks (\#C, $p$); $\varnothing$ = no-result chunks (\#C, $p$). Proportions $p$ are typeset tiny as ($.xx$).}\label{tab:train_merged_llm}\\
\toprule
\footnotesize\textbf{Dataset} & \footnotesize\textbf{$S$} & \footnotesize\textbf{$\checkmark$ (\#\!$S$, $p$)} & \footnotesize\textbf{\ding{55} (\#\!$S$, $p$)} & \footnotesize\textbf{$C$ ($\mu\!\pm\!\sigma$)} & \footnotesize\textbf{$\sum C$} & \footnotesize\textbf{$\checkmark$ (\#\!$C$, $p$)} & \footnotesize\textbf{\ding{55} (\#\!$C$, $p$)} & \footnotesize\textbf{$\varnothing$ (\#\!$C$, $p$)}\\
\midrule

\endfirsthead
\multicolumn{9}{l}{\scriptsize\emph{(Continued from previous page)}}\\
\toprule
\footnotesize\textbf{Dataset} & \footnotesize\textbf{$S$} & \footnotesize\textbf{$\checkmark$ (\#\!$S$, $p$)} & \footnotesize\textbf{\ding{55} (\#\!$S$, $p$)} & \footnotesize\textbf{$C$ ($\mu\!\pm\!\sigma$)} & \footnotesize\textbf{$\sum C$} & \footnotesize\textbf{$\checkmark$ (\#\!$C$, $p$)} & \footnotesize\textbf{\ding{55} (\#\!$C$, $p$)} & \footnotesize\textbf{$\varnothing$ (\#\!$C$, $p$)}\\
\midrule

\endhead
\midrule
\multicolumn{9}{r}{\scriptsize\emph{(Continued on next page)}}\\

\endfoot
\bottomrule
\endlastfoot
\multicolumn{9}{c}{\footnotesize\textbf{\texttt{microsoft-Phi-4-mini-flash-reasoning}}} \\
\addlinespace[2pt]
GSM8K & 849 & 824 {\scriptsize ($.97$)} & 25 {\scriptsize ($.03$)} & 6.40$\pm$6.67 & 5,436 & 2,668 {\scriptsize ($.49$)} & 1,185 {\scriptsize ($.22$)} & 1,583 {\scriptsize ($.29$)} \\
TAT-QA & 747 & 371 {\scriptsize ($.50$)} & 376 {\scriptsize ($.50$)} & 9.55$\pm$11.34 & 7,137 & 1,300 {\scriptsize ($.18$)} & 3,476 {\scriptsize ($.49$)} & 2,361 {\scriptsize ($.33$)} \\
MedQA & 791 & 252 {\scriptsize ($.32$)} & 539 {\scriptsize ($.68$)} & 3.73$\pm$9.31 & 2,948 & 415 {\scriptsize ($.14$)} & 1,671 {\scriptsize ($.57$)} & 862 {\scriptsize ($.29$)} \\
LEXam & 562 & 144 {\scriptsize ($.26$)} & 418 {\scriptsize ($.74$)} & 10.25$\pm$10.53 & 5,761 & 435 {\scriptsize ($.08$)} & 3,902 {\scriptsize ($.68$)} & 1,424 {\scriptsize ($.25$)} \\
ARC & 811 & 661 {\scriptsize ($.81$)} & 150 {\scriptsize ($.18$)} & 4.95$\pm$11.16 & 4,016 & 1,296 {\scriptsize ($.32$)} & 1,629 {\scriptsize ($.41$)} & 1,091 {\scriptsize ($.27$)} \\
CommonsenseQA2 & 814 & 569 {\scriptsize ($.70$)} & 245 {\scriptsize ($.30$)} & 8.58$\pm$11.68 & 6,985 & 1,511 {\scriptsize ($.22$)} & 2,311 {\scriptsize ($.33$)} & 3,163 {\scriptsize ($.45$)} \\
LogiQA & 760 & 338 {\scriptsize ($.44$)} & 422 {\scriptsize ($.56$)} & 3.23$\pm$7.29 & 2,453 & 484 {\scriptsize ($.20$)} & 1,330 {\scriptsize ($.54$)} & 639 {\scriptsize ($.26$)} \\
OpenBookQA & 790 & 565 {\scriptsize ($.72$)} & 225 {\scriptsize ($.28$)} & 13.17$\pm$21.77 & 10,401 & 2,072 {\scriptsize ($.20$)} & 4,109 {\scriptsize ($.40$)} & 4,220 {\scriptsize ($.41$)} \\
QuaRTz & 815 & 644 {\scriptsize ($.79$)} & 171 {\scriptsize ($.21$)} & 3.07$\pm$6.74 & 2,500 & 1,128 {\scriptsize ($.45$)} & 747 {\scriptsize ($.30$)} & 625 {\scriptsize ($.25$)} \\
ReClor & 800 & 452 {\scriptsize ($.56$)} & 348 {\scriptsize ($.43$)} & 2.36$\pm$5.87 & 1,885 & 584 {\scriptsize ($.31$)} & 937 {\scriptsize ($.50$)} & 364 {\scriptsize ($.19$)} \\
\midrule
\textbf{TOTAL} & \textbf{7,739} & \textbf{4,820} {\scriptsize ($.62$)} & \textbf{2,919} {\scriptsize ($.38$)} & \textbf{6.40$\pm$11.64} & \textbf{49,522} & \textbf{11,893} {\scriptsize ($.24$)} & \textbf{21,297} {\scriptsize ($.43$)} & \textbf{16,332} {\scriptsize ($.33$)} \\
\addlinespace[4pt]
\midrule
\multicolumn{9}{c}{\footnotesize\textbf{\texttt{Qwen-Qwen3-8B}}} \\
\addlinespace[2pt]
GSM8K & 956 & 932 {\scriptsize ($.97$)} & 24 {\scriptsize ($.03$)} & 12.49$\pm$17.23 & 11,939 & 5,461 {\scriptsize ($.46$)} & 2,322 {\scriptsize ($.19$)} & 4,156 {\scriptsize ($.35$)} \\
TAT-QA & 936 & 619 {\scriptsize ($.66$)} & 317 {\scriptsize ($.34$)} & 6.48$\pm$6.80 & 6,067 & 1,935 {\scriptsize ($.32$)} & 2,454 {\scriptsize ($.40$)} & 1,678 {\scriptsize ($.28$)} \\
MedQA & 927 & 720 {\scriptsize ($.78$)} & 207 {\scriptsize ($.22$)} & 20.42$\pm$20.04 & 18,928 & 4,409 {\scriptsize ($.23$)} & 7,427 {\scriptsize ($.39$)} & 7,092 {\scriptsize ($.37$)} \\
LEXam & 639 & 295 {\scriptsize ($.46$)} & 344 {\scriptsize ($.54$)} & 23.53$\pm$19.38 & 15,033 & 1,830 {\scriptsize ($.12$)} & 9,429 {\scriptsize ($.63$)} & 3,774 {\scriptsize ($.25$)} \\
ARC & 984 & 945 {\scriptsize ($.96$)} & 39 {\scriptsize ($.04$)} & 8.03$\pm$10.75 & 7,904 & 3,406 {\scriptsize ($.43$)} & 2,373 {\scriptsize ($.30$)} & 2,125 {\scriptsize ($.27$)} \\
CommonsenseQA2 & 988 & 802 {\scriptsize ($.81$)} & 186 {\scriptsize ($.19$)} & 6.77$\pm$6.04 & 6,692 & 2,018 {\scriptsize ($.30$)} & 1,726 {\scriptsize ($.26$)} & 2,948 {\scriptsize ($.44$)} \\
LogiQA & 906 & 683 {\scriptsize ($.75$)} & 223 {\scriptsize ($.25$)} & 13.73$\pm$13.87 & 12,441 & 2,610 {\scriptsize ($.21$)} & 5,514 {\scriptsize ($.44$)} & 4,317 {\scriptsize ($.35$)} \\
OpenBookQA & 983 & 886 {\scriptsize ($.90$)} & 97 {\scriptsize ($.10$)} & 10.18$\pm$10.99 & 10,007 & 3,106 {\scriptsize ($.31$)} & 3,716 {\scriptsize ($.37$)} & 3,185 {\scriptsize ($.32$)} \\
QuaRTz & 990 & 922 {\scriptsize ($.93$)} & 68 {\scriptsize ($.07$)} & 5.30$\pm$6.12 & 5,249 & 2,609 {\scriptsize ($.50$)} & 1,140 {\scriptsize ($.22$)} & 1,500 {\scriptsize ($.29$)} \\
ReClor & 918 & 858 {\scriptsize ($.93$)} & 60 {\scriptsize ($.07$)} & 13.24$\pm$13.78 & 12,156 & 3,821 {\scriptsize ($.31$)} & 4,690 {\scriptsize ($.39$)} & 3,645 {\scriptsize ($.30$)} \\
\midrule
\textbf{TOTAL} & \textbf{9,227} & \textbf{7,662} {\scriptsize ($.83$)} & \textbf{1,565} {\scriptsize ($.17$)} & \textbf{11.53$\pm$14.21} & \textbf{106,416} & \textbf{31,205} {\scriptsize ($.29$)} & \textbf{40,791} {\scriptsize ($.38$)} & \textbf{34,420} {\scriptsize ($.32$)} \\
\addlinespace[4pt]
\midrule
\multicolumn{9}{c}{\footnotesize\textbf{\texttt{Qwen-Qwen3-14B}}} \\
\addlinespace[2pt]
GSM8K & 978 & 948 {\scriptsize ($.97$)} & 30 {\scriptsize ($.03$)} & 8.80$\pm$7.38 & 8,605 & 4,219 {\scriptsize ($.49$)} & 1,668 {\scriptsize ($.19$)} & 2,718 {\scriptsize ($.32$)} \\
TAT-QA & 911 & 599 {\scriptsize ($.66$)} & 312 {\scriptsize ($.34$)} & 4.15$\pm$3.87 & 3,783 & 1,326 {\scriptsize ($.35$)} & 1,528 {\scriptsize ($.40$)} & 929 {\scriptsize ($.25$)} \\
MedQA & 955 & 790 {\scriptsize ($.83$)} & 165 {\scriptsize ($.17$)} & 10.05$\pm$10.12 & 9,594 & 2,786 {\scriptsize ($.29$)} & 3,235 {\scriptsize ($.34$)} & 3,573 {\scriptsize ($.37$)} \\
LEXam & 687 & 313 {\scriptsize ($.46$)} & 374 {\scriptsize ($.54$)} & 13.94$\pm$14.89 & 9,580 & 1,117 {\scriptsize ($.12$)} & 6,592 {\scriptsize ($.69$)} & 1,871 {\scriptsize ($.20$)} \\
ARC & 976 & 932 {\scriptsize ($.95$)} & 44 {\scriptsize ($.05$)} & 4.33$\pm$5.76 & 4,225 & 1,958 {\scriptsize ($.46$)} & 1,339 {\scriptsize ($.32$)} & 928 {\scriptsize ($.22$)} \\
CommonsenseQA2 & 986 & 781 {\scriptsize ($.79$)} & 205 {\scriptsize ($.21$)} & 4.42$\pm$4.41 & 4,362 & 1,426 {\scriptsize ($.33$)} & 1,295 {\scriptsize ($.30$)} & 1,641 {\scriptsize ($.38$)} \\
LogiQA & 895 & 664 {\scriptsize ($.74$)} & 231 {\scriptsize ($.26$)} & 7.59$\pm$8.49 & 6,794 & 1,584 {\scriptsize ($.23$)} & 3,088 {\scriptsize ($.45$)} & 2,122 {\scriptsize ($.31$)} \\
OpenBookQA & 975 & 893 {\scriptsize ($.92$)} & 82 {\scriptsize ($.08$)} & 5.22$\pm$5.86 & 5,093 & 1,870 {\scriptsize ($.37$)} & 1,748 {\scriptsize ($.34$)} & 1,475 {\scriptsize ($.29$)} \\
QuaRTz & 991 & 922 {\scriptsize ($.93$)} & 69 {\scriptsize ($.07$)} & 3.43$\pm$3.07 & 3,401 & 1,964 {\scriptsize ($.58$)} & 706 {\scriptsize ($.21$)} & 731 {\scriptsize ($.21$)} \\
ReClor & 964 & 904 {\scriptsize ($.94$)} & 60 {\scriptsize ($.06$)} & 6.16$\pm$6.77 & 5,938 & 2,157 {\scriptsize ($.36$)} & 2,243 {\scriptsize ($.38$)} & 1,538 {\scriptsize ($.26$)} \\
\midrule
\textbf{TOTAL} & \textbf{9,318} & \textbf{7,746} {\scriptsize ($.83$)} & \textbf{1,572} {\scriptsize ($.17$)} & \textbf{6.59$\pm$8.03} & \textbf{61,375} & \textbf{20,407} {\scriptsize ($.33$)} & \textbf{23,442} {\scriptsize ($.38$)} & \textbf{17,526} {\scriptsize ($.29$)} \\
\addlinespace[4pt]
\midrule
\multicolumn{9}{c}{\footnotesize\textbf{\texttt{mistralai-Magistral-Small-2506}}} \\
\addlinespace[2pt]
GSM8K & 909 & 126 {\scriptsize ($.14$)} & 783 {\scriptsize ($.86$)} & 7.84$\pm$2.40 & 7,129 & 266 {\scriptsize ($.04$)} & 2,370 {\scriptsize ($.33$)} & 4,493 {\scriptsize ($.63$)} \\
TAT-QA & 921 & 204 {\scriptsize ($.22$)} & 717 {\scriptsize ($.78$)} & 6.50$\pm$2.35 & 5,983 & 409 {\scriptsize ($.07$)} & 1,336 {\scriptsize ($.22$)} & 4,238 {\scriptsize ($.71$)} \\
MedQA & 970 & 694 {\scriptsize ($.72$)} & 276 {\scriptsize ($.28$)} & 4.73$\pm$1.85 & 4,590 & 1,788 {\scriptsize ($.39$)} & 1,404 {\scriptsize ($.31$)} & 1,398 {\scriptsize ($.30$)} \\
LEXam & 928 & 212 {\scriptsize ($.23$)} & 716 {\scriptsize ($.77$)} & 5.69$\pm$2.65 & 5,282 & 450 {\scriptsize ($.09$)} & 3,374 {\scriptsize ($.64$)} & 1,458 {\scriptsize ($.28$)} \\
ARC & 986 & 855 {\scriptsize ($.87$)} & 131 {\scriptsize ($.13$)} & 3.89$\pm$1.95 & 3,839 & 1,986 {\scriptsize ($.52$)} & 1,019 {\scriptsize ($.27$)} & 834 {\scriptsize ($.22$)} \\
CommonsenseQA2 & 985 & 677 {\scriptsize ($.69$)} & 308 {\scriptsize ($.31$)} & 3.95$\pm$2.04 & 3,887 & 1,305 {\scriptsize ($.34$)} & 1,244 {\scriptsize ($.32$)} & 1,338 {\scriptsize ($.34$)} \\
LogiQA & 965 & 510 {\scriptsize ($.53$)} & 455 {\scriptsize ($.47$)} & 5.23$\pm$2.24 & 5,051 & 1,223 {\scriptsize ($.24$)} & 2,363 {\scriptsize ($.47$)} & 1,465 {\scriptsize ($.29$)} \\
OpenBookQA & 985 & 835 {\scriptsize ($.85$)} & 150 {\scriptsize ($.15$)} & 3.81$\pm$1.82 & 3,750 & 1,873 {\scriptsize ($.50$)} & 900 {\scriptsize ($.24$)} & 977 {\scriptsize ($.26$)} \\
QuaRTz & 992 & 877 {\scriptsize ($.88$)} & 115 {\scriptsize ($.12$)} & 3.69$\pm$1.94 & 3,663 & 2,096 {\scriptsize ($.57$)} & 824 {\scriptsize ($.23$)} & 743 {\scriptsize ($.20$)} \\
ReClor & 975 & 749 {\scriptsize ($.77$)} & 226 {\scriptsize ($.23$)} & 4.88$\pm$1.97 & 4,755 & 1,859 {\scriptsize ($.39$)} & 1,695 {\scriptsize ($.36$)} & 1,201 {\scriptsize ($.25$)} \\
\midrule
\textbf{TOTAL} & \textbf{9,616} & \textbf{5,739} {\scriptsize ($.60$)} & \textbf{3,877} {\scriptsize ($.40$)} & \textbf{4.98$\pm$2.48} & \textbf{47,929} & \textbf{13,255} {\scriptsize ($.28$)} & \textbf{16,529} {\scriptsize ($.34$)} & \textbf{18,145} {\scriptsize ($.38$)} \\
\addlinespace[4pt]
\midrule
\multicolumn{9}{c}{\footnotesize\textbf{\texttt{LGAI-EXAONE-EXAONE-Deep-32B}}} \\
\addlinespace[2pt]
GSM8K & 977 & 921 {\scriptsize ($.94$)} & 56 {\scriptsize ($.06$)} & 18.82$\pm$15.58 & 18,391 & 6,750 {\scriptsize ($.37$)} & 4,540 {\scriptsize ($.25$)} & 7,101 {\scriptsize ($.39$)} \\
TAT-QA & 973 & 558 {\scriptsize ($.57$)} & 415 {\scriptsize ($.43$)} & 15.35$\pm$13.68 & 14,931 & 2,559 {\scriptsize ($.17$)} & 6,182 {\scriptsize ($.41$)} & 6,190 {\scriptsize ($.41$)} \\
MedQA & 972 & 652 {\scriptsize ($.67$)} & 320 {\scriptsize ($.33$)} & 26.09$\pm$20.96 & 25,362 & 3,639 {\scriptsize ($.14$)} & 10,765 {\scriptsize ($.42$)} & 10,958 {\scriptsize ($.43$)} \\
LEXam & 946 & 246 {\scriptsize ($.26$)} & 700 {\scriptsize ($.74$)} & 28.41$\pm$19.03 & 26,874 & 1,559 {\scriptsize ($.06$)} & 17,653 {\scriptsize ($.66$)} & 7,662 {\scriptsize ($.29$)} \\
ARC & 992 & 933 {\scriptsize ($.94$)} & 59 {\scriptsize ($.06$)} & 9.22$\pm$11.80 & 9,142 & 3,233 {\scriptsize ($.35$)} & 2,811 {\scriptsize ($.31$)} & 3,098 {\scriptsize ($.34$)} \\
CommonsenseQA2 & 984 & 768 {\scriptsize ($.78$)} & 216 {\scriptsize ($.22$)} & 13.04$\pm$15.57 & 12,832 & 2,715 {\scriptsize ($.21$)} & 3,755 {\scriptsize ($.29$)} & 6,362 {\scriptsize ($.50$)} \\
LogiQA & 970 & 513 {\scriptsize ($.53$)} & 457 {\scriptsize ($.47$)} & 21.45$\pm$17.01 & 20,811 & 2,246 {\scriptsize ($.11$)} & 10,050 {\scriptsize ($.48$)} & 8,515 {\scriptsize ($.41$)} \\
OpenBookQA & 979 & 886 {\scriptsize ($.91$)} & 93 {\scriptsize ($.10$)} & 14.74$\pm$18.58 & 14,427 & 3,572 {\scriptsize ($.25$)} & 4,589 {\scriptsize ($.32$)} & 6,266 {\scriptsize ($.43$)} \\
QuaRTz & 990 & 933 {\scriptsize ($.94$)} & 57 {\scriptsize ($.06$)} & 6.93$\pm$7.70 & 6,861 & 3,170 {\scriptsize ($.46$)} & 1,179 {\scriptsize ($.17$)} & 2,512 {\scriptsize ($.37$)} \\
ReClor & 975 & 823 {\scriptsize ($.84$)} & 152 {\scriptsize ($.16$)} & 13.56$\pm$12.15 & 13,217 & 3,113 {\scriptsize ($.24$)} & 5,583 {\scriptsize ($.42$)} & 4,521 {\scriptsize ($.34$)} \\
\midrule
\textbf{TOTAL} & \textbf{9,758} & \textbf{7,233} {\scriptsize ($.74$)} & \textbf{2,525} {\scriptsize ($.26$)} & \textbf{16.69$\pm$16.95} & \textbf{162,848} & \textbf{32,556} {\scriptsize ($.20$)} & \textbf{67,107} {\scriptsize ($.41$)} & \textbf{63,185} {\scriptsize ($.39$)} \\
\addlinespace[4pt]
\midrule
\multicolumn{9}{c}{\footnotesize\textbf{\texttt{Qwen-QwQ-32B}}} \\
\addlinespace[2pt]
GSM8K & 984 & 946 {\scriptsize ($.96$)} & 38 {\scriptsize ($.04$)} & 10.52$\pm$12.47 & 10,348 & 4,458 {\scriptsize ($.43$)} & 2,492 {\scriptsize ($.24$)} & 3,398 {\scriptsize ($.33$)} \\
TAT-QA & 983 & 627 {\scriptsize ($.64$)} & 356 {\scriptsize ($.36$)} & 7.65$\pm$9.09 & 7,518 & 1,974 {\scriptsize ($.26$)} & 3,098 {\scriptsize ($.41$)} & 2,446 {\scriptsize ($.33$)} \\
MedQA & 975 & 790 {\scriptsize ($.81$)} & 185 {\scriptsize ($.19$)} & 15.27$\pm$19.66 & 14,886 & 3,292 {\scriptsize ($.22$)} & 6,350 {\scriptsize ($.43$)} & 5,244 {\scriptsize ($.35$)} \\
LEXam & 947 & 304 {\scriptsize ($.32$)} & 643 {\scriptsize ($.68$)} & 28.89$\pm$27.30 & 27,359 & 2,239 {\scriptsize ($.08$)} & 19,250 {\scriptsize ($.70$)} & 5,870 {\scriptsize ($.21$)} \\
ARC & 987 & 945 {\scriptsize ($.96$)} & 42 {\scriptsize ($.04$)} & 6.11$\pm$12.05 & 6,027 & 2,449 {\scriptsize ($.41$)} & 2,016 {\scriptsize ($.33$)} & 1,562 {\scriptsize ($.26$)} \\
CommonsenseQA2 & 990 & 809 {\scriptsize ($.82$)} & 181 {\scriptsize ($.18$)} & 6.42$\pm$7.58 & 6,351 & 1,906 {\scriptsize ($.30$)} & 1,762 {\scriptsize ($.28$)} & 2,683 {\scriptsize ($.42$)} \\
LogiQA & 988 & 687 {\scriptsize ($.70$)} & 301 {\scriptsize ($.30$)} & 15.95$\pm$18.96 & 15,758 & 2,484 {\scriptsize ($.16$)} & 7,959 {\scriptsize ($.51$)} & 5,315 {\scriptsize ($.34$)} \\
OpenBookQA & 985 & 907 {\scriptsize ($.92$)} & 78 {\scriptsize ($.08$)} & 8.22$\pm$15.27 & 8,095 & 2,543 {\scriptsize ($.31$)} & 2,899 {\scriptsize ($.36$)} & 2,653 {\scriptsize ($.33$)} \\
QuaRTz & 996 & 936 {\scriptsize ($.94$)} & 60 {\scriptsize ($.06$)} & 4.79$\pm$6.38 & 4,766 & 2,367 {\scriptsize ($.50$)} & 1,058 {\scriptsize ($.22$)} & 1,341 {\scriptsize ($.28$)} \\
ReClor & 979 & 914 {\scriptsize ($.93$)} & 65 {\scriptsize ($.07$)} & 8.10$\pm$11.77 & 7,934 & 2,492 {\scriptsize ($.31$)} & 3,145 {\scriptsize ($.40$)} & 2,297 {\scriptsize ($.29$)} \\
\midrule
\textbf{TOTAL} & \textbf{9,814} & \textbf{7,865} {\scriptsize ($.80$)} & \textbf{1,949} {\scriptsize ($.20$)} & \textbf{11.11$\pm$16.68} & \textbf{109,042} & \textbf{26,204} {\scriptsize ($.24$)} & \textbf{50,029} {\scriptsize ($.46$)} & \textbf{32,809} {\scriptsize ($.30$)} \\
\end{longtable}
\endgroup
\twocolumn

\onecolumn
\begingroup
\scriptsize
\begin{longtable}{l r r r r r r r r}
\caption{A breakdown of sample- and chunk-level test statistics across various datasets, aggregated by RLMs. \textbf{Notation:} $S$ = number of samples (problems); $C$ = chunks per sample; $p$ = proportion; labels $L \in \{\checkmark,\text{\ding{55}},\varnothing\}$ denote correct, incorrect, and no-result. \textbf{Columns:} $S$; $\checkmark$ = correct samples (\#S, $p$); \ding{55} = incorrect samples (\#S, $p$); $C(\mu\pm\sigma)$ = mean$\pm$sd chunks per sample; $\sum C$ = total chunks; $\checkmark$ = correct chunks (\#C, $p$); \ding{55} = incorrect chunks (\#C, $p$); $\varnothing$ = no-result chunks (\#C, $p$). Proportions $p$ are typeset tiny as ($.xx$).}\label{tab:test_merged_llm}\\
\toprule
\footnotesize\textbf{Dataset} & \footnotesize\textbf{$S$} & \footnotesize\textbf{$\checkmark$ (\#\!$S$, $p$)} & \footnotesize\textbf{\ding{55} (\#\!$S$, $p$)} & \footnotesize\textbf{$C$ ($\mu\!\pm\!\sigma$)} & \footnotesize\textbf{$\sum C$} & \footnotesize\textbf{$\checkmark$ (\#\!$C$, $p$)} & \footnotesize\textbf{\ding{55} (\#\!$C$, $p$)} & \footnotesize\textbf{$\varnothing$ (\#\!$C$, $p$)}\\
\midrule

\endfirsthead
\multicolumn{9}{l}{\scriptsize\emph{(Continued from previous page)}}\\
\toprule
\footnotesize\textbf{Dataset} & \footnotesize\textbf{$S$} & \footnotesize\textbf{$\checkmark$ (\#\!$S$, $p$)} & \footnotesize\textbf{\ding{55} (\#\!$S$, $p$)} & \footnotesize\textbf{$C$ ($\mu\!\pm\!\sigma$)} & \footnotesize\textbf{$\sum C$} & \footnotesize\textbf{$\checkmark$ (\#\!$C$, $p$)} & \footnotesize\textbf{\ding{55} (\#\!$C$, $p$)} & \footnotesize\textbf{$\varnothing$ (\#\!$C$, $p$)}\\
\midrule

\endhead
\midrule
\multicolumn{9}{r}{\scriptsize\emph{(Continued on next page)}}\\

\endfoot
\bottomrule
\endlastfoot
\multicolumn{9}{c}{\footnotesize\textbf{\texttt{microsoft-Phi-4-mini-flash-reasoning}}} \\
\addlinespace[2pt]
MATH & 4,560 & 4,179 {\scriptsize ($.92$)} & 381 {\scriptsize ($.08$)} & 14.95$\pm$15.16 & 68,152 & 22,273 {\scriptsize ($.33$)} & 21,197 {\scriptsize ($.31$)} & 24,682 {\scriptsize ($.36$)} \\
FinQA & 979 & 256 {\scriptsize ($.26$)} & 723 {\scriptsize ($.74$)} & 10.31$\pm$13.41 & 10,097 & 719 {\scriptsize ($.07$)} & 5,834 {\scriptsize ($.58$)} & 3,544 {\scriptsize ($.35$)} \\
MedMCQA & 5,505 & 1,285 {\scriptsize ($.23$)} & 4,220 {\scriptsize ($.77$)} & 12.37$\pm$18.04 & 68,076 & 5,200 {\scriptsize ($.08$)} & 37,607 {\scriptsize ($.55$)} & 25,269 {\scriptsize ($.37$)} \\
LegalBench & 18,321 & 8,649 {\scriptsize ($.47$)} & 9,672 {\scriptsize ($.53$)} & 5.35$\pm$7.38 & 97,965 & 19,841 {\scriptsize ($.20$)} & 47,817 {\scriptsize ($.49$)} & 30,307 {\scriptsize ($.31$)} \\
MMLU-Pro & 9,404 & 5,063 {\scriptsize ($.54$)} & 4,341 {\scriptsize ($.46$)} & 15.64$\pm$19.81 & 147,104 & 22,850 {\scriptsize ($.16$)} & 73,686 {\scriptsize ($.50$)} & 50,568 {\scriptsize ($.34$)} \\
BBH & 5,640 & 4,161 {\scriptsize ($.74$)} & 1,479 {\scriptsize ($.26$)} & 14.93$\pm$17.22 & 84,177 & 21,582 {\scriptsize ($.26$)} & 31,061 {\scriptsize ($.37$)} & 31,534 {\scriptsize ($.37$)} \\
\midrule
\textbf{TOTAL} & \textbf{44,409} & \textbf{23,593} {\scriptsize ($.53$)} & \textbf{20,816} {\scriptsize ($.47$)} & \textbf{10.71$\pm$15.25} & \textbf{475,571} & \textbf{92,465} {\scriptsize ($.19$)} & \textbf{217,202} {\scriptsize ($.46$)} & \textbf{165,904} {\scriptsize ($.35$)} \\
\addlinespace[4pt]
\midrule
\multicolumn{9}{c}{\footnotesize\textbf{\texttt{Qwen-Qwen3-8B}}} \\
\addlinespace[2pt]
MATH & 4,384 & 4,122 {\scriptsize ($.94$)} & 262 {\scriptsize ($.06$)} & 27.35$\pm$16.27 & 119,893 & 33,878 {\scriptsize ($.28$)} & 36,069 {\scriptsize ($.30$)} & 49,946 {\scriptsize ($.42$)} \\
FinQA & 1,038 & 382 {\scriptsize ($.37$)} & 656 {\scriptsize ($.63$)} & 10.46$\pm$8.83 & 10,862 & 1,339 {\scriptsize ($.12$)} & 5,927 {\scriptsize ($.55$)} & 3,596 {\scriptsize ($.33$)} \\
MedMCQA & 5,893 & 1,634 {\scriptsize ($.28$)} & 4,259 {\scriptsize ($.72$)} & 18.72$\pm$16.48 & 110,315 & 10,511 {\scriptsize ($.10$)} & 60,755 {\scriptsize ($.55$)} & 39,049 {\scriptsize ($.35$)} \\
LegalBench & 20,720 & 13,997 {\scriptsize ($.68$)} & 6,723 {\scriptsize ($.32$)} & 5.39$\pm$5.97 & 111,704 & 26,217 {\scriptsize ($.23$)} & 43,053 {\scriptsize ($.39$)} & 42,434 {\scriptsize ($.38$)} \\
MMLU-Pro & 9,962 & 7,624 {\scriptsize ($.77$)} & 2,338 {\scriptsize ($.23$)} & 24.49$\pm$18.81 & 243,933 & 48,678 {\scriptsize ($.20$)} & 111,128 {\scriptsize ($.46$)} & 84,127 {\scriptsize ($.34$)} \\
BBH & 6,068 & 4,967 {\scriptsize ($.82$)} & 1,101 {\scriptsize ($.18$)} & 16.71$\pm$15.87 & 101,377 & 31,156 {\scriptsize ($.31$)} & 31,110 {\scriptsize ($.31$)} & 39,111 {\scriptsize ($.39$)} \\
\midrule
\textbf{TOTAL} & \textbf{48,065} & \textbf{32,726} {\scriptsize ($.68$)} & \textbf{15,339} {\scriptsize ($.32$)} & \textbf{14.52$\pm$15.95} & \textbf{698,084} & \textbf{151,779} {\scriptsize ($.22$)} & \textbf{288,042} {\scriptsize ($.41$)} & \textbf{258,263} {\scriptsize ($.37$)} \\
\addlinespace[4pt]
\midrule
\multicolumn{9}{c}{\footnotesize\textbf{\texttt{Qwen-Qwen3-14B}}} \\
\addlinespace[2pt]
MATH & 4,489 & 4,241 {\scriptsize ($.94$)} & 248 {\scriptsize ($.06$)} & 22.40$\pm$14.04 & 100,551 & 30,626 {\scriptsize ($.30$)} & 29,685 {\scriptsize ($.30$)} & 40,240 {\scriptsize ($.40$)} \\
FinQA & 1,075 & 382 {\scriptsize ($.36$)} & 693 {\scriptsize ($.64$)} & 5.91$\pm$5.28 & 6,349 & 847 {\scriptsize ($.13$)} & 3,450 {\scriptsize ($.54$)} & 2,052 {\scriptsize ($.32$)} \\
MedMCQA & 5,895 & 1,726 {\scriptsize ($.29$)} & 4,169 {\scriptsize ($.71$)} & 9.78$\pm$9.45 & 57,654 & 6,120 {\scriptsize ($.11$)} & 31,640 {\scriptsize ($.55$)} & 19,894 {\scriptsize ($.35$)} \\
LegalBench & 20,463 & 14,433 {\scriptsize ($.71$)} & 6,030 {\scriptsize ($.29$)} & 3.77$\pm$3.90 & 77,123 & 20,958 {\scriptsize ($.27$)} & 28,372 {\scriptsize ($.37$)} & 27,793 {\scriptsize ($.36$)} \\
MMLU-Pro & 10,312 & 8,061 {\scriptsize ($.78$)} & 2,251 {\scriptsize ($.22$)} & 15.46$\pm$13.78 & 159,452 & 35,322 {\scriptsize ($.22$)} & 69,048 {\scriptsize ($.43$)} & 55,082 {\scriptsize ($.35$)} \\
BBH & 6,281 & 5,182 {\scriptsize ($.82$)} & 1,099 {\scriptsize ($.17$)} & 9.31$\pm$9.71 & 58,501 & 20,293 {\scriptsize ($.35$)} & 16,519 {\scriptsize ($.28$)} & 21,689 {\scriptsize ($.37$)} \\
\midrule
\textbf{TOTAL} & \textbf{48,515} & \textbf{34,025} {\scriptsize ($.70$)} & \textbf{14,490} {\scriptsize ($.30$)} & \textbf{9.47$\pm$11.22} & \textbf{459,630} & \textbf{114,166} {\scriptsize ($.25$)} & \textbf{178,714} {\scriptsize ($.39$)} & \textbf{166,750} {\scriptsize ($.36$)} \\
\addlinespace[4pt]
\midrule
\multicolumn{9}{c}{\footnotesize\textbf{\texttt{mistralai-Magistral-Small-2506}}} \\
\addlinespace[2pt]
MATH & 4,589 & 193 {\scriptsize ($.04$)} & 4,396 {\scriptsize ($.96$)} & 7.54$\pm$2.33 & 34,622 & 436 {\scriptsize ($.01$)} & 8,874 {\scriptsize ($.26$)} & 25,312 {\scriptsize ($.73$)} \\
FinQA & 1,022 & 32 {\scriptsize ($.03$)} & 990 {\scriptsize ($.97$)} & 7.46$\pm$1.84 & 7,627 & 43 {\scriptsize ($.01$)} & 1,896 {\scriptsize ($.25$)} & 5,688 {\scriptsize ($.75$)} \\
MedMCQA & 5,968 & 1,386 {\scriptsize ($.23$)} & 4,582 {\scriptsize ($.77$)} & 4.58$\pm$1.94 & 27,344 & 3,481 {\scriptsize ($.13$)} & 16,642 {\scriptsize ($.61$)} & 7,221 {\scriptsize ($.26$)} \\
LegalBench & 20,810 & 10,845 {\scriptsize ($.52$)} & 9,965 {\scriptsize ($.48$)} & 3.56$\pm$1.54 & 74,066 & 19,700 {\scriptsize ($.27$)} & 33,515 {\scriptsize ($.45$)} & 20,851 {\scriptsize ($.28$)} \\
MMLU-Pro & 11,347 & 3,783 {\scriptsize ($.33$)} & 7,564 {\scriptsize ($.67$)} & 6.39$\pm$2.82 & 72,552 & 9,290 {\scriptsize ($.13$)} & 26,815 {\scriptsize ($.37$)} & 36,447 {\scriptsize ($.50$)} \\
BBH & 6,057 & 2,064 {\scriptsize ($.34$)} & 3,993 {\scriptsize ($.66$)} & 6.93$\pm$2.58 & 41,966 & 5,089 {\scriptsize ($.12$)} & 15,461 {\scriptsize ($.37$)} & 21,416 {\scriptsize ($.51$)} \\
\midrule
\textbf{TOTAL} & \textbf{49,793} & \textbf{18,303} {\scriptsize ($.37$)} & \textbf{31,490} {\scriptsize ($.63$)} & \textbf{5.18$\pm$2.66} & \textbf{258,177} & \textbf{38,039} {\scriptsize ($.15$)} & \textbf{103,203} {\scriptsize ($.40$)} & \textbf{116,935} {\scriptsize ($.45$)} \\
\addlinespace[4pt]
\midrule
\multicolumn{9}{c}{\footnotesize\textbf{\texttt{LGAI-EXAONE-EXAONE-Deep-32B}}} \\
\addlinespace[2pt]
MATH & 4,784 & 4,040 {\scriptsize ($.84$)} & 744 {\scriptsize ($.16$)} & 27.90$\pm$14.58 & 133,466 & 30,697 {\scriptsize ($.23$)} & 48,691 {\scriptsize ($.36$)} & 54,078 {\scriptsize ($.41$)} \\
FinQA & 1,107 & 380 {\scriptsize ($.34$)} & 727 {\scriptsize ($.66$)} & 24.62$\pm$15.38 & 27,249 & 2,027 {\scriptsize ($.07$)} & 13,416 {\scriptsize ($.49$)} & 11,806 {\scriptsize ($.43$)} \\
MedMCQA & 5,999 & 1,581 {\scriptsize ($.26$)} & 4,418 {\scriptsize ($.74$)} & 32.53$\pm$27.43 & 195,162 & 13,983 {\scriptsize ($.07$)} & 96,255 {\scriptsize ($.49$)} & 84,924 {\scriptsize ($.44$)} \\
LegalBench & 20,879 & 14,648 {\scriptsize ($.70$)} & 6,231 {\scriptsize ($.30$)} & 12.56$\pm$15.41 & 262,286 & 42,549 {\scriptsize ($.16$)} & 92,368 {\scriptsize ($.35$)} & 127,369 {\scriptsize ($.49$)} \\
MMLU-Pro & 11,473 & 7,119 {\scriptsize ($.62$)} & 4,354 {\scriptsize ($.38$)} & 32.72$\pm$20.99 & 375,342 & 47,973 {\scriptsize ($.13$)} & 185,339 {\scriptsize ($.49$)} & 142,030 {\scriptsize ($.38$)} \\
BBH & 6,208 & 4,902 {\scriptsize ($.79$)} & 1,306 {\scriptsize ($.21$)} & 24.95$\pm$17.54 & 154,890 & 35,943 {\scriptsize ($.23$)} & 52,970 {\scriptsize ($.34$)} & 65,977 {\scriptsize ($.43$)} \\
\midrule
\textbf{TOTAL} & \textbf{50,450} & \textbf{32,670} {\scriptsize ($.65$)} & \textbf{17,780} {\scriptsize ($.35$)} & \textbf{22.76$\pm$20.77} & \textbf{1,148,395} & \textbf{173,172} {\scriptsize ($.15$)} & \textbf{489,039} {\scriptsize ($.43$)} & \textbf{486,184} {\scriptsize ($.42$)} \\
\addlinespace[4pt]
\midrule
\multicolumn{9}{c}{\footnotesize\textbf{\texttt{Qwen-QwQ-32B}}} \\
\addlinespace[2pt]
MATH & 4,813 & 4,177 {\scriptsize ($.87$)} & 636 {\scriptsize ($.13$)} & 24.56$\pm$16.65 & 118,207 & 30,404 {\scriptsize ($.26$)} & 41,809 {\scriptsize ($.35$)} & 45,994 {\scriptsize ($.39$)} \\
FinQA & 1,112 & 406 {\scriptsize ($.37$)} & 706 {\scriptsize ($.63$)} & 13.16$\pm$15.00 & 14,630 & 1,411 {\scriptsize ($.10$)} & 7,701 {\scriptsize ($.53$)} & 5,518 {\scriptsize ($.38$)} \\
MedMCQA & 6,040 & 1,695 {\scriptsize ($.28$)} & 4,345 {\scriptsize ($.72$)} & 21.06$\pm$27.68 & 127,195 & 10,465 {\scriptsize ($.08$)} & 69,914 {\scriptsize ($.55$)} & 46,816 {\scriptsize ($.37$)} \\
LegalBench & 20,908 & 13,353 {\scriptsize ($.64$)} & 7,555 {\scriptsize ($.36$)} & 7.63$\pm$12.19 & 159,603 & 26,789 {\scriptsize ($.17$)} & 60,468 {\scriptsize ($.38$)} & 72,346 {\scriptsize ($.45$)} \\
MMLU-Pro & 11,597 & 8,165 {\scriptsize ($.70$)} & 3,432 {\scriptsize ($.30$)} & 26.58$\pm$22.85 & 308,197 & 47,887 {\scriptsize ($.16$)} & 153,966 {\scriptsize ($.50$)} & 106,344 {\scriptsize ($.35$)} \\
BBH & 6,322 & 5,266 {\scriptsize ($.83$)} & 1,056 {\scriptsize ($.17$)} & 14.24$\pm$13.61 & 90,051 & 26,997 {\scriptsize ($.30$)} & 29,290 {\scriptsize ($.33$)} & 33,764 {\scriptsize ($.37$)} \\
\midrule
\textbf{TOTAL} & \textbf{50,792} & \textbf{33,062} {\scriptsize ($.65$)} & \textbf{17,730} {\scriptsize ($.35$)} & \textbf{16.10$\pm$19.76} & \textbf{817,883} & \textbf{143,953} {\scriptsize ($.18$)} & \textbf{363,148} {\scriptsize ($.44$)} & \textbf{310,782} {\scriptsize ($.38$)} \\
\end{longtable}
\endgroup
\twocolumn

\subsection{Example of a Fully Annotated and Segmented Reasoning Trace}
\label{app:example_trace}
To make the structure of the RMCB dataset concrete, we provide a fully worked example of a single reasoning trace. This example illustrates the raw input, the model-generated reasoning segmented into coherent regions, and the final answer produced after the internal reasoning phase. The example is shown exactly as stored in the dataset, with visual grouping used to indicate segmentation.

\begin{table}[h]
\centering
\small
\renewcommand{\arraystretch}{1.3}
\caption{Illustrative example of a single RMCB reasoning trace. Gray denotes the input prompt. Colored regions indicate distinct segments of the model’s reasoning.}
\label{tab:reasoning_example}
\begin{tabular}{|p{0.95\linewidth}|}
\hline
\textbf{Input} \\
\begin{inputbox}
The pH scale measures acidity. A higher concentration of hydrogen ions (H\textsuperscript{+}) corresponds to a lower pH value. What would be the pH value of a solution with fewer hydrogen ions?
\end{inputbox}
\\
\textbf{Model Reasoning} \\
\begin{bluebox}
Okay, so the question is asking about the pH value of something with fewer hydrogen ions. Let me think. The context says that a higher concentration of hydrogen ions makes a solution more acidic and has a lower pH. So if there are more H\textsuperscript{+} ions, pH is lower. Then, the opposite should be true, right? If there are fewer hydrogen ions, the pH should be higher. Because pH is a scale where lower numbers are more acidic and higher numbers are more basic or alkaline.
\end{bluebox}
\vspace{0.5pt}
\begin{greenbox}
Wait, let me make sure I'm not mixing this up. The pH scale goes from 0 to 14. 7 is neutral. Below 7 is acidic, above is basic. Each number represents a tenfold difference in hydrogen ion concentration. So, for example, pH 3 has ten times more H\textsuperscript{+} than pH 4. So if something has less H\textsuperscript{+} ions, moving up the pH scale, like from 3 to 4, the pH increases. Therefore, less H\textsuperscript{+} means higher pH. So the answer should be (A) high.
\end{greenbox}
\vspace{0.5pt}
\begin{yellowbox}
But let me check again. The question says ``something with less hydrogen ions would have what pH value?'' Since more H\textsuperscript{+} equals lower pH, then less H\textsuperscript{+} would mean the pH is higher. So yes, the answer is A. The options are high (A) or low (B). The context directly states that higher H\textsuperscript{+} leads to lower pH, so inversely, lower H\textsuperscript{+} leads to higher pH. Therefore, the correct answer is A.
\end{yellowbox}
\\ \hline
\end{tabular}
\end{table}

\subsection{Benchmark Availability and Licensing}
\label{app:benchmark_availability}

The \textbf{RMCB} benchmark is a composite resource built upon 16 distinct public datasets, each governed by its own license. These licenses range from highly permissive (e.g., MIT, Apache 2.0) to more restrictive terms that include NonCommercial (NC), ShareAlike (SA), or research-use-only clauses. To ensure full compliance with the terms of every source dataset, we cannot distribute the complete, aggregated benchmark as a single downloadable package. Instead, we adopt a standard, script-based reconstruction approach that separates our original contributions from the source data. We make the following components publicly available:
\begin{itemize}
    \item \textbf{The Derived Data File:} A comprehensive JSONL file containing our novel contributions is released under the \textbf{CC BY 4.0 license}. Each line includes: (1) a unique, deterministic \texttt{record\_id} created by hashing the source content; (2) the full, model-generated reasoning trace; and (3) the associated correctness annotations.
    \item \textbf{The Reconstruction Script:} A Python script that automates the process of building the full RMCB benchmark on a user's local machine. All of our code, including this script, is released under the \textbf{MIT license}.
\end{itemize}

To construct the full benchmark, users must first download the original source datasets, thereby agreeing to their respective licenses. Our provided script then processes these local files, generates the corresponding \texttt{record\_id} hashes, and merges the source data with our derived data file to create the complete benchmark. This method ensures that the original data is never redistributed by us.

\subsubsection*{Important Disclaimer}
Users are solely responsible for acquiring the source datasets from their official distributors and for adhering to their original license terms. The final, reconstructed RMCB dataset is a derivative work. As such, it is governed by the most restrictive terms of its constituent components. This means the complete benchmark is intended for \textbf{non-commercial, research-use only} and is subject to all applicable \textit{ShareAlike (SA)} provisions inherited from its sources. The information provided here does not constitute legal advice.

%% file: sections/appendix_baseline_methods.tex
\section{Confidence Estimation Methods}
\label{app:confidence_estimation_methods}

This section details the confidence estimation methods evaluated in our benchmark. We categorize them into two groups: established baseline methods adapted from the literature (\ref{app:baseline_ones}), and novel methods we developed to specifically address the challenges of multi-step reasoning (\ref{app:new_ones}). For all trainable models, hyperparameters were systematically optimized using Optuna with a consistent trial budget to ensure a fair and rigorous comparison.

\subsection{Baseline Methods}
\label{app:baseline_ones}

We selected a representative set of baseline methods from recent literature, covering both simple probing techniques and more complex, reasoning-focused approaches.

\subsubsection{YVCE (Verbalized Confidence Estimation)}
\citet{YVCE} introduced a verbalized confidence estimation approach, where a reasoning model is prompted to assess its own solution after completing the reasoning process. For clarity, we refer to this method as YVCE. The procedure operates in two stages: first, the model generates its full reasoning trace and final answer; second, this output is re-used to construct a new conversational turn. A detailed system prompt, defining a 10-point confidence scale, is combined with a “nudge” phrase that encourages the model to select one of these labels as a continuation of its thought process. The final confidence score is then parsed from this verbalized self-assessment.

\begin{tcolorbox}[title=YVCE System Prompt,fonttitle=\bfseries,fontupper=\small]
First, reason through the question step by step to arrive at an answer. \\
Then, thoroughly assess your confidence in that answer by evaluating your thinking process so far. \\
Finally, classify your confidence into one of the following classes based on how likely your answer is to be correct: \\
- "Almost no chance" (0.0–0.1) \\
- "Highly unlikely" (0.1–0.2) \\
- "Chances are slight" (0.2–0.3) \\
- "Unlikely" (0.3–0.4) \\
- "Less than even" (0.4–0.5) \\
- "Better than even" (0.5–0.6) \\
- "Likely" (0.6–0.7) \\
- "Very good chance" (0.7–0.8) \\
- "Highly likely" (0.8–0.9) \\
- "Almost certain" (0.9–1.0) \\
Each category reflects the probability that your answer is correct.\\
At the very end of your output, format your answer and confidence as \\
**Answer**: \$ANSWER \\
**Confidence**: \$CLASS \\
where CLASS is one of the names (only the names without the probability ranges) of the classes above.
\end{tcolorbox}

\begin{tcolorbox}[title=YVCE Nudge Prompt,fonttitle=\bfseries,fontupper=\small]
Now, finally, if I were to briefly mention my confidence among the given classes in the system prompt, I would choose 
\end{tcolorbox}

\subsubsection{TLCC (Token-Level Chunk Classification)}
\label{app:TLCC_method}

TLCC investigates whether low-dimensional statistics derived purely from output logits can serve as a computationally efficient proxy for deep hidden-state representations. While methods like SFHS require storing and processing high-dimensional embedding vectors (often $d=4096+$), TLCC operates on a compact set of uncertainty metrics calculated during the generation pass.

\paragraph{Token-Level Feature Extraction}
For every token $x_t$ generated in the reasoning trace, we extract a vector of ten statistical features designed to capture various aspects of the model's local predictive uncertainty—ranging from confidence in the top choice to the dispersion of the probability mass. These features are detailed in Table~\ref{tab:tlcc_features}.

\begin{table*}[t]
\centering
\caption{Token-level confidence features extracted for the TLCC method. For a vocabulary size $V$, let $\mathbf{z} \in \mathbb{R}^V$ be the logit vector and $\mathbf{p} = \text{softmax}(\mathbf{z})$ be the probability distribution. Indices $(1), (2), \dots$ denote rank-ordered elements such that $p_{(1)} \ge p_{(2)} \ge \dots \ge p_{(V)}$.}
\label{tab:tlcc_features}
\resizebox{\textwidth}{!}{
\begin{tabular}{l l l}
\toprule
\textbf{Feature} & \textbf{Definition / Formulation} & \textbf{Intuition} \\
\midrule
Top-1 Probability & $p_{(1)}$ & The model's raw confidence in its selected token. \\
Log Top-1 Prob & $\log(p_{(1)})$ & Equivalent to the negative log-likelihood (NLL); sensitive to extremely low probabilities. \\
Logit Margin & $z_{(1)} - z_{(2)}$ & The distance between the best and runner-up logits (unnormalized confidence). \\
Probability Gap & $p_{(1)} - p_{(2)}$ & The gap between the best and runner-up probabilities; indicates competition between top candidates. \\
Entropy & $H(\mathbf{p}) = -\sum p_i \log p_i$ & Measures total uncertainty; high entropy implies a "flat" distribution. \\
Normalized Entropy & $H(\mathbf{p}) / \log(V)$ & Entropy scaled by vocabulary size, making it comparable across different models. \\
Top-$k$ Mass & $\sum_{i=1}^{k=5} p_{(i)}$ & Cumulative probability of the top 5 tokens; measures concentration in the head of the distribution. \\
Tail Mass & $1 - \sum_{i=1}^{k=5} p_{(i)}$ & Probability mass assigned to unlikely tokens; captures the "long tail" risk. \\
L2 Concentration & $\sum p_i^2$ & The Herfindahl index; approaches 1.0 for certainty and $1/V$ for uniform uncertainty. \\
Logit Std Dev & $\sigma(\mathbf{z})$ & Standard deviation of raw logits; indicates the sharpness of the pre-softmax distribution. \\
\bottomrule
\end{tabular}
}
\end{table*}

\paragraph{Chunk Aggregation and Classification}
To map these variable-length sequences of token features to fixed-size chunk representations, we apply four aggregation statistics—mean, standard deviation, minimum, and maximum—across the tokens within each chunk. Additionally, we include the normalized token count as a feature to capture the length of the reasoning step. This process yields a fixed-dimensional vector $v_c \in \mathbb{R}^{41}$ for each chunk (10 features $\times$ 4 statistics + 1 length feature).

Similar to the SFHS framework, the sequence of chunk vectors $[v_1, v_2, \dots, v_n]$ is treated as a time-series input. We evaluate three architectural variants to model the dependencies between these statistical summaries:
\begin{itemize}
    \item \textbf{TLCC-MLP}: A simple baseline that mean-pools the chunk vectors and classifies them using a multi-layer perceptron.
    \item \textbf{TLCC-CONV}: Applies 1D convolutions over the chunk sequence to capture local temporal patterns in uncertainty dynamics.
    \item \textbf{TLCC-LSTM}: Uses a bidirectional LSTM to model the global evolution of confidence statistics across the reasoning trace.
\end{itemize}
All variants are trained via binary cross-entropy to predict the final correctness of the answer.

\subsubsection{P(IK) (Probability of "I Know")}
This method tests the hypothesis that an LRM’s initial comprehension of a question contains a signal about its likelihood of answering correctly. It trains a lightweight classifier, an MLP, on the hidden state corresponding to the final token of the \textit{input prompt}. The goal is to predict the correctness of the eventual answer based solely on the model's state \emph{before} it begins to generate a solution.

\subsubsection{PHSV (Probing Hidden States for Self-Verification)}
\label{app:PHSV_apprndix_explanation}
\citet{PHSV} introduced Probing Hidden States for Self-Verification (PHSV), which monitors a reasoning model’s internal states throughout the reasoning process by training a lightweight classifier on the hidden representation of each intermediate reasoning chunk. The classifier is implemented as an MLP whose architecture (depth, width, and dropout) is tuned to predict the correctness of each step. To address class imbalance, the model is trained with a weighted binary cross-entropy loss. While originally proposed for early-exit classification, we adapt PHSV to our benchmark by using the confidence prediction from the \textbf{final chunk} as the overall score. In addition, we define \breakabletexttt{PHSV-half}, a variant trained on only 50\% of the data, which we employ as a crucial feature extractor for our two-stage models.

\subsection{Novel Benchmarked Methods}
\label{app:new_ones}
Building on these baselines, we developed and benchmarked a suite of novel architectures designed to better capture the sequential, relational, and dynamic properties of a reasoning trace.

\subsubsection{SFHS (Stacked Final Hidden States)} 
A natural extension of PHSV, this family of models treats the entire reasoning trace as a single sequence. Instead of classifying each chunk independently, it feeds the full stack of chunk hidden states $[h_1, h_2, \dots, h_n]$ into a more powerful sequential model to make a single, globally informed prediction. We evaluate three architectural variants: \breakabletexttt{SFHS-MLP}, \breakabletexttt{SFHS-CONV}, and \breakabletexttt{SFHS-LSTM}.

\subsubsection{GNNs (Graph Neural Networks)}
We reframe the reasoning trace as a graph where each chunk is a node, allowing us to test whether the structural properties of an argument can predict its correctness. We introduce three variants, each designed to isolate a different potential signal: the chronological flow, the logical coherence between all steps, and the evolution of the model's confidence.

\noindent\textbf{GNN-SB (Sequential Binary)}\quad
This method establishes a simple baseline by representing the reasoning trace as a sequential chain and testing whether the chronological flow of reasoning alone encodes enough signal to distinguish correct from incorrect answers. Each reasoning chunk is modeled as a node connected to its immediate successor, capturing local, step-to-step dependencies without incorporating richer relational structure. Three architectural variants are implemented, corresponding to different backbone operators from the PyTorch Geometric library: \breakabletexttt{GNN-SB-GAT}, \breakabletexttt{GNN-SB-GCN}, and \breakabletexttt{GNN-SB-GraphSAGE}. These variants differ only in how they aggregate neighborhood information.

\begin{itemize}
    \item \textbf{Graph Construction}\,: Temporal chain with directed edges $i \to i+1$.
    \item \textbf{Nodes}\,: Raw chunk hidden state $h_t$ from the reasoning model.
    \item \textbf{Edges}\,: Chronological and unweighted (no additional attributes).
    \item \textbf{Backbones}\,: \breakabletexttt{GAT} \cite{gat} (attention-based), \breakabletexttt{GCN} \cite{gcn} (spectral convolution), and \breakabletexttt{GraphSAGE} \cite{graphsage} (sample-based message passing).
\end{itemize}
Together, these variants probe whether local sequential connectivity alone can account for reasoning correctness.

\noindent\textbf{GNN-SR (Relational Graph)}\quad
This model extends the sequential baseline by explicitly encoding semantic and logical relationships between all forward pairs of reasoning chunks. The hypothesis is that correct reasoning forms a globally coherent structure where distant steps reinforce one another, whereas incorrect reasoning exhibits contradictions or semantic drift. The reasoning trace is thus modeled as a fully connected, directed acyclic graph (DAG) in which every earlier chunk can influence all later ones. We evaluate three backbone operators—\breakabletexttt{GNN-SR-GINE}, \breakabletexttt{GNN-SR-NNConv}, and \breakabletexttt{GNN-SR-Transformer}—each differing in how it leverages edge features to modulate message passing.

\begin{itemize}
    \item \textbf{Graph Construction}\,: Directed, fully connected DAG with edges $i \to j$ for all $i < j$.
    \item \textbf{Nodes}\,: Hidden state $h_t$ of each reasoning chunk.
    \item \textbf{Edge Features}\,: Five-dimensional vector per edge $\langle i,j\rangle$ comprising: (1) NLI-style entailment, contradiction, and neutral probabilities; (2) \emph{proximity} $=1-\text{normalized distance}$; and (3) cosine similarity between $h_i$ and $h_j$.
    \item \textbf{Backbones}\,: \breakabletexttt{GINEConv} \cite{gineconv} (MLP-injected messages), \breakabletexttt{NNConv} \cite{nnconv} (edge-conditioned kernels), and \breakabletexttt{TransformerConv} \cite{transformerconv} (edge-aware attention).
\end{itemize}
Collectively, these three variants—\breakabletexttt{GNN-SR-GINE}, \breakabletexttt{GNN-SR-NNConv}, and \breakabletexttt{GNN-SR-Transformer}—test whether relational and semantic coherence across all reasoning steps enhances confidence prediction.

\noindent\textbf{GNN-CD (Confidence Dynamics Graph)}\quad
This variant shifts the focus from the semantic content of the reasoning to the meta-level \emph{dynamics of confidence}. It tests the intuition that correct reasoning is characterized by a stable or increasing sense of certainty, whereas flawed reasoning involves sharp, erratic shifts in confidence. By using features from a probe trained on separate data, this model isolates the predictive power of the confidence trajectory itself.
\begin{itemize}
    \item \textbf{Graph Construction}\,: Directed, fully connected forward DAG with edges $i \to j$ for all $i < j$ (identical reachability pattern to GNN-SR).
    \item \textbf{Nodes}\,: Concatenation of per-chunk \breakabletexttt{PHSV-half} confidence $c_i$ and the probe’s penultimate layer representation.
    \item \textbf{Edge Weights}\,: Scalar values for \emph{every} forward pair $\langle i,j\rangle$ computed as the distributional distance (e.g., Wasserstein or KL) between the chunk-level token log-probability distributions of chunks $i$ and $j$; this quantifies how much the model’s belief state shifts between any two steps, not only adjacent ones.
    \item \textbf{Backbones}\,: \breakabletexttt{GCN2Conv} \cite{gcn2conv}, \breakabletexttt{APPNP} \cite{APPNP}, \breakabletexttt{TAGConv} \cite{tagconv}.
\end{itemize}
The choice of GNN backbone differs across these variants due to the nature of their edge information. Architectures that natively support scalar \emph{edge weights} (such as \breakabletexttt{GCN2Conv} and \breakabletexttt{APPNP}) do not simultaneously support rich, multi-dimensional \emph{edge features}. Conversely, operators designed for edge attributes (like \breakabletexttt{GINEConv} and \breakabletexttt{TransformerConv}) do not accommodate scalar edge weights. This distinction explains why GNN-CD and GNN-SR employ different families of models. For a concise overview of which operators support edge weights versus edge attributes, we refer the reader to the PyTorch Geometric (PyG) documentation\footnote{\url{https://pytorch-geometric.readthedocs.io/en/latest/cheatsheet/gnn_cheatsheet.html}}, which guided our model selection.

We instantiate GNN-CD with two feature strategies—\breakabletexttt{noft} (frozen \breakabletexttt{PHSV-half} features) and \breakabletexttt{ft} (end-to-end fine-tuning of \breakabletexttt{PHSV-half})—and three backbones, with \breakabletexttt{GCN2Conv} further split into \breakabletexttt{same} vs.\ \breakabletexttt{dual} initial-residual schemes. The exact names appearing in tables and plots follow the pattern \breakabletexttt{GNN-CD-\{noft,ft\}-\{GCN2Conv,APPNP,TAGConv\}[-\{same,dual\}]}, yielding:
\begin{itemize}
\item \texttt{GNN-CD-noft-GCN2Conv-same}
\item \texttt{GNN-CD-ft-GCN2Conv-same}
\item \texttt{GNN-CD-noft-GCN2Conv-dual}
\item \texttt{GNN-CD-ft-GCN2Conv-dual}
\item \texttt{GNN-CD-noft-APPNP}
\item \texttt{GNN-CD-ft-APPNP}
\item \texttt{GNN-CD-noft-TAGConv}
\item \texttt{GNN-CD-ft-TAGConv}
\end{itemize}
Here, \breakabletexttt{same} uses each layer’s input as the initial residual ($x_0{=}x$), whereas \breakabletexttt{dual} supplements message passing with a skip from the LRM’s original hidden states, providing a second information stream.

\subsubsection{CE (Chunk Ensemble)}
This method approaches confidence estimation as a classical machine learning problem, training simple classifiers on high-level features derived from the reasoning trace. It follows a strict two-stage training protocol to prevent data leakage. A \breakabletexttt{PHSV-half} model is first trained on the initial 50\% of the data and then frozen as a feature extractor. Using its outputs, we derive per-chunk confidence scores and form a fixed-length trajectory vector $[c_1, c_2, \dots, c_L]$. Standard classifiers—\breakabletexttt{LogisticRegression}, \breakabletexttt{RandomForest}, \breakabletexttt{DecisionTree}, \breakabletexttt{KNN}, and \breakabletexttt{XGBoost}—are trained on this representation, and the best-performing model is reported for each LRM.

\subsubsection{LateFusion}
This model implements a hybrid, dual-stream architecture designed to integrate semantic and confidence-based signals before final prediction. It adheres to a two-stage protocol: a pre-trained \breakabletexttt{PHSV-half} model serves as a feature extractor on the first half of the data, and the \breakabletexttt{LateFusion} model is trained on the held-out half. Each reasoning trace is processed by two parallel streams. The \emph{semantic stream} models the logical flow and content of the argument using the raw hidden states $[h_1, h_2, \dots, h_L]$, while the \emph{dynamics stream} models the evolution of certainty by processing the concatenation of each chunk’s confidence score $c_i$ and its penultimate-layer representation. The fused representations are concatenated and passed through a shared classifier. Variants differ along two axes: (i) whether the \breakabletexttt{PHSV-half} features are frozen (\breakabletexttt{noft}) or fine-tuned jointly (\breakabletexttt{ft}), and (ii) the network architecture used in both streams—\breakabletexttt{MLP}, 1D \breakabletexttt{Conv}, or bidirectional \breakabletexttt{LSTM}. The resulting six variants (\breakabletexttt{LateFusion-noft-MLP}, \breakabletexttt{LateFusion-noft-Conv}, \breakabletexttt{LateFusion-noft-LSTM}, \breakabletexttt{LateFusion-ft-MLP}, \breakabletexttt{LateFusion-ft-Conv}, and \breakabletexttt{LateFusion-ft-LSTM}) test the contribution of fine-tuning and temporal modeling.

\subsubsection{ETTIN}
This model adapts the \breakabletexttt{jhu-clsp/ettin-encoder-17m}, originally designed for hallucination detection, to the task of reasoning confidence estimation. It treats the concatenated prompt and reasoning trace as a single text sequence. The text is fed into the ETTIN encoder, whose token embeddings are mean-pooled to form a single representation summarizing the reasoning process. This vector is passed to an MLP head trained with a binary cross-entropy loss on the overall correctness label. The model thus performs holistic, text-level confidence prediction without explicit step structure, providing a strong baseline for text-encoder-based calibration.

\subsubsection{ETTIN-HGA}
\label{app:ETTIN-HGA-Model-Desc}
This variant extends ETTIN with a hierarchical architecture that explicitly models the structure and quality of intermediate reasoning steps. The reasoning trace is segmented into chunks and concatenated with \breakabletexttt{[SEP]} delimiters (\breakabletexttt{[CLS]} prompt \breakabletexttt{[SEP]} chunk\_1 \breakabletexttt{[SEP]} ... \breakabletexttt{[SEP]} chunk\_n). The ETTIN encoder processes the structured input, and hidden states at each \breakabletexttt{[SEP]} token serve as chunk-level embeddings. A hierarchical gated attention (HGA)  module combines two components: an attention head that models inter-chunk dependencies and a quality head that predicts per-chunk correctness scores, which act as gating weights to emphasize coherent, reliable chunks. The gated, context-aware representations are then mean-pooled and passed through an MLP classifier. Training employs a composite loss combining the final correctness loss with an auxiliary per-chunk loss supervising the quality head. This design allows ETTIN-HGA to explicitly reason about the consistency and reliability of intermediate steps, yielding stronger, structure-aware confidence estimates.

%% file: sections/appendix_evaluation_metrics.tex
\section{Evaluation Metrics}
\label{app:evaluation_metrics}
To comprehensively assess the quality of our confidence estimation framework, we employ a diverse collection of metrics that capture both calibration and discriminative performance. This combination enables a balanced interpretation of model behavior beyond simple accuracy, especially under class imbalance conditions that commonly arise in correctness prediction.

\subsection{Expected Calibration Error (ECE)}
Calibration reflects how well a model’s predicted confidences correspond to actual empirical frequencies. In a well-calibrated system, predictions with confidence $p$ should be correct approximately $p$ fraction of the time. We compute the Expected Calibration Error (ECE) using a standard binning approach. Specifically, the confidence scores of all $n$ samples are divided into $b$ uniform bins $\{B_j\}_{j=1}^{b}$, and the deviation between mean confidence and empirical accuracy within each bin is aggregated as:
\[
\text{ECE} = \sum_{j=1}^{b} \frac{|B_j|}{n} \big| \text{conf}(B_j) - \text{acc}(B_j) \big|
\]
where $\text{conf}(B_j)$ denotes the average predicted confidence in bin $B_j$ and $\text{acc}(B_j)$ the observed accuracy therein. Unless stated otherwise, we use $b=10$. Lower values indicate better alignment between predicted and observed probabilities.

\subsection{Brier Score}
The Brier Score quantifies the mean squared distance between each predicted probability $p_k$ and its ground-truth correctness label $o_k \in \{0,1\}$:
\[
\text{Brier Score} = \frac{1}{N} \sum_{k=1}^{N} (p_k - o_k)^2
\]
This measure simultaneously captures aspects of calibration and sharpness, penalizing both over- and under-confident predictions. A smaller value reflects superior overall reliability.

\subsection{Accuracy (ACC)}
Accuracy represents the percentage of cases in which the model’s predicted answer is correct:
\[
\text{ACC} = \frac{\text{TP} + \text{TN}}{\text{TP} + \text{FP} + \text{FN} + \text{TN}}
\]
While confidence estimation methods typically do not alter the base model’s predictions, reporting ACC provides useful reference for the inherent difficulty of the task and contextualizes other confidence-related metrics.

\subsection{F1 Score}
To comprehensively evaluate discriminative performance, we report the F1 score, which captures the balance between precision and recall. Precision quantifies the proportion of predicted correct cases that are actually correct:
\[
\text{Precision} = \frac{\text{TP}}{\text{TP} + \text{FP}}
\]
while recall (or sensitivity) measures the proportion of truly correct cases that are successfully identified by the model:
\[
\text{Recall} = \frac{\text{TP}}{\text{TP} + \text{FN}}
\]
The F1 score is then defined as the harmonic mean of precision and recall:
\[
\text{F1} = 2 \cdot \frac{\text{Precision} \cdot \text{Recall}}{\text{Precision} + \text{Recall}}
\]
A higher F1 score indicates that the confidence estimator achieves both strong precision and high recall, effectively distinguishing correct predictions while minimizing false alarms and missed detections.

\subsection{Specificity}
Specificity (Spec), also known as the True Negative Rate (TNR), measures the ability of the confidence estimation model to correctly identify instances that are truly incorrect. In other words, it quantifies how effectively the model avoids assigning high confidence to wrong predictions. It is defined as:
\[
\text{Specificity} = \frac{\text{TN}}{\text{TN} + \text{FP}}
\]
where $\text{TN}$ denotes the number of correctly identified incorrect cases and $\text{FP}$ represents the number of incorrect cases mistakenly classified as correct. High specificity indicates that the confidence estimator is conservative—rarely overconfident in wrong predictions—thereby complementing sensitivity (or recall) to provide a more complete view of discriminative reliability.

\subsection{Area Under the Precision–Recall Curve (AUCPR)}
AUCPR condenses the trade-off between precision ($\text{TP}/(\text{TP}+\text{FP})$) and recall ($\text{TP}/(\text{TP}+\text{FN})$) into a single score by integrating over varying discrimination thresholds. It is especially sensitive to performance on the positive (correct) class and therefore more informative than AUROC when class distributions are skewed.

\subsection{Area Under the ROC Curve (AUROC)}
AUROC evaluates the ability of the confidence estimator to discriminate between correct and incorrect answers. It plots the true positive rate ($\text{TPR} = \text{TP}/(\text{TP}+\text{FN})$) against the false positive rate ($\text{FPR} = \text{FP}/(\text{FP}+\text{TN})$) as the confidence threshold varies. An AUROC of 1.0 denotes perfect separability, whereas 0.5 corresponds to chance-level discrimination.

%% file: sections/appendix_optuna.tex
\section{Training and Optimization Details}
\label{app:training_details}

All trainable confidence estimation models in our benchmark were developed using a consistent and rigorous training protocol to ensure a fair comparison. Training was conducted with a fixed batch size of 32. Hyperparameter tuning was performed using the Optuna framework \citep{optuna}, with each model variant undergoing up to 100 trials to find an optimal configuration.

\subsection{Hyperparameter Optimization with Optuna}
We utilized \texttt{Optuna} to systematically explore architectural, training, and regularization parameters. A Tree-structured Parzen Estimator (TPE) sampler was employed to intelligently suggest new configurations based on past results. The objective optimized in each study was a composite score designed to address the central trade-off between discrimination and calibration:

\[
\text{CompositeScore} = \alpha \cdot \text{AUROC} + (1 - \alpha) \cdot (1 - \text{ECE})
\]

where we set $\alpha = 0.6$ to place a slight emphasis on AUROC, prioritizing the discovery of models with strong discriminative power while still imposing a significant penalty for poor calibration.

For final model selection, we imposed an additional practical constraint. A trial was only considered "feasible" if its best-performing epoch also achieved a minimum sensitivity and specificity of 0.50 at its Youden's J optimal threshold. This ensures our final reported models are not only well-balanced in terms of AUROC and ECE, but also demonstrate a tangible predictive ability better than random chance. Among all feasible trials, the one with the highest composite score was selected.

Each trial was trained for a maximum of 200 epochs with an early-stopping patience of 20 epochs based on the composite validation score. To accelerate the search, Optuna’s Median Pruner was applied with a "moderate" schedule to terminate unpromising trials early.

\begin{table*}[t]
\centering
\caption{Hyperparameter search space. All methods inherit the shared space; rows list only method-specific additions.}
\label{tab:all_hparams}
\scriptsize
\begin{tabularx}{\textwidth}{llL}
\toprule
\textbf{Method} & \textbf{Variant} & \textbf{Search space} \\
\midrule
\multicolumn{2}{l}{\textbf{Shared (all methods)}} & 
learning\_rate $\in \{1\text{e-4},\,1\text{e-3}\}$;\; 
weight\_decay $\in \{1\text{e-5},\,1\text{e-4}\}$;\; 
classifier\_layers $\in \{$128,64; 128,32; 64,32; 32,16; 128; 64; 32; 0; 256,128; 512,256; 256; 512$\}$;\; 
classifier\_dropout $\in \{0.1,\,0.25,\,0.4\}$;\; 
budget $[1,\,3.2$M params$]$ \\
\midrule
PIK & -- & -- (no additional hyperparameters) \\
PHSV & -- & -- (no additional hyperparameters) \\
\midrule
SFHS & MLP & -- (uses only shared) \\
     & Conv & conv\_layers $\in \{$32,64; 64,128$\}$;\; kernel\_sizes $\in \{$3,3; 5,3$\}$;\; dropout $\in \{0.1,\,0.25,\,0.4\}$ \\
     & LSTM & hidden\_dim $\in \{16,\,32,\,64\}$;\; num\_layers $\in \{1,2\}$;\; bidirectional $\in \{\texttt{True},\,\texttt{False}\}$;\; dropout $\in \{0.1,\,0.25,\,0.4\}$ \\
\midrule
TLCC & MLP & -- (uses only shared) \\
     & Conv & conv\_layers $\in \{$32,64; 64,128$\}$;\; kernel\_sizes $\in \{$3,3; 5,3$\}$;\; dropout $\in \{0.1,\,0.25,\,0.4\}$ \\
     & LSTM & hidden\_dim $\in \{16,\,32,\,64\}$;\; num\_layers $\in \{1,2\}$;\; bidirectional $\in \{\texttt{True},\,\texttt{False}\}$;\; dropout $\in \{0.1,\,0.25,\,0.4\}$ \\
\midrule
LateFusion & MLP & semantic\_hidden $\in \{$128,64; 64; 0$\}$;\; dynamics\_hidden $\in \{$64,32; 32; 0$\}$ \\
           & Conv & semantic\_conv $\in \{$32,64$\}$;\; semantic\_kernels $\in \{$3,3; 5,3$\}$;\; dynamics\_conv $\in \{$16,32; 32,32$\}$;\; dynamics\_kernels $\in \{$3,3$\}$;\; dropout $\in \{0.1,\,0.25,\,0.4\}$ \\
           & LSTM & semantic\_hidden\_dim $\in \{32,\,64\}$;\; semantic\_num\_layers $=1$;\; semantic\_bidirectional $=\texttt{True}$;\; dynamics\_hidden\_dim $\in \{16,\,32\}$;\; dynamics\_num\_layers $=1$;\; dynamics\_bidirectional $=\texttt{True}$;\; dropout $\in \{0.1,\,0.25,\,0.4\}$ \\
\midrule
GNN\_EdgeAttr & GINE & hidden\_dim $\in \{64,\,128,\,256\}$;\; num\_layers $\in \{1,2\}$;\; pooling $\in \{\texttt{mean},\texttt{max},\texttt{sum},\texttt{attention},\texttt{last\_node}\}$;\; edge\_nn $\in \{\texttt{linear},\texttt{mlp\_small},\texttt{mlp\_medium}\}$ \\
              & Transformer & hidden\_dim $\in \{64,\,128,\,256\}$;\; num\_layers $\in \{1,2\}$;\; pooling as above;\; heads $\in \{2,4,8\}$;\; concat $\in \{\texttt{True},\,\texttt{False}\}$ \\
              & NNConv & hidden\_dim $\in \{32,\,64,\,128,\,256\}$;\; num\_layers $\in \{1,2\}$;\; pooling as above;\; edge\_nn=\texttt{linear} \\
\midrule
GNN\_Conf & GCN2\_Same & hidden\_dim $\in \{128,\,256,\,512\}$;\; num\_layers $\in \{1,2\}$;\; pooling as above;\; $\alpha \in \{0.1,0.3,0.5\}$;\; $\theta \in \{1.0,1.5,2.0\}$;\; shared\_weights=True \\
          & GCN2\_Dual & hidden\_dim $\in \{128,\,256\}$;\; num\_layers $\in \{1,2\}$;\; pooling as above;\; $\alpha, \theta$ as above;\; shared\_weights=True \\
          & TAGConv & hidden\_dim $\in \{128,\,256,\,512\}$;\; num\_layers $\in \{1,2\}$;\; $K \in \{2,3\}$;\; pooling as above \\
          & APPNP & hidden\_dim $\in \{128,\,256,\,512\}$;\; num\_layers $\in \{1,2\}$;\; $K \in \{2,3\}$;\; appnp\_alpha $\in \{0.1,0.5,0.9\}$;\; pooling as above \\
\midrule
GNN\_SB & GCN & gnn\_type=gcn;\; hidden\_dim $\in \{64,128,256\}$;\; num\_layers $\in \{1,2,3,4\}$;\; pooling $\in \{\texttt{mean},\texttt{max},\texttt{sum}\}$ \\
        & GAT & gnn\_type=gat;\; hidden\_dim as above;\; num\_layers as above;\; pooling as above;\; heads $\in \{1,2,4\}$;\; concat $\in \{\texttt{True},\texttt{False}\}$ \\
        & GraphSAGE & gnn\_type=graphsage;\; hidden\_dim as above;\; num\_layers as above;\; pooling as above;\; aggr $\in \{\texttt{mean},\texttt{max},\texttt{add}\}$ \\
\midrule
ETTIN & -- & -- (no additional hyperparameters) \\
ETTIN-HGA & -- & attention\_dropout $\in \{0.1,\,0.25,\,0.4\}$;\; quality\_layers same as classifier\_layers above \\
\bottomrule
\end{tabularx}
\end{table*}

\subsection{Hyperparameter Search Spaces}
The specific hyperparameter search spaces for each model family are detailed in Table~\ref{tab:all_hparams}. All methods shared a common search space for the final classifier's architecture (e.g., \texttt{classifier\_layers}, \texttt{classifier\_dropout}), learning rate, and weight decay. To ensure a fair comparison of architectural efficiency and prevent model complexity from being conflated with raw parameter count, all configurations were constrained to a maximum of 3.2 million trainable parameters. Any trial suggesting a model outside this budget was immediately discarded. Upon completion of each 100-trial study, the best feasible configuration was selected as the final model for evaluation.

%% file: sections/appendix_extended_results.tex
\section{Comprehensive Results}
\label{app:comprehensive_tables}
This appendix presents complementary views of performance so readers can inspect both granular and aggregated behavior across models, datasets, and methods. We report \emph{ECE}, \emph{Brier}, \emph{Acc}, \emph{F1}, \emph{Spec}, \emph{AUCPR}, and \emph{AUROC}. Lower is better for ECE and Brier; higher is better for the others. Within any comparison group (e.g., a given dataset inside a per-LLM table), the best value per metric is shown in \textbf{bold} using the appropriate direction. Tables \ref{tab:Phi_4_mini_flash_reasoning_dataset_method}, \ref{tab:Qwen3_8B_dataset_method}, \ref{tab:Qwen3_14B_dataset_method}, \ref{tab:Magistral_Small_2506_dataset_method}, \ref{tab:QwQ_32B_dataset_method}, and \ref{tab:EXAONE_Deep_32B_dataset_method} enumerate, for each LLM, every method’s performance on each test dataset without aggregation, supporting fine-grained, within-model comparisons. Tables \ref{tab:Phi_4_mini_flash_reasoning_method_avg}, \ref{tab:Qwen3_8B_method_avg}, \ref{tab:Qwen3_14B_method_avg}, \ref{tab:Magistral_Small_2506_method_avg}, \ref{tab:QwQ_32B_method_avg}, and \ref{tab:EXAONE_Deep_32B_method_avg} summarize, for each LLM, the \emph{unweighted} mean $\pm$ standard deviation of each method across all datasets, providing a concise per-model overview that averages out dataset-level variability. Table \ref{tab:dataset_method_avg} aggregates each \emph{dataset--method} pair across LLMs, reporting the \emph{unweighted} mean $\pm$ standard deviation to indicate which methods generalize well on a given dataset independent of the underlying model. Table \ref{tab:llm_method_avg} groups each \emph{LLM--method} pair across datasets as \emph{unweighted} mean $\pm$ standard deviation, emphasizing which methods work best for a particular LLM after averaging out dataset effects. Finally, Table \ref{tab:method_only_appendix} provides a method-only view using a two-stage \emph{unweighted} average: first average a method across datasets within each LLM, then average those LLM-level means across LLMs; we report the corresponding standard deviations.

\onecolumn
\begingroup
\scriptsize
\setlength{\tabcolsep}{3pt}
\setlength{\LTleft}{\fill}
\setlength{\LTright}{\fill}
\setlength{\LTcapwidth}{\textwidth}
\begin{longtable}{@{}l l c c c c c c c@{}}
\caption{Performance metrics for \texttt{Phi-4-mini-flash-reasoning} showing results per method within each test dataset. Each metric value represents the performance of the specified method on the specified dataset for this LLM. \textbf{Bold} entries mark the best-performing method for each metric within each dataset.}\label{tab:Phi_4_mini_flash_reasoning_dataset_method}\\
\toprule
\textbf{Dataset} & \textbf{Method} & \textbf{ECE}$\downarrow$ & \textbf{Brier}$\downarrow$ & \textbf{Acc}$\uparrow$ & \textbf{F1}$\uparrow$ & \textbf{Spec}$\uparrow$ & \textbf{AUCPR}$\uparrow$ & \textbf{AUROC}$\uparrow$ \\
\midrule
\endfirsthead
\multicolumn{9}{l}{\scriptsize\emph{(Continued from previous page)}}\\
\toprule
\textbf{Dataset} & \textbf{Method} & \textbf{ECE}$\downarrow$ & \textbf{Brier}$\downarrow$ & \textbf{Acc}$\uparrow$ & \textbf{F1}$\uparrow$ & \textbf{Spec}$\uparrow$ & \textbf{AUCPR}$\uparrow$ & \textbf{AUROC}$\uparrow$ \\
\midrule
\endhead
\midrule
\multicolumn{9}{r}{\scriptsize\emph{(Continued on next page)}}\\
\endfoot
\bottomrule
\endlastfoot
BBH & YVCE & 0.171 & 0.245 & 0.648 & 0.767 & 0.295 & 0.845 & 0.625 \\
 & TLCC-MLP & 0.110 & 0.171 & 0.723 & 0.809 & 0.577 & 0.906 & 0.755 \\
 & TLCC-CONV & 0.099 & 0.169 & 0.719 & 0.807 & 0.557 & 0.907 & 0.756 \\
 & TLCC-LSTM & 0.083 & 0.164 & 0.741 & 0.827 & 0.530 & 0.905 & 0.757 \\
 & P(IK) & 0.033 & 0.171 & \textbf{0.758} & \textbf{0.862} & 0.012 & 0.868 & 0.682 \\
 & PHSV-half & 0.078 & 0.197 & 0.710 & 0.817 & 0.212 & 0.830 & 0.683 \\
 & PHSV & 0.117 & 0.204 & 0.691 & 0.801 & 0.231 & 0.837 & 0.693 \\
 & SFHS-MLP & 0.076 & 0.165 & 0.719 & 0.813 & 0.449 & 0.889 & 0.744 \\
\end{longtable}
\endgroup
\twocolumn

\onecolumn
\begingroup
\scriptsize
\setlength{\tabcolsep}{3pt}
\setlength{\LTleft}{\fill}
\setlength{\LTright}{\fill}
\setlength{\LTcapwidth}{\textwidth}
\begin{longtable}{@{}l l c c c c c c c@{}}
\caption{Performance metrics for \texttt{Qwen3-8B} showing results per method within each test dataset. Each metric value represents the performance of the specified method on the specified dataset for this LLM. \textbf{Bold} entries mark the best-performing method for each metric within each dataset.}\label{tab:Qwen3_8B_dataset_method}\\
\toprule
\textbf{Dataset} & \textbf{Method} & \textbf{ECE}$\downarrow$ & \textbf{Brier}$\downarrow$ & \textbf{Acc}$\uparrow$ & \textbf{F1}$\uparrow$ & \textbf{Spec}$\uparrow$ & \textbf{AUCPR}$\uparrow$ & \textbf{AUROC}$\uparrow$ \\
\midrule
\endfirsthead
\multicolumn{9}{l}{\scriptsize\emph{(Continued from previous page)}}\\
\toprule
\textbf{Dataset} & \textbf{Method} & \textbf{ECE}$\downarrow$ & \textbf{Brier}$\downarrow$ & \textbf{Acc}$\uparrow$ & \textbf{F1}$\uparrow$ & \textbf{Spec}$\uparrow$ & \textbf{AUCPR}$\uparrow$ & \textbf{AUROC}$\uparrow$ \\
\midrule
\endhead
\midrule
\multicolumn{9}{r}{\scriptsize\emph{(Continued on next page)}}\\
\endfoot
\bottomrule
\endlastfoot
BBH & YVCE & \textbf{0.018} & 0.140 & 0.824 & 0.903 & 0.032 & 0.897 & 0.637 \\
 & TLCC-MLP & 0.173 & 0.181 & 0.724 & 0.824 & 0.424 & 0.883 & 0.654 \\
 & TLCC-CONV & 0.139 & 0.179 & 0.732 & 0.831 & 0.391 & 0.888 & 0.650 \\
 & TLCC-LSTM & 0.027 & 0.137 & 0.814 & 0.894 & 0.171 & 0.892 & 0.697 \\
 & P(IK) & 0.034 & 0.142 & 0.821 & 0.901 & 0.012 & 0.881 & 0.634 \\
 & PHSV-half & 0.095 & 0.187 & 0.741 & 0.846 & 0.127 & 0.833 & 0.672 \\
 & PHSV & 0.060 & 0.173 & 0.754 & 0.854 & 0.140 & 0.849 & 0.685 \\
 & SFHS-MLP & 0.072 & 0.135 & 0.815 & 0.891 & 0.359 & 0.915 & 0.742 \\
\end{longtable}
\endgroup
\twocolumn

\onecolumn
\begingroup
\scriptsize
\setlength{\tabcolsep}{3pt}
\setlength{\LTleft}{\fill}
\setlength{\LTright}{\fill}
\setlength{\LTcapwidth}{\textwidth}
\begin{longtable}{@{}l l c c c c c c c@{}}
\caption{Performance metrics for \texttt{Qwen3-14B} showing results per method within each test dataset. Each metric value represents the performance of the specified method on the specified dataset for this LLM. \textbf{Bold} entries mark the best-performing method for each metric within each dataset.}\label{tab:Qwen3_14B_dataset_method}\\
\toprule
\textbf{Dataset} & \textbf{Method} & \textbf{ECE}$\downarrow$ & \textbf{Brier}$\downarrow$ & \textbf{Acc}$\uparrow$ & \textbf{F1}$\uparrow$ & \textbf{Spec}$\uparrow$ & \textbf{AUCPR}$\uparrow$ & \textbf{AUROC}$\uparrow$ \\
\midrule
\endfirsthead
\multicolumn{9}{l}{\scriptsize\emph{(Continued from previous page)}}\\
\toprule
\textbf{Dataset} & \textbf{Method} & \textbf{ECE}$\downarrow$ & \textbf{Brier}$\downarrow$ & \textbf{Acc}$\uparrow$ & \textbf{F1}$\uparrow$ & \textbf{Spec}$\uparrow$ & \textbf{AUCPR}$\uparrow$ & \textbf{AUROC}$\uparrow$ \\
\midrule
\endhead
\midrule
\multicolumn{9}{r}{\scriptsize\emph{(Continued on next page)}}\\
\endfoot
\bottomrule
\endlastfoot
BBH & YVCE & 0.078 & 0.133 & \textbf{0.833} & \textbf{0.908} & 0.027 & \textbf{0.933} & 0.686 \\
 & TLCC-MLP & 0.134 & 0.154 & 0.788 & 0.874 & 0.308 & 0.911 & 0.689 \\
 & TLCC-CONV & 0.082 & 0.151 & 0.791 & 0.878 & 0.234 & 0.908 & 0.676 \\
 & TLCC-LSTM & 0.078 & 0.143 & 0.796 & 0.879 & 0.308 & 0.923 & 0.720 \\
 & P(IK) & 0.114 & 0.149 & 0.831 & 0.908 & 0.000 & 0.889 & 0.643 \\
 & PHSV-half & 0.057 & 0.171 & 0.745 & 0.847 & 0.170 & 0.862 & 0.690 \\
 & PHSV & 0.075 & 0.180 & 0.737 & 0.842 & 0.152 & 0.848 & 0.678 \\
 & SFHS-MLP & 0.078 & 0.141 & 0.802 & 0.885 & 0.244 & 0.921 & 0.724 \\
\end{longtable}
\endgroup
\twocolumn

\onecolumn
\begingroup
\scriptsize
\setlength{\tabcolsep}{3pt}
\setlength{\LTleft}{\fill}
\setlength{\LTright}{\fill}
\setlength{\LTcapwidth}{\textwidth}
\begin{longtable}{@{}l l c c c c c c c@{}}
\caption{Performance metrics for \texttt{Magistral-Small-2506} showing results per method within each test dataset. Each metric value represents the performance of the specified method on the specified dataset for this LLM. \textbf{Bold} entries mark the best-performing method for each metric within each dataset.}\label{tab:Magistral_Small_2506_dataset_method}\\
\toprule
\textbf{Dataset} & \textbf{Method} & \textbf{ECE}$\downarrow$ & \textbf{Brier}$\downarrow$ & \textbf{Acc}$\uparrow$ & \textbf{F1}$\uparrow$ & \textbf{Spec}$\uparrow$ & \textbf{AUCPR}$\uparrow$ & \textbf{AUROC}$\uparrow$ \\
\midrule
\endfirsthead
\multicolumn{9}{l}{\scriptsize\emph{(Continued from previous page)}}\\
\toprule
\textbf{Dataset} & \textbf{Method} & \textbf{ECE}$\downarrow$ & \textbf{Brier}$\downarrow$ & \textbf{Acc}$\uparrow$ & \textbf{F1}$\uparrow$ & \textbf{Spec}$\uparrow$ & \textbf{AUCPR}$\uparrow$ & \textbf{AUROC}$\uparrow$ \\
\midrule
\endhead
\midrule
\multicolumn{9}{r}{\scriptsize\emph{(Continued on next page)}}\\
\endfoot
\bottomrule
\endlastfoot
BBH & YVCE & 0.507 & 0.517 & 0.398 & 0.546 & 0.059 & 0.598 & 0.466 \\
 & TLCC-MLP & 0.041 & 0.173 & 0.764 & 0.669 & 0.839 & 0.676 & 0.800 \\
 & TLCC-CONV & 0.055 & 0.173 & 0.765 & 0.670 & 0.840 & 0.686 & 0.805 \\
 & TLCC-LSTM & 0.045 & 0.170 & 0.766 & 0.677 & 0.833 & 0.697 & 0.807 \\
 & P(IK) & 0.075 & 0.217 & 0.649 & 0.608 & 0.608 & 0.511 & 0.693 \\
 & PHSV-half & 0.087 & 0.214 & 0.696 & 0.517 & 0.817 & 0.537 & 0.675 \\
 & PHSV & 0.068 & 0.209 & 0.685 & 0.510 & 0.798 & 0.537 & 0.681 \\
 & SFHS-MLP & 0.036 & 0.183 & 0.739 & 0.631 & 0.829 & 0.681 & 0.774 \\
\end{longtable}
\endgroup
\twocolumn

\onecolumn
\begingroup
\scriptsize
\setlength{\tabcolsep}{3pt}
\setlength{\LTleft}{\fill}
\setlength{\LTright}{\fill}
\setlength{\LTcapwidth}{\textwidth}
\begin{longtable}{@{}l l c c c c c c c@{}}
\caption{Performance metrics for \texttt{QwQ-32B} showing results per method within each test dataset. Each metric value represents the performance of the specified method on the specified dataset for this LLM. \textbf{Bold} entries mark the best-performing method for each metric within each dataset.}\label{tab:QwQ_32B_dataset_method}\\
\toprule
\textbf{Dataset} & \textbf{Method} & \textbf{ECE}$\downarrow$ & \textbf{Brier}$\downarrow$ & \textbf{Acc}$\uparrow$ & \textbf{F1}$\uparrow$ & \textbf{Spec}$\uparrow$ & \textbf{AUCPR}$\uparrow$ & \textbf{AUROC}$\uparrow$ \\
\midrule
\endfirsthead
\multicolumn{9}{l}{\scriptsize\emph{(Continued from previous page)}}\\
\toprule
\textbf{Dataset} & \textbf{Method} & \textbf{ECE}$\downarrow$ & \textbf{Brier}$\downarrow$ & \textbf{Acc}$\uparrow$ & \textbf{F1}$\uparrow$ & \textbf{Spec}$\uparrow$ & \textbf{AUCPR}$\uparrow$ & \textbf{AUROC}$\uparrow$ \\
\midrule
\endhead
\midrule
\multicolumn{9}{r}{\scriptsize\emph{(Continued on next page)}}\\
\endfoot
\bottomrule
\endlastfoot
BBH & YVCE & 0.047 & 0.115 & 0.855 & \textbf{0.922} & 0.025 & \textbf{0.940} & 0.690 \\
 & TLCC-MLP & 0.211 & 0.183 & 0.696 & 0.806 & 0.463 & 0.918 & 0.663 \\
 & TLCC-CONV & 0.173 & 0.191 & 0.712 & 0.820 & 0.370 & 0.897 & 0.607 \\
 & TLCC-LSTM & 0.179 & 0.175 & 0.733 & 0.834 & 0.404 & 0.913 & 0.653 \\
 & P(IK) & 0.089 & 0.136 & 0.813 & 0.893 & 0.221 & 0.914 & 0.679 \\
 & PHSV-half & 0.061 & 0.169 & 0.763 & 0.862 & 0.096 & 0.845 & 0.651 \\
 & PHSV & 0.048 & 0.164 & 0.766 & 0.864 & 0.120 & 0.874 & 0.677 \\
 & SFHS-MLP & 0.095 & 0.133 & 0.819 & 0.895 & 0.314 & 0.931 & 0.718 \\
\end{longtable}
\endgroup
\twocolumn

\onecolumn
\begingroup
\scriptsize
\setlength{\tabcolsep}{3pt}
\setlength{\LTleft}{\fill}
\setlength{\LTright}{\fill}
\setlength{\LTcapwidth}{\textwidth}
\begin{longtable}{@{}l l c c c c c c c@{}}
\caption{Performance metrics for \texttt{EXAONE-Deep-32B} showing results per method within each test dataset. Each metric value represents the performance of the specified method on the specified dataset for this LLM. \textbf{Bold} entries mark the best-performing method for each metric within each dataset.}\label{tab:EXAONE_Deep_32B_dataset_method}\\
\toprule
\textbf{Dataset} & \textbf{Method} & \textbf{ECE}$\downarrow$ & \textbf{Brier}$\downarrow$ & \textbf{Acc}$\uparrow$ & \textbf{F1}$\uparrow$ & \textbf{Spec}$\uparrow$ & \textbf{AUCPR}$\uparrow$ & \textbf{AUROC}$\uparrow$ \\
\midrule
\endfirsthead
\multicolumn{9}{l}{\scriptsize\emph{(Continued from previous page)}}\\
\toprule
\textbf{Dataset} & \textbf{Method} & \textbf{ECE}$\downarrow$ & \textbf{Brier}$\downarrow$ & \textbf{Acc}$\uparrow$ & \textbf{F1}$\uparrow$ & \textbf{Spec}$\uparrow$ & \textbf{AUCPR}$\uparrow$ & \textbf{AUROC}$\uparrow$ \\
\midrule
\endhead
\midrule
\multicolumn{9}{r}{\scriptsize\emph{(Continued on next page)}}\\
\endfoot
\bottomrule
\endlastfoot
BBH & YVCE & 0.056 & 0.145 & 0.813 & 0.892 & 0.228 & 0.928 & 0.727 \\
 & TLCC-MLP & 0.089 & 0.141 & 0.808 & 0.887 & 0.327 & 0.902 & 0.685 \\
 & TLCC-CONV & 0.057 & 0.134 & 0.817 & 0.894 & 0.259 & 0.915 & 0.702 \\
 & TLCC-LSTM & 0.046 & 0.125 & \textbf{0.841} & \textbf{0.911} & 0.153 & 0.923 & 0.729 \\
 & P(IK) & 0.049 & 0.140 & 0.831 & 0.907 & 0.019 & 0.870 & 0.596 \\
 & PHSV-half & 0.264 & 0.240 & 0.644 & 0.742 & 0.526 & 0.783 & 0.633 \\
 & PHSV & 0.264 & 0.242 & 0.635 & 0.731 & 0.542 & 0.769 & 0.623 \\
 & SFHS-MLP & 0.051 & 0.133 & 0.820 & 0.897 & 0.201 & 0.921 & 0.728 \\
\end{longtable}
\endgroup
\twocolumn

\onecolumn
\begingroup
\scriptsize
\setlength{\tabcolsep}{3pt}
\setlength{\LTleft}{\fill}
\setlength{\LTright}{\fill}
\setlength{\LTcapwidth}{\textwidth}
\begin{longtable}{@{}l c c c c c c c@{}}
\caption{Dataset-averaged performance metrics for \texttt{Phi-4-mini-flash-reasoning}. Each metric is the \emph{unweighted} mean $\pm$ standard deviation of a method across all test datasets for this LLM. \textbf{Bold} entries mark the best method per metric across datasets.}\label{tab:Phi_4_mini_flash_reasoning_method_avg}\\
\toprule
\textbf{Method} & \textbf{ECE}$\downarrow$ & \textbf{Brier}$\downarrow$ & \textbf{Acc}$\uparrow$ & \textbf{F1}$\uparrow$ & \textbf{Spec}$\uparrow$ & \textbf{AUCPR}$\uparrow$ & \textbf{AUROC}$\uparrow$ \\
\midrule
\endfirsthead
\multicolumn{8}{l}{\scriptsize\emph{(Continued from previous page)}}\\
\toprule
\textbf{Method} & \textbf{ECE}$\downarrow$ & \textbf{Brier}$\downarrow$ & \textbf{Acc}$\uparrow$ & \textbf{F1}$\uparrow$ & \textbf{Spec}$\uparrow$ & \textbf{AUCPR}$\uparrow$ & \textbf{AUROC}$\uparrow$ \\
\midrule
\endhead
\midrule
\multicolumn{8}{r}{\scriptsize\emph{(Continued on next page)}}\\
\endfoot
\bottomrule
\endlastfoot
CE-DT & 0.181 {\tiny $\pm$0.11} & 0.252 {\tiny $\pm$0.08} & 0.588 {\tiny $\pm$0.18} & 0.607 {\tiny $\pm$0.23} & 0.355 {\tiny $\pm$0.11} & 0.600 {\tiny $\pm$0.28} & 0.602 {\tiny $\pm$0.07} \\
CE-KNN & 0.176 {\tiny $\pm$0.10} & 0.239 {\tiny $\pm$0.08} & 0.626 {\tiny $\pm$0.15} & 0.591 {\tiny $\pm$0.25} & 0.469 {\tiny $\pm$0.13} & 0.586 {\tiny $\pm$0.30} & 0.618 {\tiny $\pm$0.09} \\
CE-LogReg & 0.224 {\tiny $\pm$0.13} & 0.249 {\tiny $\pm$0.07} & 0.513 {\tiny $\pm$0.26} & 0.644 {\tiny $\pm$0.23} & 0.002 {\tiny $\pm$0.00} & 0.601 {\tiny $\pm$0.30} & 0.630 {\tiny $\pm$0.11} \\
CE-RF & 0.151 {\tiny $\pm$0.13} & 0.222 {\tiny $\pm$0.08} & 0.613 {\tiny $\pm$0.18} & 0.606 {\tiny $\pm$0.25} & 0.388 {\tiny $\pm$0.16} & 0.600 {\tiny $\pm$0.30} & 0.634 {\tiny $\pm$0.10} \\
CE-XGB & 0.153 {\tiny $\pm$0.13} & 0.223 {\tiny $\pm$0.08} & 0.616 {\tiny $\pm$0.17} & 0.605 {\tiny $\pm$0.25} & 0.404 {\tiny $\pm$0.16} & 0.601 {\tiny $\pm$0.30} & 0.632 {\tiny $\pm$0.10} \\
ETTIN & 0.149 {\tiny $\pm$0.09} & 0.207 {\tiny $\pm$0.08} & 0.648 {\tiny $\pm$0.17} & 0.606 {\tiny $\pm$0.26} & 0.485 {\tiny $\pm$0.20} & 0.625 {\tiny $\pm$0.30} & 0.634 {\tiny $\pm$0.11} \\
ETTIN-HGA & \textbf{0.106} {\tiny $\pm$0.09} & \textbf{0.194} {\tiny $\pm$0.08} & \textbf{0.685} {\tiny $\pm$0.16} & 0.613 {\tiny $\pm$0.28} & 0.469 {\tiny $\pm$0.28} & 0.630 {\tiny $\pm$0.31} & 0.641 {\tiny $\pm$0.12} \\
GNN-CD-ft-APPNP & 0.164 {\tiny $\pm$0.11} & 0.223 {\tiny $\pm$0.07} & 0.641 {\tiny $\pm$0.15} & 0.602 {\tiny $\pm$0.25} & 0.482 {\tiny $\pm$0.18} & 0.611 {\tiny $\pm$0.29} & 0.616 {\tiny $\pm$0.08} \\
\end{longtable}
\endgroup
\twocolumn

\onecolumn
\begingroup
\scriptsize
\setlength{\tabcolsep}{3pt}
\setlength{\LTleft}{\fill}
\setlength{\LTright}{\fill}
\setlength{\LTcapwidth}{\textwidth}
\begin{longtable}{@{}l c c c c c c c@{}}
\caption{Dataset-averaged performance metrics for \texttt{Qwen3-8B}. Each metric is the \emph{unweighted} mean $\pm$ standard deviation of a method across all test datasets for this LLM. \textbf{Bold} entries mark the best method per metric across datasets.}\label{tab:Qwen3_8B_method_avg}\\
\toprule
\textbf{Method} & \textbf{ECE}$\downarrow$ & \textbf{Brier}$\downarrow$ & \textbf{Acc}$\uparrow$ & \textbf{F1}$\uparrow$ & \textbf{Spec}$\uparrow$ & \textbf{AUCPR}$\uparrow$ & \textbf{AUROC}$\uparrow$ \\
\midrule
\endfirsthead
\multicolumn{8}{l}{\scriptsize\emph{(Continued from previous page)}}\\
\toprule
\textbf{Method} & \textbf{ECE}$\downarrow$ & \textbf{Brier}$\downarrow$ & \textbf{Acc}$\uparrow$ & \textbf{F1}$\uparrow$ & \textbf{Spec}$\uparrow$ & \textbf{AUCPR}$\uparrow$ & \textbf{AUROC}$\uparrow$ \\
\midrule
\endhead
\midrule
\multicolumn{8}{r}{\scriptsize\emph{(Continued on next page)}}\\
\endfoot
\bottomrule
\endlastfoot
CE-DT & 0.209 {\tiny $\pm$0.18} & 0.265 {\tiny $\pm$0.14} & 0.634 {\tiny $\pm$0.20} & 0.723 {\tiny $\pm$0.20} & 0.217 {\tiny $\pm$0.09} & 0.686 {\tiny $\pm$0.26} & 0.636 {\tiny $\pm$0.10} \\
CE-KNN & 0.204 {\tiny $\pm$0.19} & 0.257 {\tiny $\pm$0.14} & 0.610 {\tiny $\pm$0.22} & 0.725 {\tiny $\pm$0.20} & 0.073 {\tiny $\pm$0.04} & 0.693 {\tiny $\pm$0.26} & 0.661 {\tiny $\pm$0.12} \\
CE-LogReg & 0.241 {\tiny $\pm$0.21} & 0.278 {\tiny $\pm$0.15} & 0.596 {\tiny $\pm$0.23} & 0.723 {\tiny $\pm$0.20} & 0.000 & 0.698 {\tiny $\pm$0.26} & 0.684 {\tiny $\pm$0.12} \\
CE-RF & 0.201 {\tiny $\pm$0.20} & 0.253 {\tiny $\pm$0.14} & 0.600 {\tiny $\pm$0.23} & 0.724 {\tiny $\pm$0.20} & 0.019 {\tiny $\pm$0.01} & 0.699 {\tiny $\pm$0.27} & \textbf{0.687} {\tiny $\pm$0.13} \\
CE-XGB & 0.204 {\tiny $\pm$0.19} & 0.259 {\tiny $\pm$0.14} & 0.616 {\tiny $\pm$0.21} & 0.722 {\tiny $\pm$0.20} & 0.116 {\tiny $\pm$0.05} & 0.683 {\tiny $\pm$0.28} & 0.663 {\tiny $\pm$0.13} \\
ETTIN & 0.164 {\tiny $\pm$0.20} & 0.224 {\tiny $\pm$0.16} & 0.679 {\tiny $\pm$0.22} & 0.752 {\tiny $\pm$0.21} & 0.191 {\tiny $\pm$0.13} & 0.722 {\tiny $\pm$0.27} & 0.649 {\tiny $\pm$0.08} \\
ETTIN-HGA & 0.164 {\tiny $\pm$0.16} & 0.214 {\tiny $\pm$0.13} & 0.670 {\tiny $\pm$0.24} & \textbf{0.757} {\tiny $\pm$0.22} & 0.085 {\tiny $\pm$0.07} & 0.729 {\tiny $\pm$0.27} & 0.653 {\tiny $\pm$0.09} \\
GNN-CD-ft-APPNP & 0.385 {\tiny $\pm$0.11} & 0.384 {\tiny $\pm$0.11} & 0.606 {\tiny $\pm$0.11} & 0.552 {\tiny $\pm$0.25} & 0.678 {\tiny $\pm$0.21} & 0.730 {\tiny $\pm$0.25} & 0.591 {\tiny $\pm$0.08} \\
\end{longtable}
\endgroup
\twocolumn

\onecolumn
\begingroup
\scriptsize
\setlength{\tabcolsep}{3pt}
\setlength{\LTleft}{\fill}
\setlength{\LTright}{\fill}
\setlength{\LTcapwidth}{\textwidth}
\begin{longtable}{@{}l c c c c c c c@{}}
\caption{Dataset-averaged performance metrics for \texttt{Qwen3-14B}. Each metric is the \emph{unweighted} mean $\pm$ standard deviation of a method across all test datasets for this LLM. \textbf{Bold} entries mark the best method per metric across datasets.}\label{tab:Qwen3_14B_method_avg}\\
\toprule
\textbf{Method} & \textbf{ECE}$\downarrow$ & \textbf{Brier}$\downarrow$ & \textbf{Acc}$\uparrow$ & \textbf{F1}$\uparrow$ & \textbf{Spec}$\uparrow$ & \textbf{AUCPR}$\uparrow$ & \textbf{AUROC}$\uparrow$ \\
\midrule
\endfirsthead
\multicolumn{8}{l}{\scriptsize\emph{(Continued from previous page)}}\\
\toprule
\textbf{Method} & \textbf{ECE}$\downarrow$ & \textbf{Brier}$\downarrow$ & \textbf{Acc}$\uparrow$ & \textbf{F1}$\uparrow$ & \textbf{Spec}$\uparrow$ & \textbf{AUCPR}$\uparrow$ & \textbf{AUROC}$\uparrow$ \\
\midrule
\endhead
\midrule
\multicolumn{8}{r}{\scriptsize\emph{(Continued on next page)}}\\
\endfoot
\bottomrule
\endlastfoot
CE-DT & 0.198 {\tiny $\pm$0.20} & 0.257 {\tiny $\pm$0.15} & 0.633 {\tiny $\pm$0.22} & 0.726 {\tiny $\pm$0.21} & 0.191 {\tiny $\pm$0.10} & 0.675 {\tiny $\pm$0.27} & 0.616 {\tiny $\pm$0.08} \\
CE-KNN & 0.180 {\tiny $\pm$0.21} & 0.243 {\tiny $\pm$0.15} & 0.621 {\tiny $\pm$0.24} & 0.734 {\tiny $\pm$0.21} & 0.047 {\tiny $\pm$0.02} & 0.699 {\tiny $\pm$0.28} & 0.665 {\tiny $\pm$0.12} \\
CE-LogReg & 0.181 {\tiny $\pm$0.22} & 0.244 {\tiny $\pm$0.16} & 0.636 {\tiny $\pm$0.23} & 0.737 {\tiny $\pm$0.21} & 0.116 {\tiny $\pm$0.08} & 0.700 {\tiny $\pm$0.29} & \textbf{0.673} {\tiny $\pm$0.13} \\
CE-RF & 0.178 {\tiny $\pm$0.21} & 0.243 {\tiny $\pm$0.15} & 0.626 {\tiny $\pm$0.24} & 0.737 {\tiny $\pm$0.21} & 0.055 {\tiny $\pm$0.03} & 0.694 {\tiny $\pm$0.28} & 0.664 {\tiny $\pm$0.11} \\
CE-XGB & 0.181 {\tiny $\pm$0.21} & 0.247 {\tiny $\pm$0.15} & 0.630 {\tiny $\pm$0.23} & 0.735 {\tiny $\pm$0.21} & 0.086 {\tiny $\pm$0.04} & 0.690 {\tiny $\pm$0.28} & 0.654 {\tiny $\pm$0.11} \\
ETTIN & 0.178 {\tiny $\pm$0.22} & 0.229 {\tiny $\pm$0.18} & 0.660 {\tiny $\pm$0.26} & 0.762 {\tiny $\pm$0.22} & 0.063 {\tiny $\pm$0.07} & 0.721 {\tiny $\pm$0.29} & 0.649 {\tiny $\pm$0.11} \\
ETTIN-HGA & 0.157 {\tiny $\pm$0.20} & 0.216 {\tiny $\pm$0.16} & 0.685 {\tiny $\pm$0.23} & 0.762 {\tiny $\pm$0.21} & 0.150 {\tiny $\pm$0.15} & 0.742 {\tiny $\pm$0.27} & 0.658 {\tiny $\pm$0.09} \\
GNN-CD-ft-APPNP & 0.186 {\tiny $\pm$0.09} & 0.214 {\tiny $\pm$0.10} & 0.669 {\tiny $\pm$0.18} & 0.632 {\tiny $\pm$0.28} & 0.572 {\tiny $\pm$0.26} & 0.711 {\tiny $\pm$0.28} & 0.632 {\tiny $\pm$0.09} \\
\end{longtable}
\endgroup
\twocolumn

\onecolumn
\begingroup
\scriptsize
\setlength{\tabcolsep}{3pt}
\setlength{\LTleft}{\fill}
\setlength{\LTright}{\fill}
\setlength{\LTcapwidth}{\textwidth}
\begin{longtable}{@{}l c c c c c c c@{}}
\caption{Dataset-averaged performance metrics for \texttt{Magistral-Small-2506}. Each metric is the \emph{unweighted} mean $\pm$ standard deviation of a method across all test datasets for this LLM. \textbf{Bold} entries mark the best method per metric across datasets.}\label{tab:Magistral_Small_2506_method_avg}\\
\toprule
\textbf{Method} & \textbf{ECE}$\downarrow$ & \textbf{Brier}$\downarrow$ & \textbf{Acc}$\uparrow$ & \textbf{F1}$\uparrow$ & \textbf{Spec}$\uparrow$ & \textbf{AUCPR}$\uparrow$ & \textbf{AUROC}$\uparrow$ \\
\midrule
\endfirsthead
\multicolumn{8}{l}{\scriptsize\emph{(Continued from previous page)}}\\
\toprule
\textbf{Method} & \textbf{ECE}$\downarrow$ & \textbf{Brier}$\downarrow$ & \textbf{Acc}$\uparrow$ & \textbf{F1}$\uparrow$ & \textbf{Spec}$\uparrow$ & \textbf{AUCPR}$\uparrow$ & \textbf{AUROC}$\uparrow$ \\
\midrule
\endhead
\midrule
\multicolumn{8}{r}{\scriptsize\emph{(Continued on next page)}}\\
\endfoot
\bottomrule
\endlastfoot
CE-DT & 0.237 {\tiny $\pm$0.15} & 0.245 {\tiny $\pm$0.14} & 0.642 {\tiny $\pm$0.21} & 0.392 {\tiny $\pm$0.24} & 0.609 {\tiny $\pm$0.29} & 0.353 {\tiny $\pm$0.23} & 0.604 {\tiny $\pm$0.07} \\
CE-KNN & 0.226 {\tiny $\pm$0.16} & 0.236 {\tiny $\pm$0.14} & 0.666 {\tiny $\pm$0.22} & 0.394 {\tiny $\pm$0.22} & 0.660 {\tiny $\pm$0.30} & 0.364 {\tiny $\pm$0.24} & 0.621 {\tiny $\pm$0.09} \\
CE-LogReg & 0.199 {\tiny $\pm$0.15} & 0.219 {\tiny $\pm$0.10} & 0.693 {\tiny $\pm$0.22} & 0.362 {\tiny $\pm$0.21} & 0.745 {\tiny $\pm$0.28} & 0.356 {\tiny $\pm$0.24} & 0.599 {\tiny $\pm$0.08} \\
CE-RF & 0.185 {\tiny $\pm$0.15} & 0.209 {\tiny $\pm$0.11} & 0.695 {\tiny $\pm$0.24} & 0.399 {\tiny $\pm$0.19} & 0.724 {\tiny $\pm$0.31} & 0.367 {\tiny $\pm$0.24} & 0.630 {\tiny $\pm$0.08} \\
CE-XGB & 0.228 {\tiny $\pm$0.16} & 0.234 {\tiny $\pm$0.14} & 0.664 {\tiny $\pm$0.22} & 0.399 {\tiny $\pm$0.22} & 0.653 {\tiny $\pm$0.30} & 0.369 {\tiny $\pm$0.24} & 0.628 {\tiny $\pm$0.10} \\
ETTIN & 0.152 {\tiny $\pm$0.23} & 0.204 {\tiny $\pm$0.20} & \textbf{0.716} {\tiny $\pm$0.26} & \textbf{0.540} {\tiny $\pm$0.20} & 0.647 {\tiny $\pm$0.38} & 0.526 {\tiny $\pm$0.25} & \textbf{0.737} {\tiny $\pm$0.12} \\
ETTIN-HGA & 0.160 {\tiny $\pm$0.20} & 0.201 {\tiny $\pm$0.17} & 0.710 {\tiny $\pm$0.25} & 0.493 {\tiny $\pm$0.21} & 0.669 {\tiny $\pm$0.36} & 0.496 {\tiny $\pm$0.24} & 0.724 {\tiny $\pm$0.12} \\
GNN-CD-ft-APPNP & 0.200 {\tiny $\pm$0.13} & 0.218 {\tiny $\pm$0.11} & 0.665 {\tiny $\pm$0.20} & 0.394 {\tiny $\pm$0.21} & 0.696 {\tiny $\pm$0.25} & 0.396 {\tiny $\pm$0.26} & 0.652 {\tiny $\pm$0.11} \\
\end{longtable}
\endgroup
\twocolumn

\onecolumn
\begingroup
\scriptsize
\setlength{\tabcolsep}{3pt}
\setlength{\LTleft}{\fill}
\setlength{\LTright}{\fill}
\setlength{\LTcapwidth}{\textwidth}
\begin{longtable}{@{}l c c c c c c c@{}}
\caption{Dataset-averaged performance metrics for \texttt{QwQ-32B}. Each metric is the \emph{unweighted} mean $\pm$ standard deviation of a method across all test datasets for this LLM. \textbf{Bold} entries mark the best method per metric across datasets.}\label{tab:QwQ_32B_method_avg}\\
\toprule
\textbf{Method} & \textbf{ECE}$\downarrow$ & \textbf{Brier}$\downarrow$ & \textbf{Acc}$\uparrow$ & \textbf{F1}$\uparrow$ & \textbf{Spec}$\uparrow$ & \textbf{AUCPR}$\uparrow$ & \textbf{AUROC}$\uparrow$ \\
\midrule
\endfirsthead
\multicolumn{8}{l}{\scriptsize\emph{(Continued from previous page)}}\\
\toprule
\textbf{Method} & \textbf{ECE}$\downarrow$ & \textbf{Brier}$\downarrow$ & \textbf{Acc}$\uparrow$ & \textbf{F1}$\uparrow$ & \textbf{Spec}$\uparrow$ & \textbf{AUCPR}$\uparrow$ & \textbf{AUROC}$\uparrow$ \\
\midrule
\endhead
\midrule
\multicolumn{8}{r}{\scriptsize\emph{(Continued on next page)}}\\
\endfoot
\bottomrule
\endlastfoot
CE-DT & 0.213 {\tiny $\pm$0.20} & 0.272 {\tiny $\pm$0.16} & 0.627 {\tiny $\pm$0.22} & 0.721 {\tiny $\pm$0.20} & 0.200 {\tiny $\pm$0.09} & 0.655 {\tiny $\pm$0.26} & 0.612 {\tiny $\pm$0.10} \\
CE-KNN & 0.198 {\tiny $\pm$0.20} & 0.254 {\tiny $\pm$0.16} & 0.625 {\tiny $\pm$0.23} & 0.728 {\tiny $\pm$0.20} & 0.145 {\tiny $\pm$0.09} & 0.695 {\tiny $\pm$0.26} & 0.651 {\tiny $\pm$0.12} \\
CE-LogReg & 0.242 {\tiny $\pm$0.18} & 0.268 {\tiny $\pm$0.14} & 0.615 {\tiny $\pm$0.24} & 0.737 {\tiny $\pm$0.20} & 0.000 & 0.685 {\tiny $\pm$0.27} & 0.649 {\tiny $\pm$0.12} \\
CE-RF & 0.187 {\tiny $\pm$0.20} & 0.249 {\tiny $\pm$0.15} & 0.626 {\tiny $\pm$0.23} & 0.729 {\tiny $\pm$0.20} & 0.135 {\tiny $\pm$0.09} & 0.685 {\tiny $\pm$0.27} & 0.652 {\tiny $\pm$0.13} \\
CE-XGB & 0.211 {\tiny $\pm$0.19} & 0.262 {\tiny $\pm$0.16} & 0.631 {\tiny $\pm$0.22} & 0.724 {\tiny $\pm$0.20} & 0.188 {\tiny $\pm$0.09} & 0.679 {\tiny $\pm$0.28} & 0.640 {\tiny $\pm$0.12} \\
ETTIN & 0.171 {\tiny $\pm$0.19} & 0.224 {\tiny $\pm$0.16} & 0.682 {\tiny $\pm$0.22} & 0.760 {\tiny $\pm$0.20} & 0.195 {\tiny $\pm$0.12} & 0.734 {\tiny $\pm$0.27} & 0.659 {\tiny $\pm$0.09} \\
ETTIN-HGA & 0.155 {\tiny $\pm$0.20} & 0.220 {\tiny $\pm$0.17} & \textbf{0.698} {\tiny $\pm$0.23} & \textbf{0.769} {\tiny $\pm$0.21} & 0.148 {\tiny $\pm$0.19} & 0.726 {\tiny $\pm$0.26} & 0.650 {\tiny $\pm$0.08} \\
GNN-CD-ft-APPNP & 0.332 {\tiny $\pm$0.11} & 0.331 {\tiny $\pm$0.13} & 0.609 {\tiny $\pm$0.12} & 0.645 {\tiny $\pm$0.19} & 0.549 {\tiny $\pm$0.11} & 0.706 {\tiny $\pm$0.27} & 0.618 {\tiny $\pm$0.09} \\
\end{longtable}
\endgroup
\twocolumn

\onecolumn
\begingroup
\scriptsize
\setlength{\tabcolsep}{3pt}
\setlength{\LTleft}{\fill}
\setlength{\LTright}{\fill}
\setlength{\LTcapwidth}{\textwidth}
\begin{longtable}{@{}l c c c c c c c@{}}
\caption{Dataset-averaged performance metrics for \texttt{EXAONE-Deep-32B}. Each metric is the \emph{unweighted} mean $\pm$ standard deviation of a method across all test datasets for this LLM. \textbf{Bold} entries mark the best method per metric across datasets.}\label{tab:EXAONE_Deep_32B_method_avg}\\
\toprule
\textbf{Method} & \textbf{ECE}$\downarrow$ & \textbf{Brier}$\downarrow$ & \textbf{Acc}$\uparrow$ & \textbf{F1}$\uparrow$ & \textbf{Spec}$\uparrow$ & \textbf{AUCPR}$\uparrow$ & \textbf{AUROC}$\uparrow$ \\
\midrule
\endfirsthead
\multicolumn{8}{l}{\scriptsize\emph{(Continued from previous page)}}\\
\toprule
\textbf{Method} & \textbf{ECE}$\downarrow$ & \textbf{Brier}$\downarrow$ & \textbf{Acc}$\uparrow$ & \textbf{F1}$\uparrow$ & \textbf{Spec}$\uparrow$ & \textbf{AUCPR}$\uparrow$ & \textbf{AUROC}$\uparrow$ \\
\midrule
\endhead
\midrule
\multicolumn{8}{r}{\scriptsize\emph{(Continued on next page)}}\\
\endfoot
\bottomrule
\endlastfoot
CE-DT & 0.217 {\tiny $\pm$0.17} & 0.304 {\tiny $\pm$0.13} & 0.632 {\tiny $\pm$0.18} & 0.713 {\tiny $\pm$0.18} & 0.323 {\tiny $\pm$0.19} & 0.679 {\tiny $\pm$0.20} & 0.595 {\tiny $\pm$0.09} \\
CE-KNN & 0.176 {\tiny $\pm$0.19} & 0.245 {\tiny $\pm$0.14} & 0.651 {\tiny $\pm$0.22} & 0.733 {\tiny $\pm$0.20} & 0.306 {\tiny $\pm$0.26} & 0.686 {\tiny $\pm$0.25} & 0.671 {\tiny $\pm$0.15} \\
CE-LogReg & 0.165 {\tiny $\pm$0.19} & 0.261 {\tiny $\pm$0.10} & 0.588 {\tiny $\pm$0.21} & 0.720 {\tiny $\pm$0.18} & 0.000 & 0.676 {\tiny $\pm$0.26} & 0.669 {\tiny $\pm$0.14} \\
CE-RF & 0.167 {\tiny $\pm$0.19} & 0.239 {\tiny $\pm$0.14} & 0.655 {\tiny $\pm$0.21} & 0.730 {\tiny $\pm$0.20} & 0.340 {\tiny $\pm$0.26} & 0.680 {\tiny $\pm$0.26} & 0.670 {\tiny $\pm$0.15} \\
CE-XGB & 0.165 {\tiny $\pm$0.19} & 0.240 {\tiny $\pm$0.14} & 0.653 {\tiny $\pm$0.21} & 0.730 {\tiny $\pm$0.20} & 0.312 {\tiny $\pm$0.25} & 0.681 {\tiny $\pm$0.26} & 0.669 {\tiny $\pm$0.15} \\
ETTIN & 0.146 {\tiny $\pm$0.20} & 0.215 {\tiny $\pm$0.15} & 0.680 {\tiny $\pm$0.22} & \textbf{0.770} {\tiny $\pm$0.20} & 0.159 {\tiny $\pm$0.11} & 0.752 {\tiny $\pm$0.26} & \textbf{0.706} {\tiny $\pm$0.11} \\
ETTIN-HGA & 0.143 {\tiny $\pm$0.19} & 0.218 {\tiny $\pm$0.15} & 0.684 {\tiny $\pm$0.21} & 0.765 {\tiny $\pm$0.20} & 0.171 {\tiny $\pm$0.14} & 0.745 {\tiny $\pm$0.26} & 0.697 {\tiny $\pm$0.10} \\
GNN-CD-ft-APPNP & 0.256 {\tiny $\pm$0.12} & 0.273 {\tiny $\pm$0.08} & 0.647 {\tiny $\pm$0.18} & 0.703 {\tiny $\pm$0.22} & 0.406 {\tiny $\pm$0.31} & 0.697 {\tiny $\pm$0.25} & 0.615 {\tiny $\pm$0.11} \\
\end{longtable}
\endgroup
\twocolumn

\onecolumn
\begingroup
\scriptsize
\setlength{\tabcolsep}{3pt}
\setlength{\LTleft}{\fill}
\setlength{\LTright}{\fill}
\setlength{\LTcapwidth}{\textwidth}
\begin{longtable}{@{}l l c c c c c c c@{}}
\caption{Performance for each dataset--method pair averaged across all LLMs. Metrics are \emph{unweighted} means $\pm$ standard deviations across models, showing which methods generalize best on each dataset independent of the underlying LLM. \textbf{Bold} entries mark the best method per metric within each dataset.}\label{tab:dataset_method_avg}\\
\toprule
\textbf{Dataset} & \textbf{Method} & \textbf{ECE}$\downarrow$ & \textbf{Brier}$\downarrow$ & \textbf{Acc}$\uparrow$ & \textbf{F1}$\uparrow$ & \textbf{Spec}$\uparrow$ & \textbf{AUCPR}$\uparrow$ & \textbf{AUROC}$\uparrow$ \\
\midrule
\endfirsthead
\multicolumn{9}{l}{\scriptsize\emph{(Continued from previous page)}}\\
\toprule
\textbf{Dataset} & \textbf{Method} & \textbf{ECE}$\downarrow$ & \textbf{Brier}$\downarrow$ & \textbf{Acc}$\uparrow$ & \textbf{F1}$\uparrow$ & \textbf{Spec}$\uparrow$ & \textbf{AUCPR}$\uparrow$ & \textbf{AUROC}$\uparrow$ \\
\midrule
\endhead
\midrule
\multicolumn{9}{r}{\scriptsize\emph{(Continued on next page)}}\\
\endfoot
\bottomrule
\endlastfoot
BBH & CE-DT & 0.103 {\tiny $\pm$0.02} & 0.202 {\tiny $\pm$0.03} & 0.729 {\tiny $\pm$0.06} & 0.795 {\tiny $\pm$0.13} & 0.253 {\tiny $\pm$0.21} & 0.792 {\tiny $\pm$0.15} & 0.651 {\tiny $\pm$0.04} \\
 & CE-KNN & 0.075 {\tiny $\pm$0.02} & 0.181 {\tiny $\pm$0.03} & 0.746 {\tiny $\pm$0.05} & 0.803 {\tiny $\pm$0.14} & 0.236 {\tiny $\pm$0.27} & 0.803 {\tiny $\pm$0.14} & 0.682 {\tiny $\pm$0.01} \\
 & CE-LogReg & 0.070 {\tiny $\pm$0.03} & 0.183 {\tiny $\pm$0.02} & 0.749 {\tiny $\pm$0.04} & 0.800 {\tiny $\pm$0.16} & 0.144 {\tiny $\pm$0.34} & 0.809 {\tiny $\pm$0.14} & 0.696 {\tiny $\pm$0.02} \\
 & CE-RF & 0.049 {\tiny $\pm$0.02} & 0.173 {\tiny $\pm$0.02} & 0.756 {\tiny $\pm$0.04} & 0.804 {\tiny $\pm$0.14} & 0.243 {\tiny $\pm$0.31} & 0.811 {\tiny $\pm$0.14} & 0.696 {\tiny $\pm$0.02} \\
 & CE-XGB & 0.071 {\tiny $\pm$0.02} & 0.177 {\tiny $\pm$0.02} & 0.748 {\tiny $\pm$0.05} & 0.806 {\tiny $\pm$0.14} & 0.245 {\tiny $\pm$0.25} & 0.809 {\tiny $\pm$0.14} & 0.694 {\tiny $\pm$0.01} \\
 & ETTIN & 0.061 {\tiny $\pm$0.05} & 0.143 {\tiny $\pm$0.03} & 0.803 {\tiny $\pm$0.05} & \textbf{0.853} {\tiny $\pm$0.09} & 0.305 {\tiny $\pm$0.31} & \textbf{0.885} {\tiny $\pm$0.07} & \textbf{0.742} {\tiny $\pm$0.04} \\
 & ETTIN-HGA & 0.046 {\tiny $\pm$0.01} & \textbf{0.142} {\tiny $\pm$0.02} & \textbf{0.805} {\tiny $\pm$0.04} & 0.853 {\tiny $\pm$0.11} & 0.232 {\tiny $\pm$0.32} & 0.872 {\tiny $\pm$0.09} & 0.723 {\tiny $\pm$0.05} \\
 & GNN-CD-ft-APPNP & 0.185 {\tiny $\pm$0.11} & 0.227 {\tiny $\pm$0.08} & 0.709 {\tiny $\pm$0.07} & 0.771 {\tiny $\pm$0.12} & 0.503 {\tiny $\pm$0.20} & 0.829 {\tiny $\pm$0.13} & 0.662 {\tiny $\pm$0.05} \\
\end{longtable}
\endgroup
\twocolumn

\onecolumn
\begingroup
\scriptsize
\setlength{\tabcolsep}{3pt}
\setlength{\LTleft}{\fill}
\setlength{\LTright}{\fill}
\setlength{\LTcapwidth}{\textwidth}
\begin{longtable}{@{}l c c c c c c c@{}}
\caption{Performance for each LLM--method pair averaged across datasets. Metrics are \emph{unweighted} means $\pm$ standard deviations across datasets, emphasizing which methods work best for a given LLM after averaging out dataset effects. \textbf{Bold} entries mark the best method per metric within each LLM.}\label{tab:llm_method_avg}\\
\toprule
\textbf{Method} & \textbf{ECE}$\downarrow$ & \textbf{Brier}$\downarrow$ & \textbf{Acc}$\uparrow$ & \textbf{F1}$\uparrow$ & \textbf{Spec}$\uparrow$ & \textbf{AUCPR}$\uparrow$ & \textbf{AUROC}$\uparrow$ \\
\midrule
\endfirsthead
\multicolumn{8}{l}{\scriptsize\emph{(Continued from previous page)}}\\
\toprule
\textbf{Method} & \textbf{ECE}$\downarrow$ & \textbf{Brier}$\downarrow$ & \textbf{Acc}$\uparrow$ & \textbf{F1}$\uparrow$ & \textbf{Spec}$\uparrow$ & \textbf{AUCPR}$\uparrow$ & \textbf{AUROC}$\uparrow$ \\
\midrule
\endhead
\midrule
\multicolumn{8}{r}{\scriptsize\emph{(Continued on next page)}}\\
\endfoot
\bottomrule
\endlastfoot
\multicolumn{8}{c}{\footnotesize\textbf{\texttt{Phi-4-mini-flash-reasoning}}} \\
\addlinespace[2pt]
CE-DT & 0.181 {\tiny $\pm$0.11} & 0.252 {\tiny $\pm$0.08} & 0.588 {\tiny $\pm$0.18} & 0.607 {\tiny $\pm$0.23} & 0.355 {\tiny $\pm$0.11} & 0.600 {\tiny $\pm$0.28} & 0.602 {\tiny $\pm$0.07} \\
CE-KNN & 0.176 {\tiny $\pm$0.10} & 0.239 {\tiny $\pm$0.08} & 0.626 {\tiny $\pm$0.15} & 0.591 {\tiny $\pm$0.25} & 0.469 {\tiny $\pm$0.13} & 0.586 {\tiny $\pm$0.30} & 0.618 {\tiny $\pm$0.09} \\
CE-LogReg & 0.224 {\tiny $\pm$0.13} & 0.249 {\tiny $\pm$0.07} & 0.513 {\tiny $\pm$0.26} & 0.644 {\tiny $\pm$0.23} & 0.002 {\tiny $\pm$0.00} & 0.601 {\tiny $\pm$0.30} & 0.630 {\tiny $\pm$0.11} \\
CE-RF & 0.151 {\tiny $\pm$0.13} & 0.222 {\tiny $\pm$0.08} & 0.613 {\tiny $\pm$0.18} & 0.606 {\tiny $\pm$0.25} & 0.388 {\tiny $\pm$0.16} & 0.600 {\tiny $\pm$0.30} & 0.634 {\tiny $\pm$0.10} \\
CE-XGB & 0.153 {\tiny $\pm$0.13} & 0.223 {\tiny $\pm$0.08} & 0.616 {\tiny $\pm$0.17} & 0.605 {\tiny $\pm$0.25} & 0.404 {\tiny $\pm$0.16} & 0.601 {\tiny $\pm$0.30} & 0.632 {\tiny $\pm$0.10} \\
ETTIN & 0.149 {\tiny $\pm$0.09} & 0.207 {\tiny $\pm$0.08} & 0.648 {\tiny $\pm$0.17} & 0.606 {\tiny $\pm$0.26} & 0.485 {\tiny $\pm$0.20} & 0.625 {\tiny $\pm$0.30} & 0.634 {\tiny $\pm$0.11} \\
ETTIN-HGA & \textbf{0.106} {\tiny $\pm$0.09} & \textbf{0.194} {\tiny $\pm$0.08} & \textbf{0.685} {\tiny $\pm$0.16} & 0.613 {\tiny $\pm$0.28} & 0.469 {\tiny $\pm$0.28} & 0.630 {\tiny $\pm$0.31} & 0.641 {\tiny $\pm$0.12} \\
\end{longtable}
\endgroup
\twocolumn

\onecolumn
\begingroup
\scriptsize
\setlength{\tabcolsep}{3pt}
\setlength{\LTleft}{\fill}
\setlength{\LTright}{\fill}
\setlength{\LTcapwidth}{\textwidth}
\begin{longtable}{@{}l c c c c c c c@{}}
\caption{Overall method performance aggregated across LLMs and datasets using a two-stage \emph{unweighted} average: first average a method across datasets within each LLM, then average those LLM-level means across LLMs. We report mean $\pm$ standard deviation; \textbf{bold} marks the best method per metric.}
\label{tab:method_only_appendix}\\
\toprule
\textbf{Method} & \textbf{ECE}$\downarrow$ & \textbf{Brier}$\downarrow$ & \textbf{Acc}$\uparrow$ & \textbf{F1}$\uparrow$ & \textbf{Spec}$\uparrow$ & \textbf{AUCPR}$\uparrow$ & \textbf{AUROC}$\uparrow$ \\
\midrule
\endfirsthead
\multicolumn{8}{l}{\scriptsize\emph{(Continued from previous page)}}\\
\toprule
\textbf{Method} & \textbf{ECE}$\downarrow$ & \textbf{Brier}$\downarrow$ & \textbf{Acc}$\uparrow$ & \textbf{F1}$\uparrow$ & \textbf{Spec}$\uparrow$ & \textbf{AUCPR}$\uparrow$ & \textbf{AUROC}$\uparrow$ \\
\midrule
\endhead
\midrule
\multicolumn{8}{r}{\scriptsize\emph{(Continued on next page)}}\\
\endfoot
\bottomrule
\endlastfoot
CE-DT & 0.209 {\tiny $\pm$0.02} & 0.266 {\tiny $\pm$0.02} & 0.626 {\tiny $\pm$0.02} & 0.647 {\tiny $\pm$0.13} & 0.316 {\tiny $\pm$0.16} & 0.608 {\tiny $\pm$0.13} & 0.611 {\tiny $\pm$0.01} \\
CE-KNN & 0.193 {\tiny $\pm$0.02} & 0.246 {\tiny $\pm$0.01} & 0.633 {\tiny $\pm$0.02} & 0.651 {\tiny $\pm$0.14} & 0.283 {\tiny $\pm$0.24} & 0.620 {\tiny $\pm$0.13} & 0.648 {\tiny $\pm$0.02} \\
CE-LogReg & 0.209 {\tiny $\pm$0.03} & 0.253 {\tiny $\pm$0.02} & 0.607 {\tiny $\pm$0.06} & 0.654 {\tiny $\pm$0.15} & 0.144 {\tiny $\pm$0.30} & 0.619 {\tiny $\pm$0.13} & 0.651 {\tiny $\pm$0.03} \\
CE-RF & 0.178 {\tiny $\pm$0.02} & 0.236 {\tiny $\pm$0.02} & 0.636 {\tiny $\pm$0.03} & 0.654 {\tiny $\pm$0.13} & 0.277 {\tiny $\pm$0.27} & 0.621 {\tiny $\pm$0.13} & 0.656 {\tiny $\pm$0.02} \\
CE-XGB & 0.190 {\tiny $\pm$0.03} & 0.244 {\tiny $\pm$0.01} & 0.635 {\tiny $\pm$0.02} & 0.652 {\tiny $\pm$0.13} & 0.293 {\tiny $\pm$0.21} & 0.617 {\tiny $\pm$0.13} & 0.648 {\tiny $\pm$0.02} \\
ETTIN & 0.160 {\tiny $\pm$0.01} & 0.217 {\tiny $\pm$0.01} & 0.677 {\tiny $\pm$0.02} & \textbf{0.698} {\tiny $\pm$0.10} & 0.290 {\tiny $\pm$0.22} & 0.680 {\tiny $\pm$0.09} & \textbf{0.672} {\tiny $\pm$0.04} \\
ETTIN-HGA & \textbf{0.148} {\tiny $\pm$0.02} & 0.211 {\tiny $\pm$0.01} & \textbf{0.689} {\tiny $\pm$0.01} & 0.693 {\tiny $\pm$0.12} & 0.282 {\tiny $\pm$0.23} & 0.678 {\tiny $\pm$0.10} & 0.670 {\tiny $\pm$0.03} \\
GNN-CD-ft-APPNP & 0.254 {\tiny $\pm$0.09} & 0.274 {\tiny $\pm$0.07} & 0.640 {\tiny $\pm$0.03} & 0.588 {\tiny $\pm$0.11} & 0.564 {\tiny $\pm$0.11} & 0.642 {\tiny $\pm$0.13} & 0.620 {\tiny $\pm$0.02} \\
\end{longtable}
\endgroup
\twocolumn

%% file: sections/appendix_ellipse_plots.tex
\section{Supplementary Performance Visualization with Ellipse Plots}
\label{app:allipse_plots}

\begin{figure*}[t]
\centering
\makebox[\textwidth][c]{\includegraphics[width=0.9\textwidth]{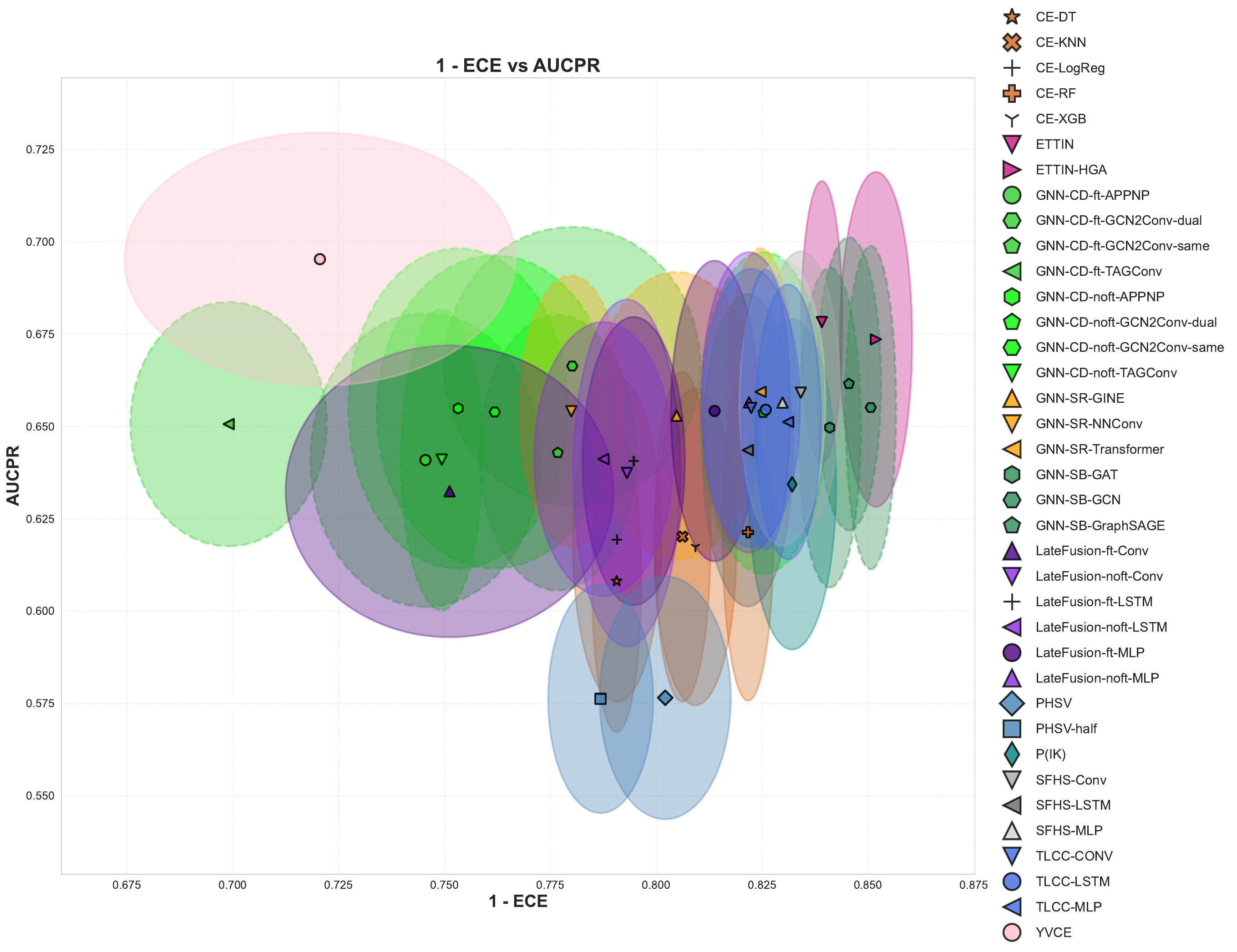}}
\caption{Performance trade-off between calibration (1-ECE) and discrimination focused on the positive class (AUCPR). This plot confirms that methods with the best calibration are not necessarily the best at identifying correct answers.}
\label{fig:ECE_vs_AUCPR_ellipse_distribution}
\end{figure*}

\begin{figure*}[t]
\centering
\makebox[\textwidth][c]{\includegraphics[width=0.9\textwidth]{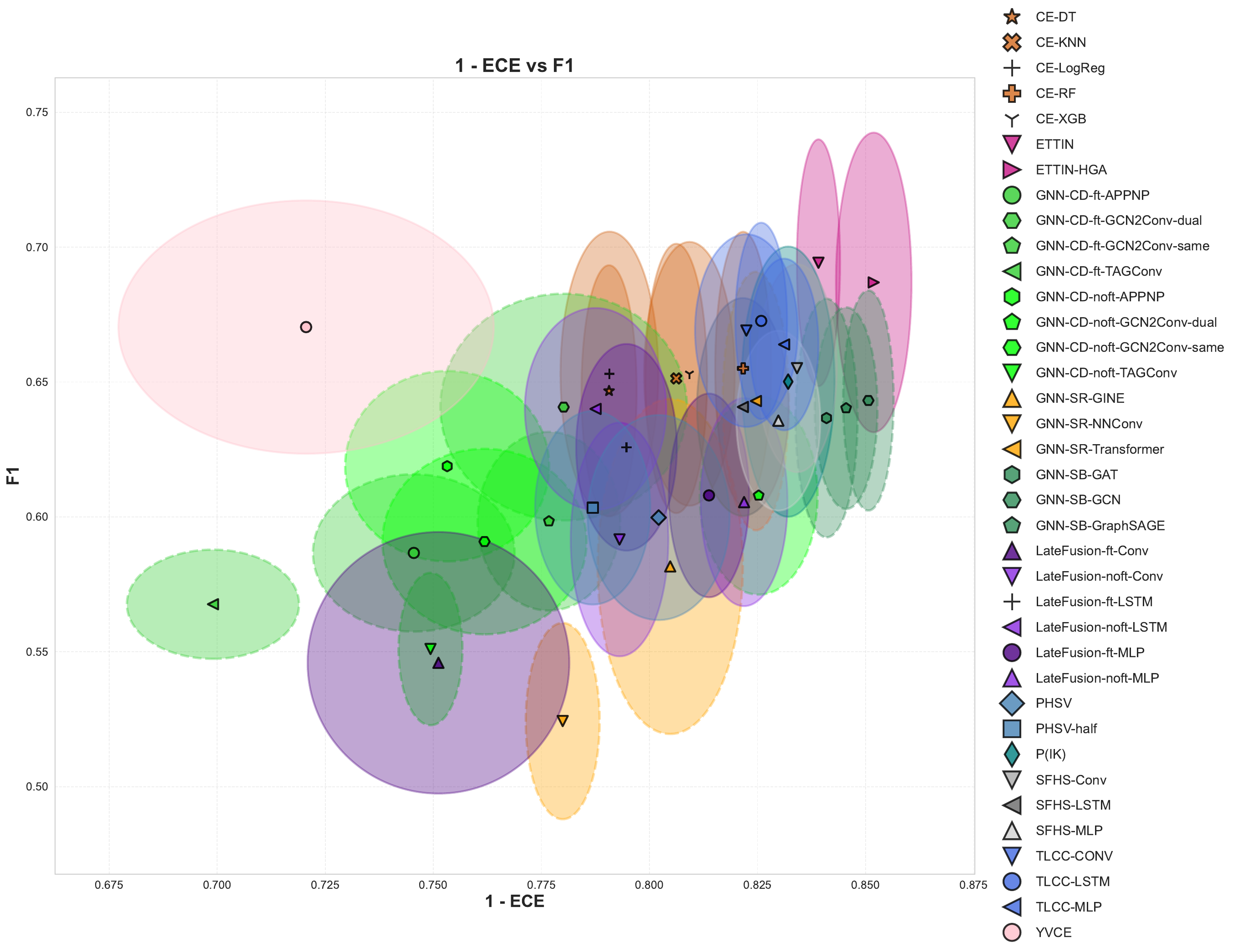}}
\caption{Performance trade-off between calibration (1-ECE) and F1 Score. This view highlights the relationship between probabilistic accuracy and the harmonic mean of precision and recall.}
\label{fig:ECE_vs_F1_ellipse_distribution}
\end{figure*}

\begin{figure*}[t]
\centering
\makebox[\textwidth][c]{\includegraphics[width=0.9\textwidth]{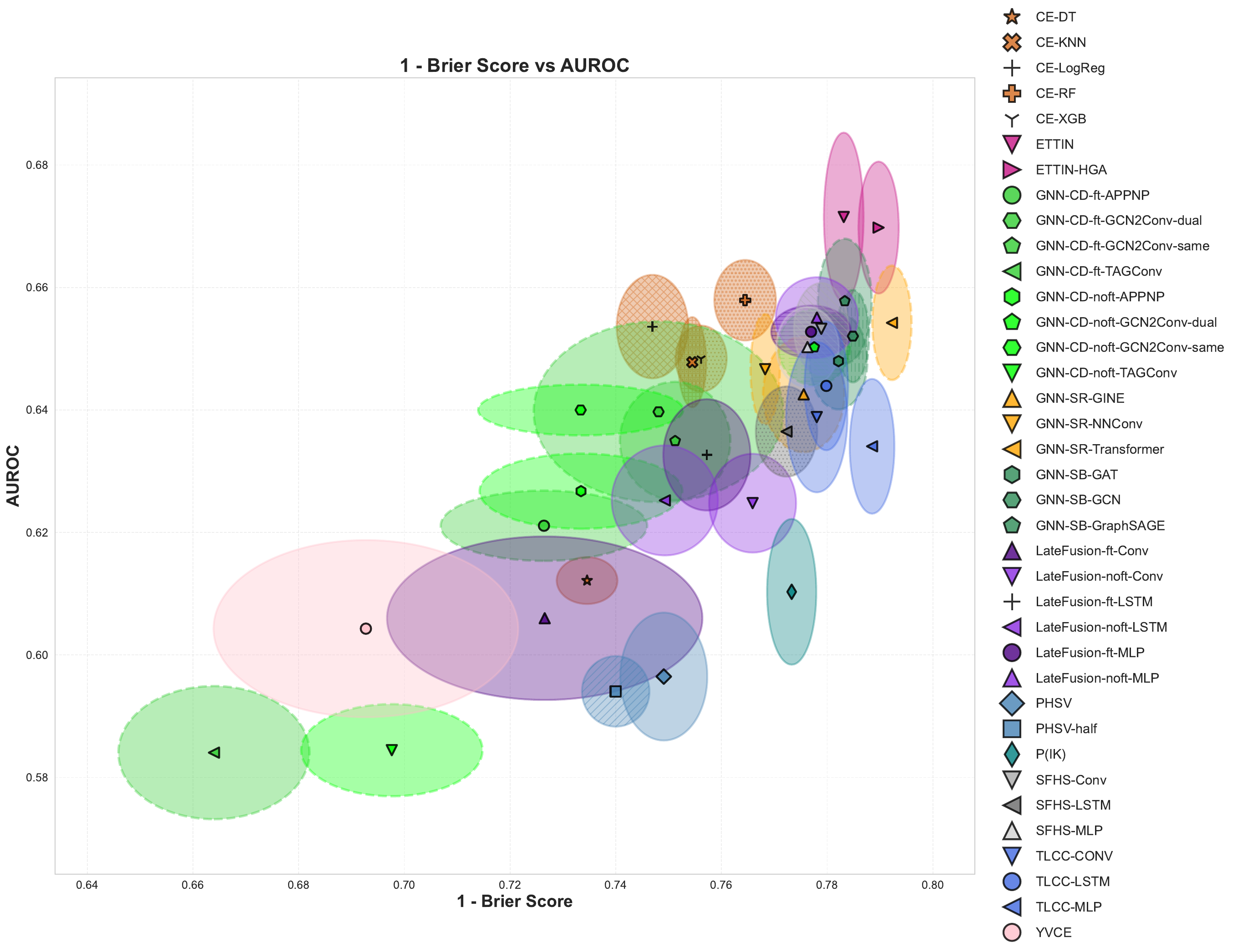}}
\caption{Performance trade-off between the Brier Score (plotted as 1-Brier), which combines calibration and discrimination, and pure discrimination (AUROC).}
\label{fig:Brier_Score_vs_AUROC_ellipse_distribution}
\end{figure*}

\begin{figure*}[t]
\centering
\makebox[\textwidth][c]{\includegraphics[width=0.9\textwidth]{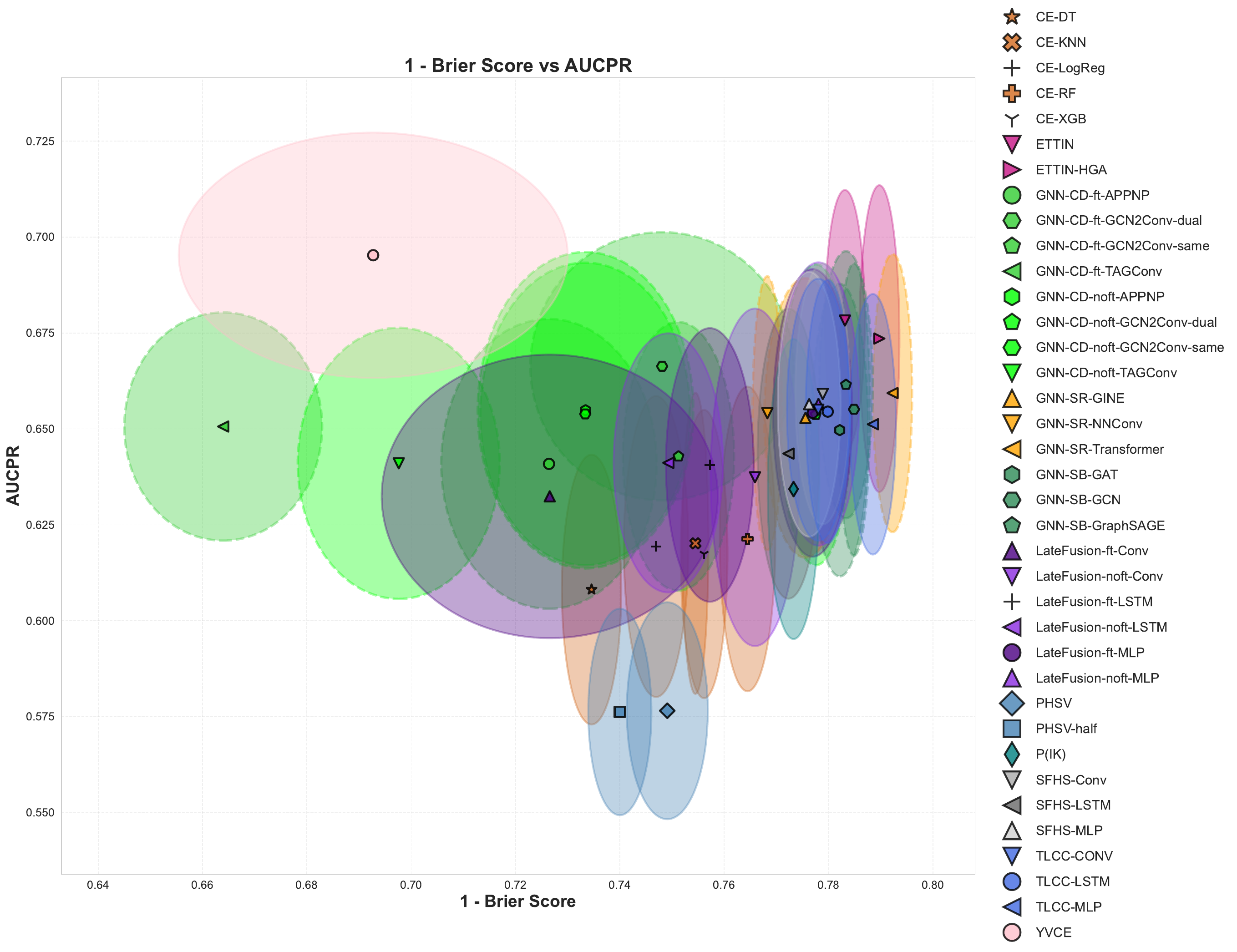}}
\caption{Performance trade-off between the Brier Score (1-Brier) and discrimination focused on the positive class (AUCPR).}
\label{fig:Brier_Score_vs_AUCPR_ellipse_distribution}
\end{figure*}

\begin{figure*}[t]
\centering
\makebox[\textwidth][c]{\includegraphics[width=0.9\textwidth]{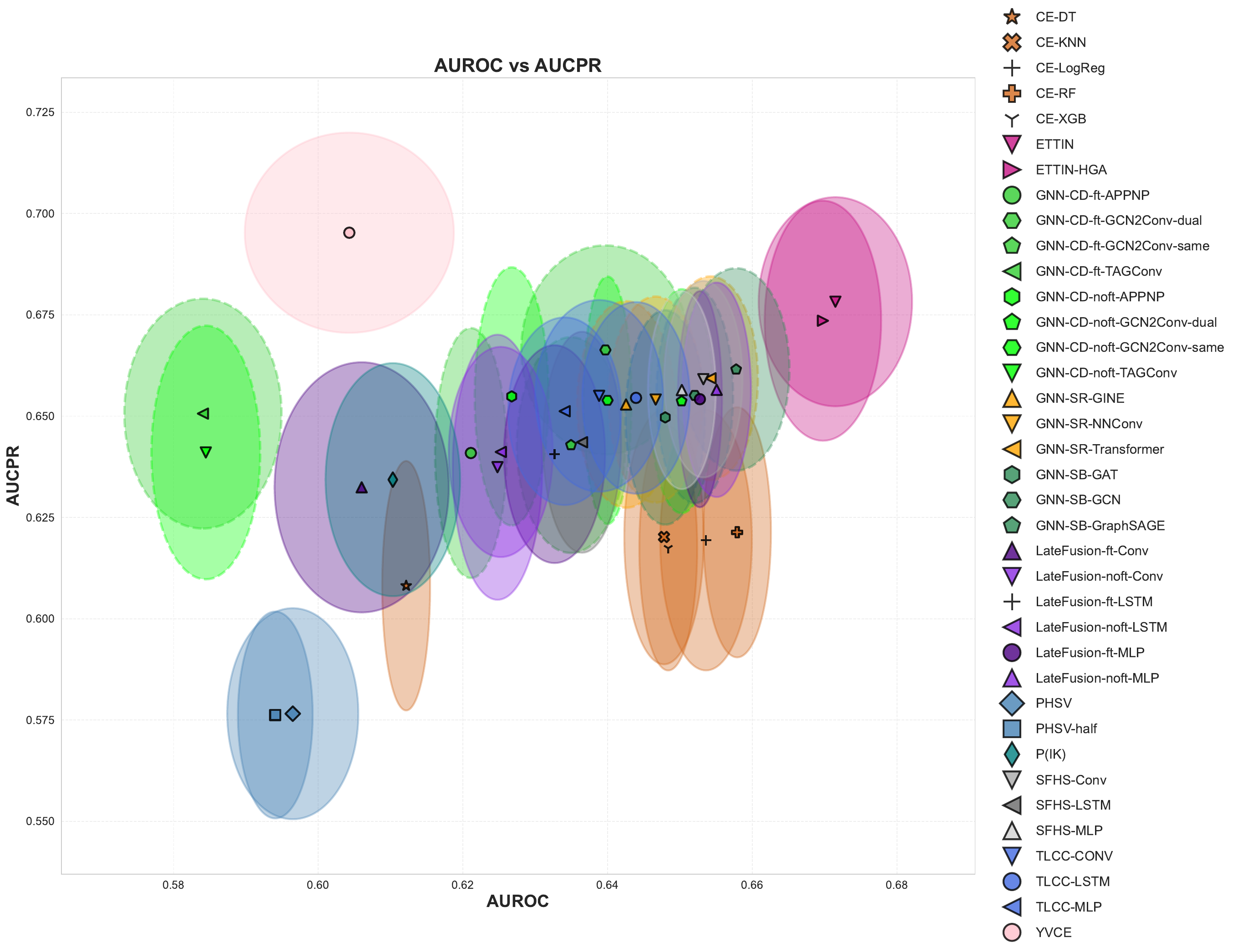}}
\caption{A comparison of the two primary discrimination metrics, AUROC and AUCPR. The strong positive correlation indicates that most methods that are good at general ranking are also good at ranking the positive class specifically.}
\label{fig:AUROC_vs_AUCPR_ellipse_distribution}
\end{figure*}

\begin{figure*}[t]
\centering
\makebox[\textwidth][c]{\includegraphics[width=0.9\textwidth]{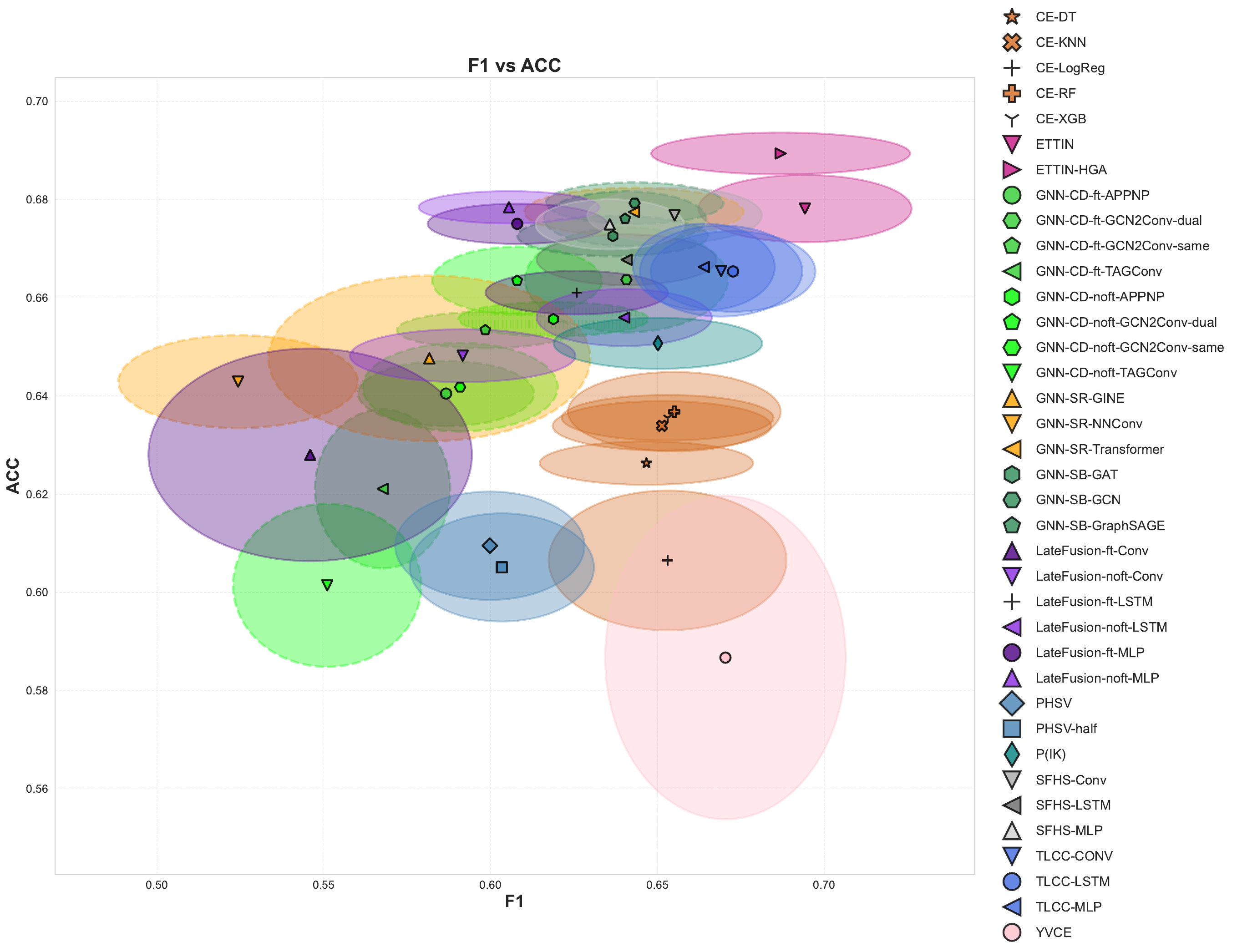}}
\caption{A comparison of two threshold-dependent classification metrics, F1 Score and Accuracy. This plot shows the relationship between balanced performance (F1) and overall correctness (Accuracy) at each method's optimal threshold.}
\label{fig:F1_vs_ACC_ellipse_distribution}
\end{figure*}

This section provides supplementary ellipse plots to complement the main 1-ECE vs. AUROC analysis presented in Figure~\ref{fig:ECE_vs_AUROC_ellipse_distribution}. Together, Figures~\ref{fig:ECE_vs_AUCPR_ellipse_distribution}, \ref{fig:ECE_vs_F1_ellipse_distribution}, \ref{fig:Brier_Score_vs_AUROC_ellipse_distribution}, \ref{fig:Brier_Score_vs_AUCPR_ellipse_distribution}, \ref{fig:AUROC_vs_AUCPR_ellipse_distribution}, and \ref{fig:F1_vs_ACC_ellipse_distribution} visualize performance trade-offs across a broad set of calibration, discrimination, and threshold-dependent evaluation metrics.

The data for each ellipse is aggregated in a two-stage process to ensure a balanced comparison. First, for each confidence estimation method, its performance on a given metric is computed separately for each of the six LRMs by taking an unweighted average across all test datasets. This yields an LRM-specific mean score. The center of each ellipse shown in Figures~\ref{fig:ECE_vs_AUCPR_ellipse_distribution} through~\ref{fig:F1_vs_ACC_ellipse_distribution} represents the final mean performance, obtained by averaging these six LRM-specific means. The width and height of each ellipse correspond to the standard deviation of the LRM-specific means, illustrating the consistency of each method across different model architectures. For metrics where lower values indicate better performance, including ECE and Brier Score, we plot their inverse quantities such as 1-ECE and 1-Brier so that the optimal region is consistently located in the top-right corner of each plot.

%% file: sections/appendix_calibration_diagrams.tex
\section{Detailed Calibration Analysis by Dataset}
\label{app:appendix_calibration_diagrams}
\begin{figure*}[t]
\centering
\makebox[\textwidth][c]{\includegraphics[width=0.9\textwidth]{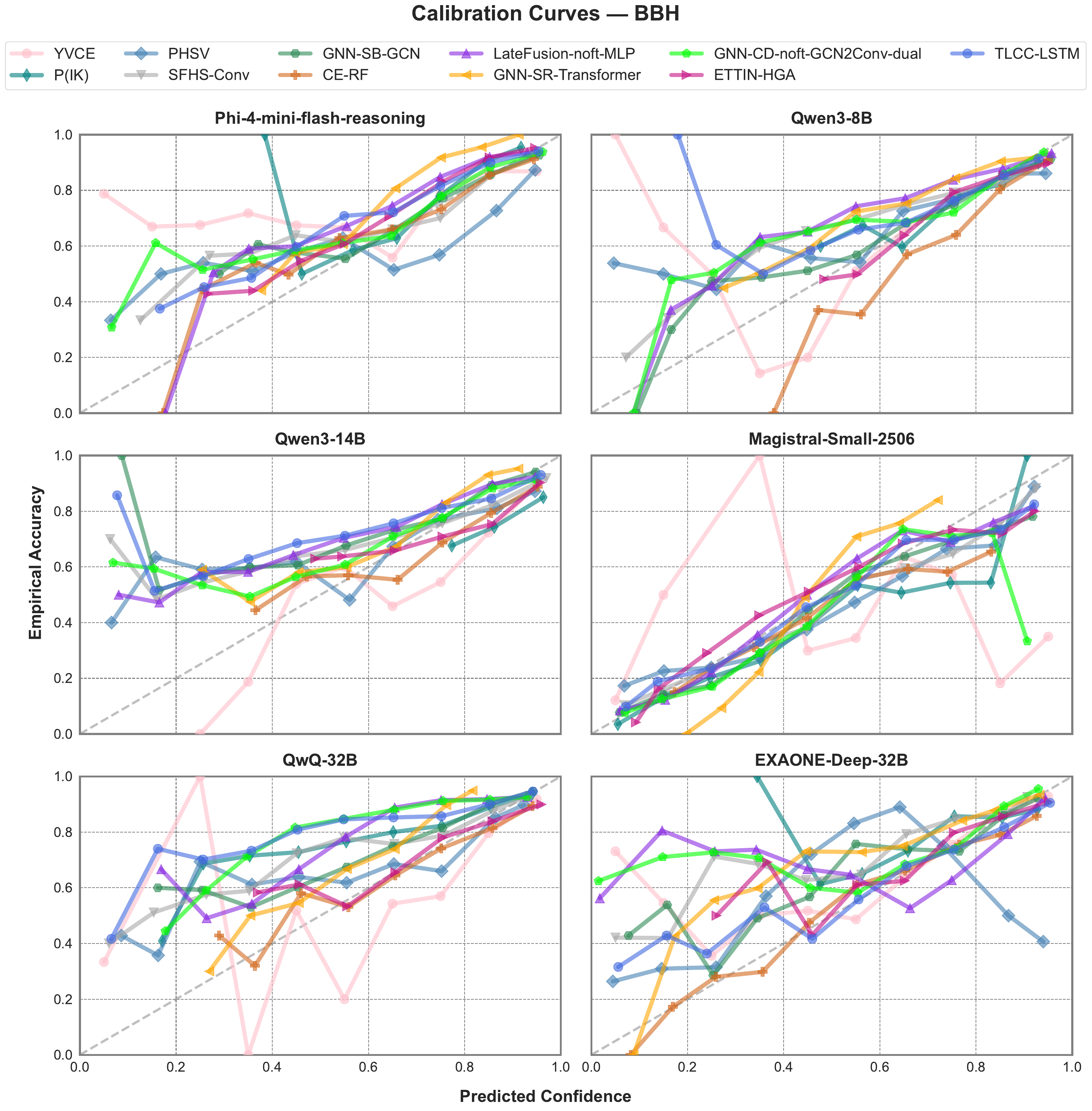}}
\caption{Reliability diagrams for the top-performing method variants on the BBH dataset, aggregated across all LRMs.}
\label{fig:calibration_grid_v3_bbh_best_10bins}
\end{figure*}

\begin{figure*}[t]
\centering
\makebox[\textwidth][c]{\includegraphics[width=0.9\textwidth]{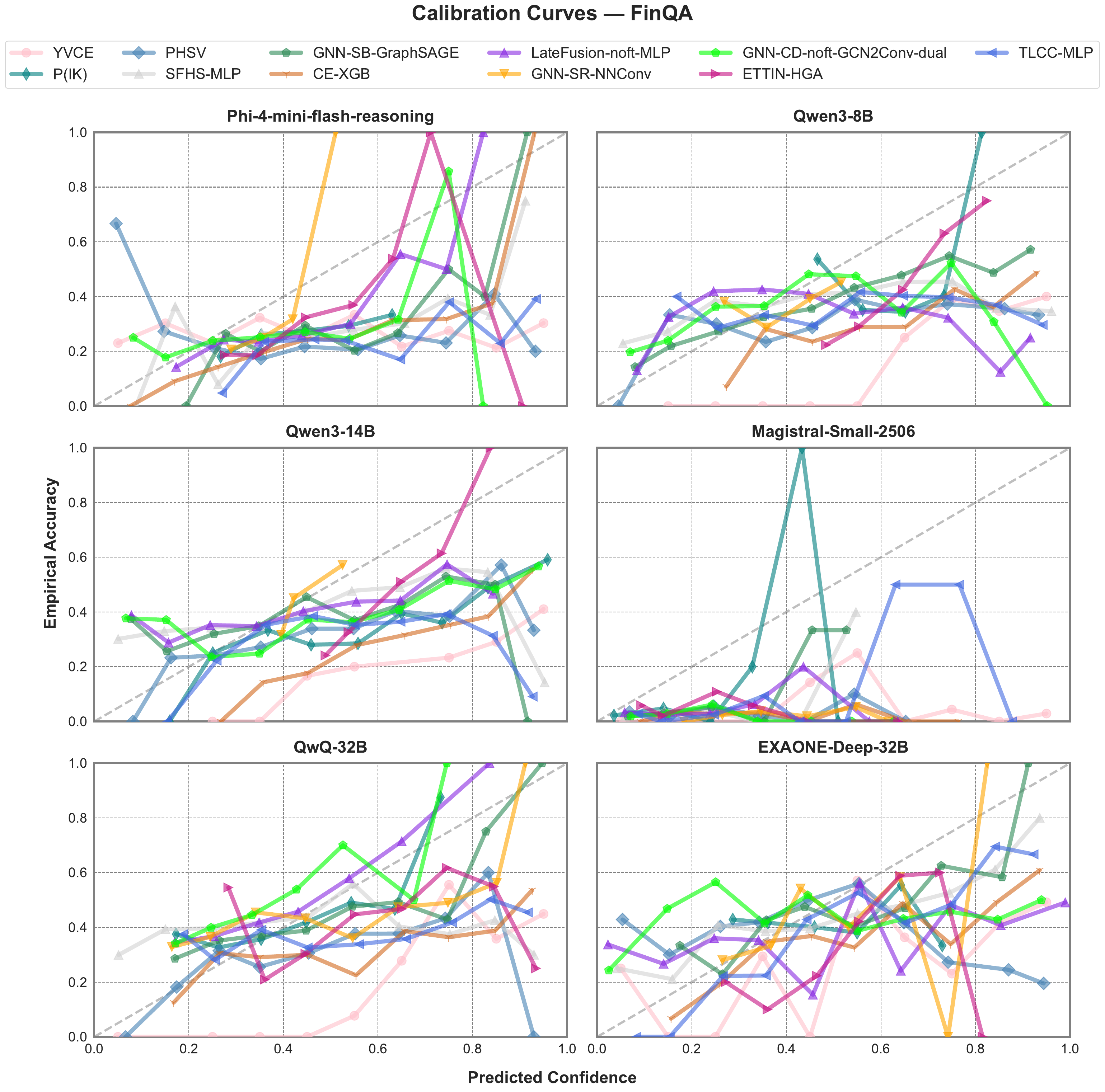}}
\caption{Reliability diagrams for the top-performing method variants on the FinQA dataset, aggregated across all LRMs.}
\label{fig:calibration_grid_v3_finqa_best_10bins}
\end{figure*}

\begin{figure*}[t]
\centering
\makebox[\textwidth][c]{\includegraphics[width=0.9\textwidth]{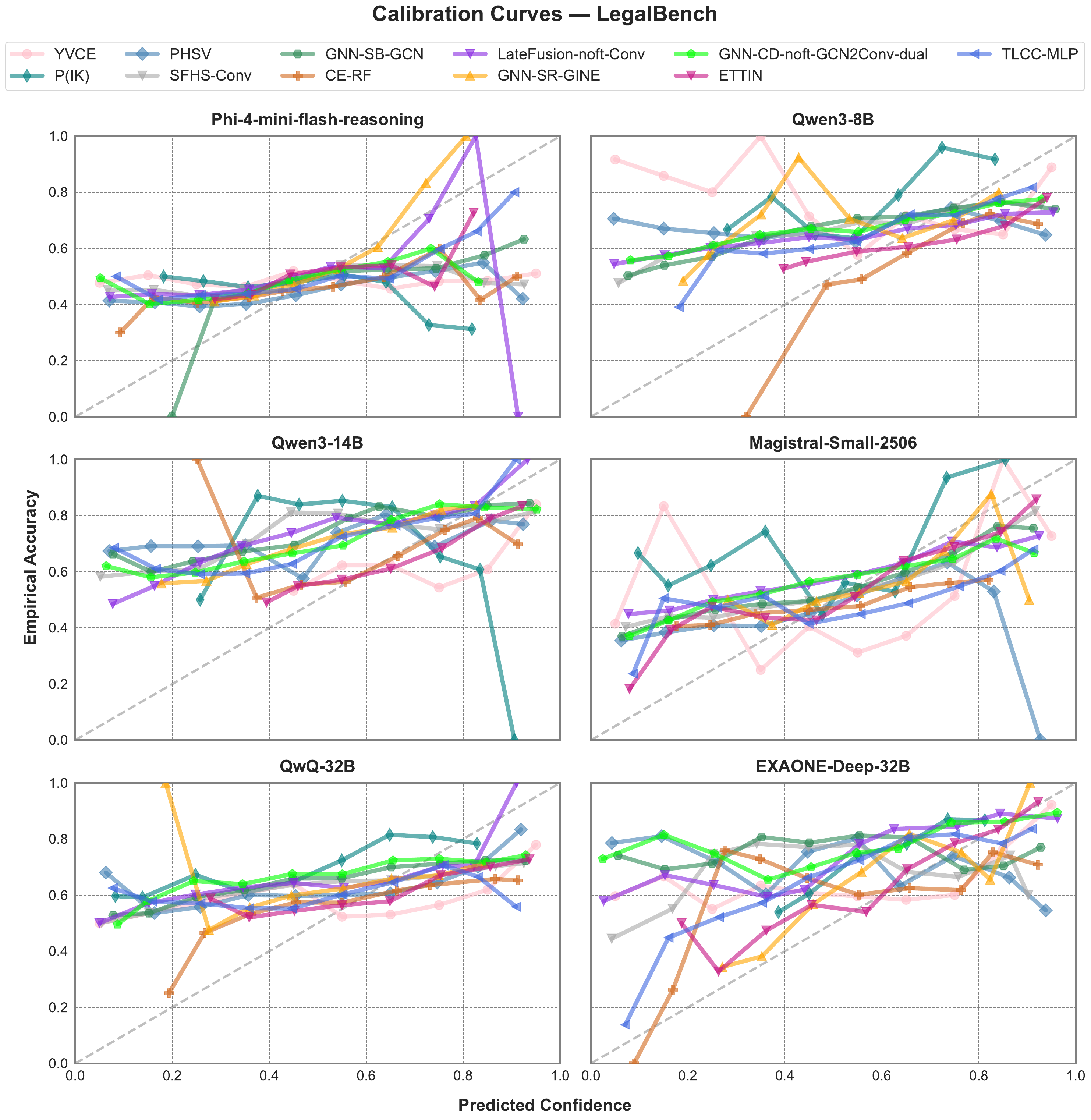}}
\caption{Reliability diagrams for the top-performing method variants on the LegalBench dataset, aggregated across all LRMs.}
\label{fig:calibration_grid_v3_legalbench_best_10bins}
\end{figure*}

\begin{figure*}[t]
\centering
\makebox[\textwidth][c]{\includegraphics[width=0.9\textwidth]{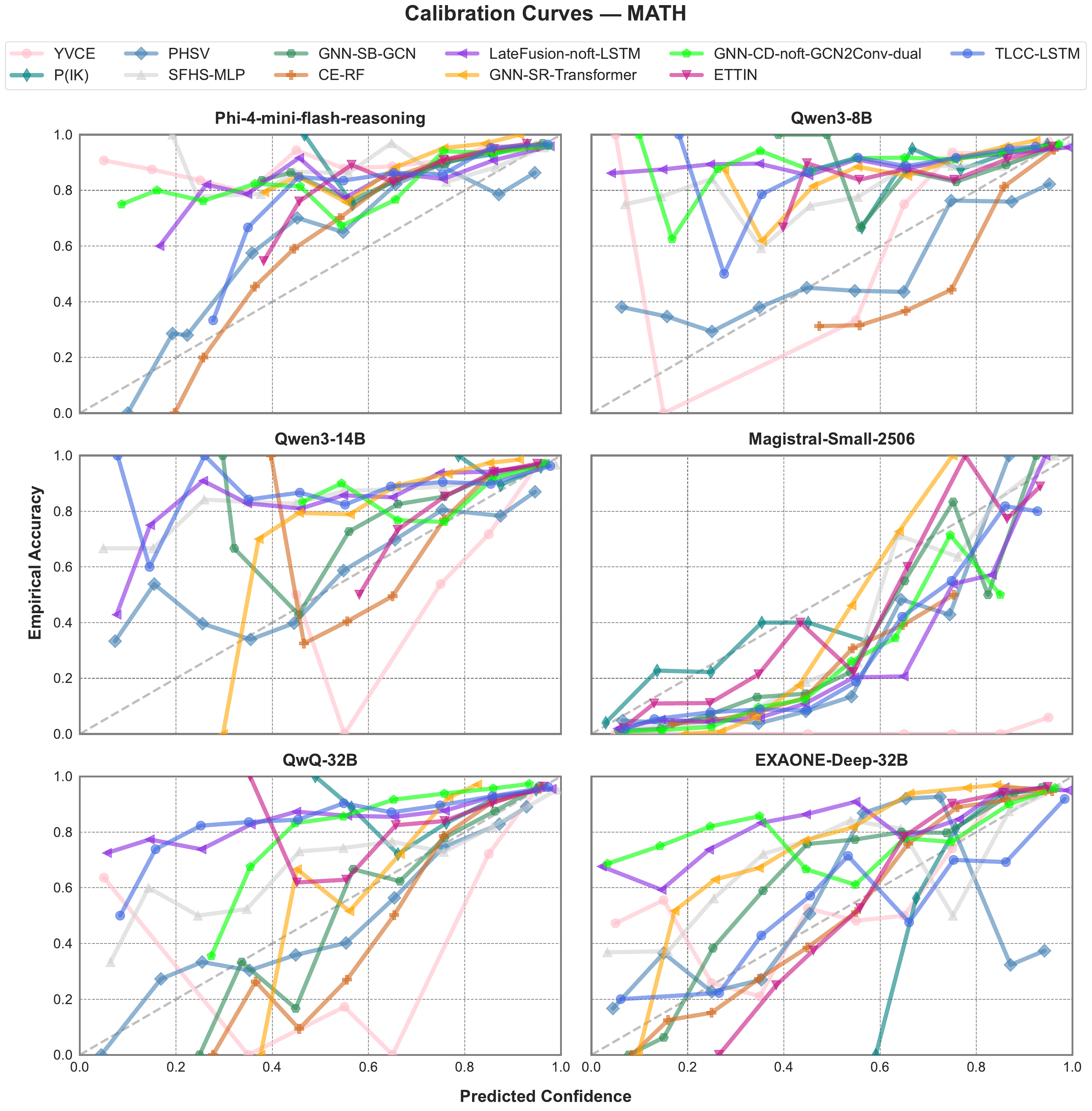}}
\caption{Reliability diagrams for the top-performing method variants on the MATH dataset, aggregated across all LRMs.}
\label{fig:calibration_grid_v3_math_best_10bins}
\end{figure*}

\begin{figure*}[t]
\centering
\makebox[\textwidth][c]{\includegraphics[width=0.9\textwidth]{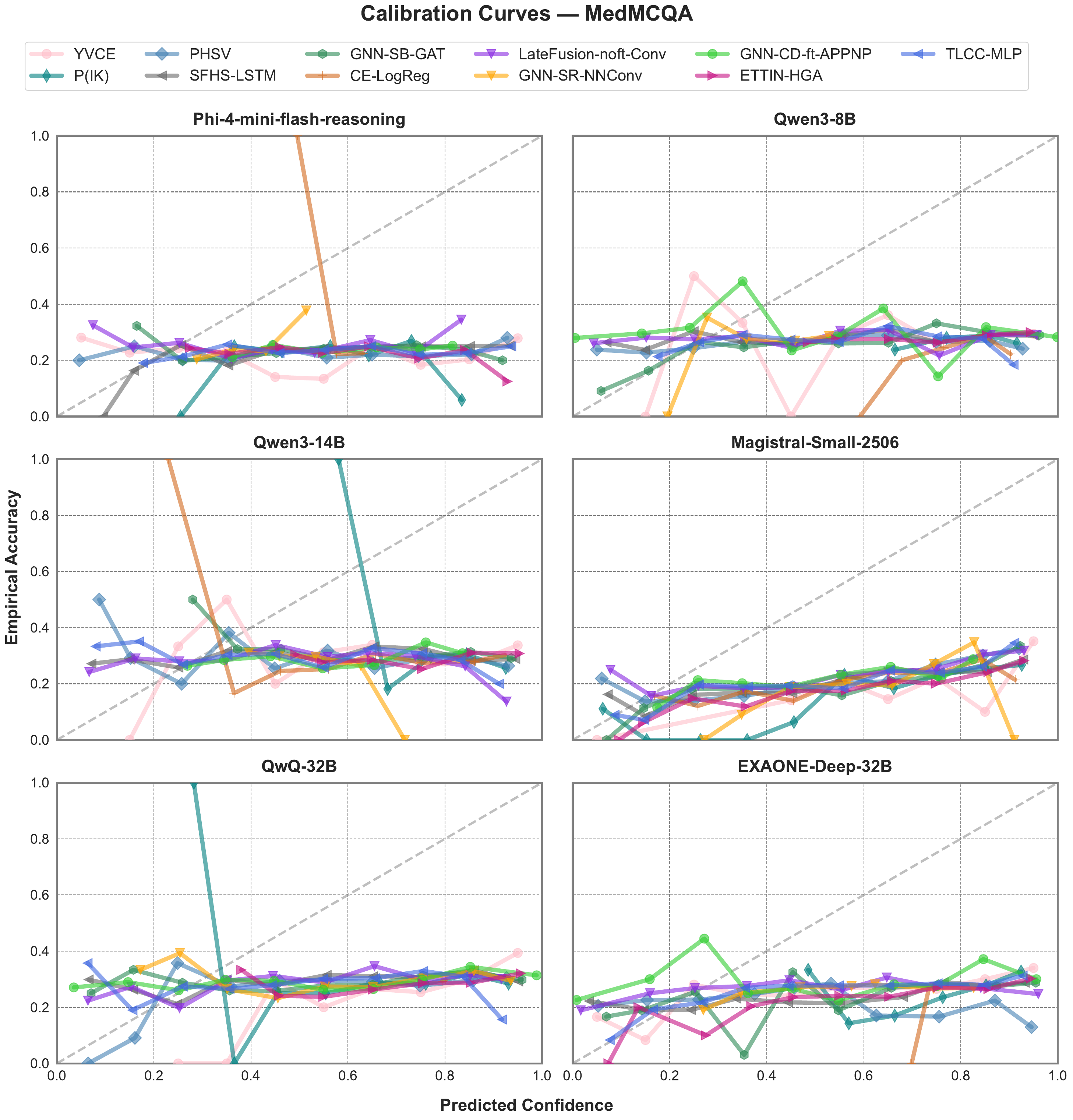}}
\caption{Reliability diagrams for the top-performing method variants on the MedMCQA dataset, aggregated across all LRMs.}
\label{fig:calibration_grid_v3_medmcqa_best_10bins}
\end{figure*}

\begin{figure*}[t]
\centering
\makebox[\textwidth][c]{\includegraphics[width=0.9\textwidth]{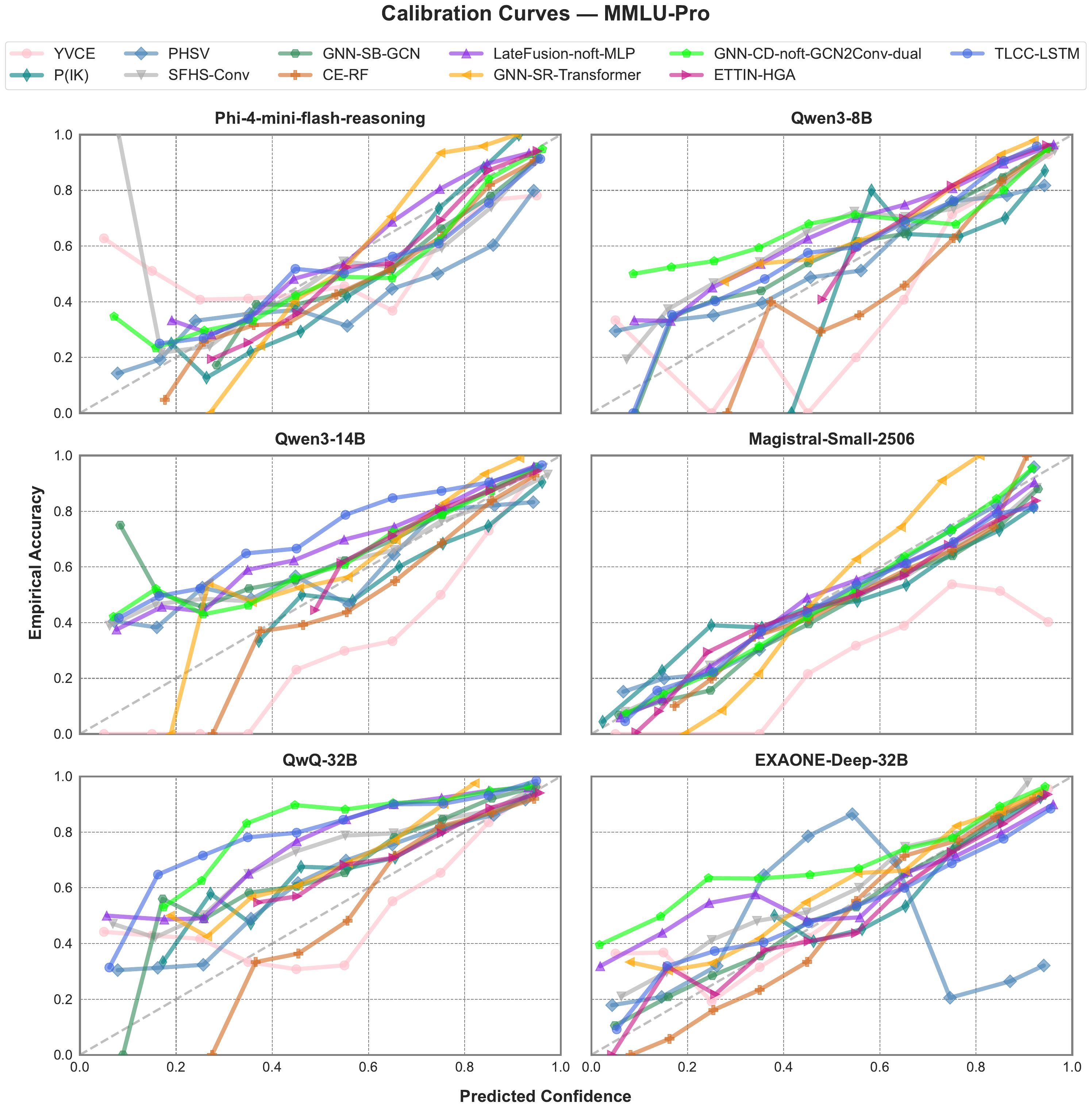}}
\caption{Reliability diagrams for the top-performing method variants on the MMLU-Pro dataset, aggregated across all LRMs.}
\label{fig:calibration_grid_v3_mmlupro_best_10bins}
\end{figure*}

This section provides a more granular breakdown of the calibration performance discussed in the main paper, complementing the overall reliability diagrams shown in Figure~\ref{fig:calibration_grid_v2_dataset_cells_best_10bins_2x3}. Figures~\ref{fig:calibration_grid_v3_bbh_best_10bins}, \ref{fig:calibration_grid_v3_finqa_best_10bins}, \ref{fig:calibration_grid_v3_legalbench_best_10bins}, \ref{fig:calibration_grid_v3_math_best_10bins}, \ref{fig:calibration_grid_v3_medmcqa_best_10bins}, and \ref{fig:calibration_grid_v3_mmlupro_best_10bins} present dataset-specific reliability diagrams for the best-performing variant within each method family, where “best” is defined as the variant achieving the lowest average ECE on the corresponding dataset.

Each plot visualizes calibration quality by plotting the average predicted confidence for a given bin on the x-axis against the empirical accuracy of the predictions within that bin on the y-axis. The dashed diagonal line represents perfect calibration, where the predicted confidence exactly matches the observed accuracy. A method’s curve deviating from this diagonal indicates miscalibration:
\begin{itemize}
\item \textbf{Below the diagonal:} Indicates \textbf{over-confidence}. For example, if predictions in the 50\% confidence bin are correct only 40\% of the time, the corresponding point falls below the line.
\item \textbf{Above the diagonal:} Indicates \textbf{under-confidence}. For example, if predictions in the 60\% confidence bin are correct 70\% of the time, the corresponding point rises above the line.
\end{itemize}
Taken together, Figures~\ref{fig:calibration_grid_v3_bbh_best_10bins} through~\ref{fig:calibration_grid_v3_mmlupro_best_10bins} enable a direct comparison of calibration behavior across domains, illustrating how confidence reliability varies with the underlying reasoning task.

